\pdfoutput=1
\RequirePackage{rotating}

\documentclass[format=acmsmall]{acmart}

\usepackage{
	amsmath,
	booktabs,
	footnote,
	graphicx,
	multirow,
	pgfplots,
	subcaption,
	url,
    hyperref,
    siunitx,
    nicefrac,
    pifont,
    threeparttable
}
\usepackage[keeplastbox]{flushend}
\usepackage[graphicx]{realboxes}

\graphicspath{{fig/}}

\pgfkeys{/pgf/number format/.cd, 1000 sep={\,}}
\pgfplotsset{
	discard if not/.style 2 args={
		x filter/.code={
			\edef\tempa{\thisrow{#1}}
			\edef\tempb{#2}
			\ifx\tempa\tempb
			\else
				
			\fi
		}
	}
}

\newcommand\yup{\ding{52}}
\newcommand\nope{\ding{56}}

\makeatletter
\renewcommand*\env@matrix[1][*\c@MaxMatrixCols c]{%
    \hskip -\arraycolsep
    \let\@ifnextchar\new@ifnextchar
    \array{#1}}
\makeatother

\acmJournal{CSUR}
\acmVolume{52}
\acmNumber{2}
\acmArticle{40}
\acmYear{2019}
\acmMonth{5}
\acmArticleSeq{2}

\setcopyright{acmcopyright}

\acmDOI{10.1145/3309551}

\received{28/09/2018}
\received[revised]{31/12/2018}
\received[accepted]{15/01/2019}

\begin{document}

    \title[Deep Neural Network Approximation for Custom Hardware]{Deep Neural Network Approximation for Custom Hardware:\\Where We've Been, Where We're Going}
    
    \author{Erwei Wang}
    \author{James J. Davis}
    \author{Ruizhe Zhao}
    \author{Ho-Cheung Ng}
    \affiliation{%
        \institution{Imperial College London}
    }
    \author{Xinyu Niu}
    \affiliation{%
        \institution{Corerain Technologies}
    }
    \author{Wayne Luk}
    \author{Peter Y. K. Cheung}
    \author{George A. Constantinides}
    \affiliation{%
        \institution{Imperial College London}
    }
    
    \begin{abstract}
    	Deep neural networks have proven to be particularly effective in visual and audio recognition tasks.
    	Existing models tend to be computationally expensive and memory intensive, however, and so methods for hardware-oriented approximation have become a hot topic.
    	Research has shown that custom hardware-based neural network accelerators can surpass their general-purpose processor equivalents in terms of both throughput and energy efficiency.
        Application-tailored accelerators, when co-designed with approximation-based network training methods, transform large, dense and computationally expensive networks into small, sparse and hardware-efficient alternatives, increasing the feasibility of network deployment.
    	In this article, we provide a comprehensive evaluation of approximation methods for high-performance network inference along with in-depth discussion of their effectiveness for custom hardware implementation.
    	We also include proposals for future research based on a thorough analysis of current trends.
    	This article represents the first survey providing detailed comparisons of custom hardware accelerators featuring approximation for both convolutional and recurrent neural networks, through which we hope to inspire exciting new developments in the field.
    \end{abstract}

%
%

%
%

    
    \thanks{
        The support of the United Kingdom EPSRC (grant numbers EP/K034448/1, EP/P010040/1, EP/N031768/1, EP/I012036/1, EP/L00058X/1 and EP/L016796/1), European Union Horizon 2020 Research and Innovation Programme (grant number 671653), Corerain, Imagination Technologies, Intel, Maxeler, Royal Academy of Engineering, SGIIT, China Scholarship Council and Lee Family Scholarship is gratefully acknowledged.
        
        Authors' addresses: E. Wang, J. J. Davis, P. Y. K. Cheung, G. A. Constantinides, Department of Electrical and Electronic Engineering, Imperial College London, London, SW7 2AZ, United Kingdom. E-mail: \texttt{\{erwei.wang13, james.davis, p.cheung, g.constantinides\}@imperial.ac.uk}. X. Niu, Corerain Technologies, Shenzhen, China. E-mail: \texttt{xinyu.niu@corerain.com}. R. Zhao, H.-C. Ng, W. Luk, Department of Computing, Imperial College London, London, United Kingdom. E-mail: \texttt{\{ruizhe.zhao15, h.ng16, w.luk\}@imperial.ac.uk}.
    }
    
    \maketitle
    
    \renewcommand{\shortauthors}{E. Wang et al.}

	\section{Introduction}
		
        The exponentially growing availability of digital data such as images, videos and speech from myriad sources, including social media and the Internet of Things, is driving the demand for high-performance data analysis.
        Compared to other machine learning algorithms, deep neural networks (DNNs) have achieved dramatic accuracy improvements over the past decade. 
        They have now been employed in a vast range of application domains, from image classification~\cite{BG_CLASSIFICATION_2} and object detection~\cite{BG_OBJECT_DETECTION_1} to autonomous driving~\cite{BG_AUTO_DRIVING_1} and drone navigation~\cite{BG_DRONE_1}.
		Two classes of DNN---convolutional and recurrent (CNNs and RNNs)---are particularly popular.
		While CNNs excel in learning spatial features, RNNs are more suited to problems involving time series.
		
		As tasks increase in complexity, inference architectures become deeper and more computationally expensive.
		For example, a small LeNet-5 model targetting the simple MNIST handwritten digit-classification task requires 680~kop/cl (thousand arithmetic operations per classification, where an arithmetic operation is either an addition or multiplication), while a VGG16 implementation executing the 1000-class ImageNet task requires 31~Gop/cl along with 550~MiB of 32-bit floating-point weight storage~\cite{SURV_EFFICIENT_DNN}.
		The development of algorithms for reducing the computational and storage costs of DNN inference is therefore essential for throughput-, latency- and energy-critical applications.
		Recent work has shown that, with the use of approximation, DNN deployment becomes more feasible thanks to its resultant reductions in memory use and compute complexity.
		
		DNN approximation algorithms can be classified into two broad categories: quantisation and weight reduction.
		Quantisation methods reduce the precision of weights, activations (neuron outputs) or both, while weight reduction removes redundant parameters through pruning and structural simplification.
		By doing so, the latter commonly leads to reductions in numbers of activations per network as well.
		We assess methods of both types in this article since they both contribute to DNN acceleration.
		
		For many years, general-purpose processors (GPPs), particularly multi-core CPUs and GPUs, have been the dominant hardware platforms for DNN inference.
		For uncompressed DNN models, layer operations are mapped to dense floating-point matrix multiplications, which can be efficiently processed in parallel by GPPs following the single-instruction, multiple-data (SIMD) or single-instruction, multiple-thread (SIMT) parallel-processing paradigms.
		With DNN approximation, however, there is an emerging trend of using custom hardware platforms, such as field-programmable gate arrays (FPGAs) and application-specific integrated circuits (ASICs), to accelerate inference instead.
		While GPUs still excel at dense floating-point computation, researchers have reported higher throughput and energy efficiency with custom hardware through the use of low-precision fixed-point quantisation~\cite{FXP_CNN_FPGA_TOWARDS_A_UNIFORM,FXP_CNN_TPU}.
		Moreover, SIMD and SIMT architectures often perform poorly when operating on sparse data; DNNs compressed via fine-grained weight reduction have been shown to execute more efficiently in custom hardware~\cite{R_LSTM_FXP_ESE,PRU_FPGA_POSEWSKY_ZERO_SKIPPING}.
		Logic and memory hierarchy customisability often make custom hardware DNN inference faster and significantly more energy efficient than through the use of GPPs.
		
		A significant number of world-leading information technology firms have selected custom hardware over GPPs for the implementation of their next-generation DNN architectures.
		These include ASICs, \emph{e.g.} Google's Tensor Processing Unit (TPU)~\cite{FXP_CNN_TPU_JOURNAL}, Intel Nervana~\cite{INTEL_NERVANA} and IBM TrueNorth~\cite{FXP_CNN_TRUENORTH}, as well as FPGA-based designs such as Microsoft Brainwave~\cite{MICROSOFT_BRAINWAVE} and Xilinx Everest~\cite{XILINX_EVEREST}.
		In general, ASIC designs can achieve state-of-the art throughput and energy efficiency.
		Their time-consuming and resource-demanding design and fabrication processes, however, make it hard for them to keep up with the rapid evolution of DNN algorithms~\cite{FXP_CNN_TPU,MICROSOFT_BRAINWAVE}.
		
        High-level implementation tools, including Intel's OpenCL Software Development Kit and Xilinx Vivado High-Level Synthesis, and Python-to-netlist neural network frameworks, such as DNNWeaver~\cite{DNNWEAVER}, make the DNN hardware design process for both FPGAs and ASICs faster and simpler.
		Such software allows DNN architects unfamiliar with hardware development to migrate their designs to custom hardware with relative ease.
		Reconfigurability, meanwhile, enables rapid design iteration, making FPGAs ideal prototyping and deployment devices for cutting-edge DNNs.

		Through this survey, we aim to equip researchers new to the field with a comprehensive grounding of DNN approximation, revealing how custom hardware is able to achieve greater performance than GPPs for inference.
		More specifically, we make the following novel contributions:
		
		\begin{itemize}
			\item
				We motivate DNN approximation for custom hardware by comparing the so-called \emph{roofline models}~\cite{BG_ROOFLINE_MODEL} of comparable FPGA, ASIC, CPU and GPU platforms of different scales.
			\item
				We survey key trends in approximation for state-of-the-art DNNs.
				We detail low-precision quantisation and weight-reduction methods, introducing recent algorithmic developments and assessing their relative strengths and weaknesses.
			\item
				We evaluate the performance of custom hardware implementations of each method, focussing on accuracy, compression, throughput, latency and energy efficiency.
			\item
				Based on identified trends, we propose several promising directions for future research.
		\end{itemize}

		There are some existing surveys on DNN approximation.
		Cheng \emph{et al.}~\cite{SURV_CNN_ALGO}, Guo \emph{et al.}~\cite{SURV_FPGA_BASED_NEURAL_NETWORK_ACC}, Cheng \emph{et al.}~\cite{SURV_CNN_FPGA} and Sze \emph{et al.}~\cite{SURV_EFFICIENT_DNN} surveyed algorithms for DNN compression and acceleration.
		Of these, Cheng \emph{et al.}~\cite{SURV_CNN_FPGA} briefly evaluated system-level designs for FPGA implementation.
		Guo \emph{et al.} only surveyed quantisation methods; weight reduction was not mentioned.
		Nurvitadhi \emph{et al.} compared Intel FPGA performance to that of GPU platforms for CNN inference benchmarks~\cite{FXP_CNN_FPGA_GPU}.
		This article represents the first survey that provides not only a comprehensive evaluation of approximation algorithms for efficient DNN inference, but also in-depth analysis and comparison of these algorithms' implementations in custom hardware, covering both CNNs and RNNs.

	\section{Performance Evaluation Metrics}
	\label{sec:metrics}
    
    	We evaluate the effectiveness of DNN approximation by considering the following factors.
        \begin{itemize}
			\item
				\emph{Accuracy}.
                The two accuracy metrics commonly used in machine learning research are \emph{training} and \emph{testing} accuracy, which respectively capture the proportions of correct classifications over training and testing datasets.
                Throughout this article, ``accuracy'' always refers to testing accuracy, which is indicative of a particular DNN's generalisability.
                \emph{Top-$n$} accuracy captures the proportion of testing data for which any of the $n$ highest-probability predictions match the correct results.
                Accuracies are reported as percentages, with changes expressed in percentage points (pp).
                Where comparisons are drawn against baselines, these are uncompressed implementations of the same networks, trained and tested using identical datasets, with all data in IEEE-754 single-precision floating-point format (FP32).
			\item
				\emph{Compression ratio}.
				A network's weight storage requirement \emph{vs} that of the above baseline.
			\item
				\emph{Throughput}.
                Classifications produced per second (cl/s).
                Also known as \emph{classification rate}.
			\item
				\emph{Latency}.
				The end-to-end processing time for one classification, in seconds (s).
			\item
				\emph{Energy efficiency}.
                The throughput obtained per unit power, expressed in cl/J.
		\end{itemize}
    	We also discuss application-specific considerations, \emph{e.g.} parameter tuning time and design flexibility.
    
    \section{Why Custom Hardware? A Roofline Model Analysis}
	\label{sec:motivation}
    
    	For DNN inference, approximation contributes to increases in throughput in three ways: \emph{increased parallelism}, \emph{memory transfer reductions} and \emph{workload reductions}.
        With the help of roofline modelling, we can explain each factor's contribution, revealing why custom hardware can squeeze more speedup from approximation than GPPs.
        
        A roofline model captures the theoretical peak performance of an acceleration platform while also reflecting the effects of off-chip memory data transfers.
        For any high-performance computing engine, the peak \emph{arithmetic performance}, expressed in op/s, is limited by two factors: memory bandwidth and the amount of available compute resources.
        In the context of DNN inference, memory bandwidth limits the rate at which activations can be read and written, as well as that at which parameters stored off-chip can be fetched.
        By compute resources, we mean on-chip parallel-processing units able to perform operations: chiefly multiplication.
        When \emph{memory bound}, the arithmetic performance of a platform does not scale with any increase in parallelism.
        At the \emph{compute bound}, meanwhile, all available processing resources are saturated.
        
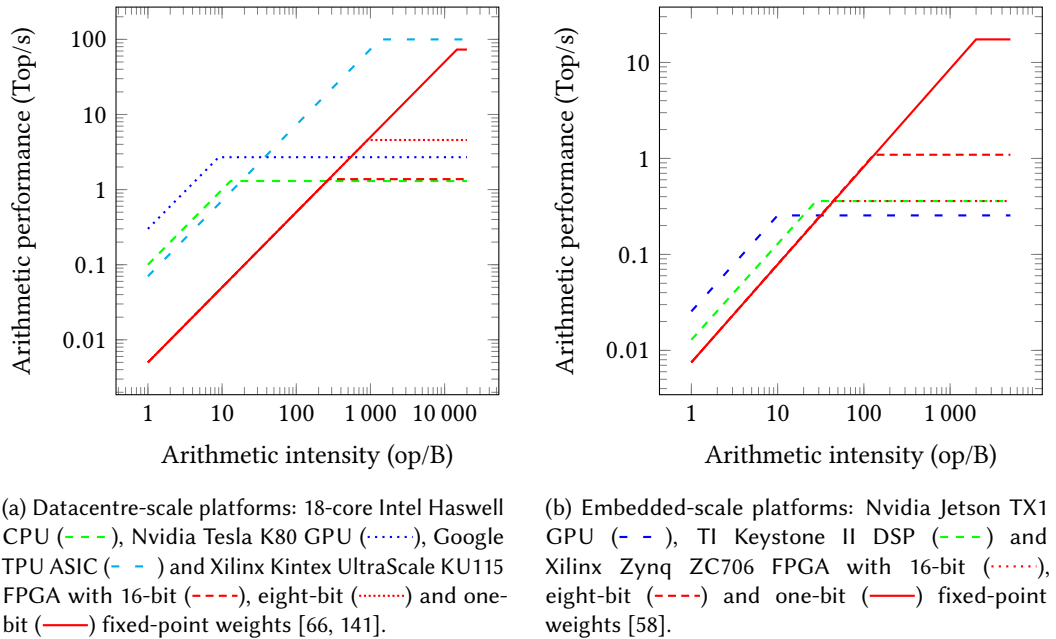
\begin{figure}
    		\begin{subfigure}[t]{0.48\textwidth} 
    			\begin{tikzpicture}

	\begin{axis}[
		width=\textwidth,
		height=\textwidth,
		xlabel={Arithmetic intensity (op/B)},
		ylabel={Arithmetic performance (Top/s)},
		xmode=log,
		ymode=log,
		log ticks with fixed point,
		every axis plot/.append style={thick}
	]

		\addplot[
			loosely dashed,
			color=cyan,
			discard if not={platform}{tpu}
		]
		table[x=intensity, y=throughput]{data/roof.txt};
		\label{plt:roof_tpu}
		
		\addplot[
			dashed,
			color=green,
			discard if not={platform}{cpu}
		]
		table[x=intensity, y=throughput]{data/roof.txt};
		\label{plt:roof_cpu}
		
		\addplot[
			dotted,
			color=blue,
			discard if not={platform}{gpu}
		]
		table[x=intensity, y=throughput]{data/roof.txt};
		\label{plt:roof_gpu}
		
		\addplot[
			densely dashed,
			color=red,
			discard if not={platform}{fpga_16b}
		]
		table[x=intensity, y=throughput]{data/roof.txt};
		\label{plt:roof_fpga_16b}
		
		\addplot[
			densely dotted,
			color=red,
			discard if not={platform}{fpga_8b}
		]
		table[x=intensity, y=throughput]{data/roof.txt};
		\label{plt:roof_fpga_8b}
		
		\addplot[
			solid,
			color=red,
			discard if not={platform}{fpga_bin}
		]
		table[x=intensity, y=throughput]{data/roof.txt};
		\label{plt:roof_fpga_bin}
		
		
		
		
	
	\end{axis}

\end{tikzpicture}
	
    			\caption{Datacentre-scale platforms: 18-core Intel Haswell CPU~(\ref{plt:roof_cpu}), Nvidia Tesla K80 GPU~(\ref{plt:roof_gpu}), Google TPU ASIC~(\ref{plt:roof_tpu}) and Xilinx Kintex UltraScale KU115 FPGA with 16-bit~(\ref{plt:roof_fpga_16b}), eight-bit~(\ref{plt:roof_fpga_8b}) and one-bit~(\ref{plt:roof_fpga_bin}) fixed-point weights~\cite{FXP_CNN_TPU,BNN_CNN_FINN}.}
    			\label{THROUGHPUT_ROOFLINE_DATACENTRE}
    		\end{subfigure}
    		\hfill
    		\begin{subfigure}[t]{0.48\textwidth} 
    			\begin{tikzpicture}

	\begin{axis}[
		width=\textwidth,
		height=\textwidth,
		xlabel={Arithmetic intensity (op/B)},
		ylabel={Arithmetic performance (Top/s)},
		xmode=log,
		ymode=log,
		log ticks with fixed point,
		every axis plot/.append style={thick},
		xtick={1,10,100,1000}
	]

		\addplot[
			loosely dashed,
			color=blue,
			discard if not={platform}{gpu}
		]
		table[x=intensity, y=throughput]{data/roof_embedded.txt};
		\label{plt:roof_embedded_gpu}
		
		\addplot[
			dashed,
			color=green,
			discard if not={platform}{dsp}
		]
		table[x=intensity, y=throughput]{data/roof_embedded.txt};
		\label{plt:roof_embedded_dsp}
		
		\addplot[
			dotted,
			color=red,
			discard if not={platform}{fpga_16b}
		]
		table[x=intensity, y=throughput]{data/roof_embedded.txt};
		\label{plt:roof_embedded_fpga_16b}
		
		\addplot[
			densely dashed,
			color=red,
			discard if not={platform}{fpga_8b}
		]
		table[x=intensity, y=throughput]{data/roof_embedded.txt};
		\label{plt:roof_embedded_fpga_8b}
		
		\addplot[
			solid,
			color=red,
			discard if not={platform}{fpga_bin}
		]
		table[x=intensity, y=throughput]{data/roof_embedded.txt};
		\label{plt:roof_embedded_fpga_bin}
	
	\end{axis}

\end{tikzpicture}
	
    			\caption{Embedded-scale platforms: Nvidia Jetson TX1 GPU~(\ref{plt:roof_embedded_gpu}), TI Keystone~II DSP~(\ref{plt:roof_embedded_dsp}) and Xilinx Zynq ZC706 FPGA with 16-bit~(\ref{plt:roof_embedded_fpga_16b}), eight-bit~(\ref{plt:roof_embedded_fpga_8b}) and one-bit~(\ref{plt:roof_embedded_fpga_bin}) fixed-point weights~\cite{CAFFEPRESSO}.}
    			\label{THROUGHPUT_ROOFLINE_EMBEDDED}
    		\end{subfigure}
    		\caption{Comparison of roofline models of datacentre- and embedded-scale DNN inference platforms.}
    		\label{THROUGHPUT_ROOFLINE}
    	\end{figure}

        Figure~\ref{THROUGHPUT_ROOFLINE} overlays the estimated rooflines of DNN inference accelerators on several hardware platforms.
        The abscissa shows the \emph{arithmetic intensity} of DNN inference, while the ordinate indicates the peak attainable arithmetic performance.
        Arithmetic intensity, also commonly referred to as \emph{operational intensity} or \emph{compute-to-communication (CTC) ratio}, is expressed as the number of arithmetic operations performed per byte of off-chip memory traffic (op/B).
        Arithmetic performance is memory bound when the arithmetic intensity is to the left of the break point.
        When to the right, it is compute bound: resource limitations prevent further scaling.
    
        For fairness, platforms were divided into datacentre and embedded scales and compared accordingly.
        For FPGA-based accelerators, compute bounds were approximated under the assumption that the cost per fixed-point multiply-accumulate (MAC) unit was 2.5 lookup tables (LUTs) for one-bit (binary), 40 LUTs for eight-bit and eight LUTs and half a digital signal processing (DSP) block for 16-bit precision, as suggested by Umuroglu \emph{et al.}~\cite{BNN_CNN_FINN}.
        Both weights and activations were quantised at the same precision.
        We assumed that both Xilinx FPGAs featured, the Kintex UltraScale KU115 and Zynq ZC706, had 4.8~GB/s of off-chip memory bandwidth, and that implementations on the two devices were clocked at 350 and 200~MHz, respectively~\cite{BNN_CNN_FINN}.
    
    	\subsection{Compute Bound Flexibility}
        
        	From Figure~\ref{THROUGHPUT_ROOFLINE}, we can observe that, due to their specialised support for floating-point arithmetic operations, GPUs can deliver the highest arithmetic performance for FP32 DNN inference.
            When moving from floating-point to lower-precision fixed-point data representations, however, custom hardware design flexibility facilitates the trading off of precision for increased performance.
            Being robust to reductions in precision, DNNs can take great advantage of this flexibility~\cite{R_CNN_FXP_DFXP}.
            The ASIC implementation featured, the TPU, has the greatest compute bound---92~Top/s---following which is the KU115 FPGA.
            Since FPGAs afford their users total post-fabrication architectural freedom, different compute bounds are reachable, dependent upon the chosen precision, for the same device.
            As a result, the KU115 has compute bounds of 1.0~Top/s with 16-bit, 3.0~Top/s for eight-bit and 50~Top/s for one-bit fixed-point representations.
            Similarly, the embedded-scale ZC706 can reach 360~Mop/s for 16-bit, 1.0~Top/s for eight-bit and 17~Top/s for binary.
            Compared with custom hardware platforms, GPPs have lower compute bounds since their arithmetic units are designed to perform high-precision operations and are thus inefficient when operating on low-precision data.
            
        \subsection{Arithmetic Performance Increases from Network Compression}
        
        	Reaching a platform's compute bound is only possible if the executing application is not limited by its memory.
            If it is, then, to achieve higher arithmetic performance, higher arithmetic intensity is required.
            With network compression in the form of precision reductions, less off-chip memory needs to be accessed per operation performed, hence higher arithmetic intensity---and subsequently performance, if the application is not compute bound---is achievable.
            Networks can also be compressed via weight reduction, which both saves memory and removes the need to perform the associated operations.
            This can also lead to increased arithmetic intensity and thus performance: a smaller network can use on-chip caching more efficiently, reducing, or even entirely eliminating, off-chip memory traffic~\cite{BNN_CNN_FINN}.
            Performance gains from network compression can be supported from observations from the roofline models, in which, when bounded by memory, an increase in arithmetic intensity means a rightward shift along a roofline, resulting in an increase in arithmetic performance.
            Although all hardware platforms can benefit from network compression, custom hardware implementations, featuring higher compute bounds than GPPs, stand to gain the most; GPPs hit their compute bounds earlier when arithmetic intensity increases.
            
        \subsection{Limitations}
        
        	While roofline models can allow one to predict increases in arithmetic performance (in op/s) that will arise from increased parallelism and memory transfer reductions gained through approximation, they can capture the corresponding changes in throughput (in cl/s) to only a limited extent.
        	To understand the throughput impacts of weight-reduction methods, we must consider an additional factor.
            Arithmetic performance and throughput are related by \emph{workload} (op/cl): the number of arithmetic operations performed per classification.
            Since weight reduction removes unimportant parameters, these methods achieve simultaneous memory transfer and workload reductions.
            As memory transfer reductions can facilitate arithmetic performance increases, it is possible for throughput increases to outpace those in arithmetic performance realised through their employment.
            Quantisation methods, on the other hand, do not cause reductions in workload since the numbers of operations performed per classification remain the same.
            For these, increases in arithmetic performance result in proportionate increases in throughput.
            
        	Roofline modelling does not account for the discrepancies in accuracy that arise from approximation.
            In general, while DNN approximation results in information loss and subsequent accuracy degradation, the majority of works surveyed in this article suggest that the acceptance of low to moderate sacrifices in accuracy can result in significant performance improvement.
            Some show that, in certain scenarios, the introduction of approximation can actually improve accuracy by reducing model overfitting.
            The remainder of this article places great emphasis on the analysis of tradeoffs between network compression and accuracy.
            
            Latency-critical DNN applications, such as advanced driver assistance systems, require the swift production of classifications.
            Many user-interfacing applications also require low latency to maintain adequate user experience~\cite{LATENCY_USER_EXPERIENCE}.
            Roofline models do not inherently capture latency.
            Herein, we detail how custom hardware can achieve state-of-the-art DNN inference latency, as well as throughput, thanks to its flexibility.
            
            Approximation in custom hardware can also achieve superior energy efficiency---another metric whose behaviour is not natively observable through roofline modelling---\emph{vs} competing platforms.
            Custom hardware-based DNN inferencing applications operate at lower clock frequencies and hence consume less power, while also attaining higher throughput and/or lower latency, than those running on GPPs.
            Furthermore, some implementations, by exploiting customisability, outperform GPU-based versions in terms of memory energy efficiency.

	\section{Quantisation}
	\label{sec:Quantisation}

		The first major approximation theme we consider is that of quantisation.
		FPGA and ASIC flexibility permits the implementation of low-precision DNNs, thereby increasing throughput through parallelisation and by reducing reliance on slow off-chip memory.

		\subsection{Fixed-point Representation}
		\label{sec:FXP}

			\subsubsection{Algorithmic Development}
            
                A floating-point-quantised DNN typically allows for an arbitrary binary point position, \emph{i.e.} exponent value, for each individual parameter. 
                This flexibility in data representation range comes at the expense of high resource use, power consumption and arithmetic operation latency, however.
                Fixed-point-quantised DNNs generally use consistent, predetermined precisions and binary point locations, \emph{i.e.} equal maximum and minimum representable magnitudes, for entire networks. 
                This allows for fast, cheap and power-efficient arithmetic operations in hardware, but enforces the use of constant data representation ranges.
                Early works, such as Courbariaux \emph{et al.}'s~\cite{R_CNN_FXP_DFXP}, surveyed this topic, signalling that the accuracy of CNN inference can be preserved even with forward propagation conducted in low-precision fixed-point formats.
                Jacob \emph{et al.} performed eight-bit quantisation of a popular CNN model, MobileNet, reporting an up-to 50\% reduction in inference latency on an ARM CPU with only a 1.8~pp accuracy drop for the Common Objects in Context (COCO) dataset~\cite{FXP_CNN_GOOGLE_MOBILENET}.
                Thereafter, many authors presented FPGA-based CNN and RNN inference frameworks using low-precision fixed-point formats that achieved superior throughputs to their floating-point counterparts with negligible accuracy drops~\cite{NR_CNN_FXP_OPTIM_LOOP,NR_CNN_FXP_CAFFEINE}.
                However, since data in different layers can have very different ranges, using a constant quantisation resolution for an entire network can provide suboptimal bandwidth efficiency.

				Courbariaux \emph{et al.}~\cite{R_CNN_FXP_DFXP}, Qiu \emph{et al.}~\cite{NR_CNN_FXP_GOING_DEEPER} and Shin \emph{et al.}~\cite{R_LSTM_FXP_DFXP} explored using \emph{block floating point (BFP)} for weight and activation quantisation.
                With BFP, often unfortunately referred to as ``dynamic fixed point"~\cite{DFXP_WILLIAMSON}, groups of variables share common binary point locations represented as scaling factors updated during training based on data distributions.
                As such, it can be seen as a compromise between fully floating- and fixed-point formats.
                These authors associated each layer's parameters with a scaling factor, updated after each arithmetic operation by checking the parameters' overflow status during training.
                Their experiments showed that, for both CNNs and RNNs, BFP quantisation of both weights and activations can result in the incursion of below-1.0~pp accuracy losses.
                Since then, BFP has become common in the hardware inference of DNNs as well.
                
                Many authors have explored methods allowing for the automatic selection of layer-wise precision.
                Inspired by Sung \emph{et al.}~\cite{Q-sensitivity-check}, Shin \emph{et al.} proposed the exhaustive search for cost-optimal precisions to use within long short-term memories (LSTMs) through analysis of the tradeoff between signal-to-quantisation-noise ratio (SQNR) and precision~\cite{R_LSTM_FXP_DFXP}. 
                The time complexity of such searches is too high to be practical, however.
				Qiu \emph{et al.} formulated an optimisation problem for minimising quantisation error with respect to changes in precision and binary point location~\cite{NR_CNN_FXP_GOING_DEEPER}.
                A greedy method was proposed for its solution, resulting in desirable layer-wise CNN quantisations.
                Lin \emph{et al.}~\cite{NR_CNN_FXP_SQNR} formulated and solved an SQNR-based optimisation problem to identify the optimal fixed-point precision per layer of a custom-designed CNN, showing that the proposed scheme offered over 1.2$\times$ compression for the CIFAR-10 dataset with no loss in accuracy.
                Their method converts pretrained networks from FP32 into further-quantised equivalents without retraining.
                
                Many authors have focussed on reducing accuracy losses through the modification of rounding schemes.
                Gupta \emph{et al.} trained CNNs with 16-bit fixed-point weight representation using stochastic rounding, achieving lossless compression for the MNIST and CIFAR-10 datasets~\cite{R_CNN_FXP_STOCH}. 
                By following
                \begin{equation}
                    \label{eqn:stochastic_rounding}
                    \text{round}{\left(x\right)} = \begin{cases}
                        \lfloor x \rfloor           & \text{with probability} \ 1-\frac{x-\lfloor x \rfloor}{2^{-f}}    \\
                        \lfloor x \rfloor+2^{-f}    & \text{otherwise},
                    \end{cases}
                \end{equation}
                stochastic rounding results in input $x$ being rounded with resolution $2^{-f}$, where $f$ is the fractional width of the result.
                The probability of rounding $x$ to $\lfloor x \rfloor$ is proportional to the proximity of $x$ to $\lfloor x \rfloor$. 
                Stochastic rounding is thus an unbiased scheme, \emph{i.e.} $\text{E}\left[\text{round}{\left(x\right)}\right] = x$.
				Wu \emph{et al.} proposed WAGE, a CNN framework that discretises gradients using stochastic rounding~\cite{R_CNN_FXP_WAGE}.
                Using two bits for weights and eight bits for activations, gradients and errors, AlexNet trained to classify ImageNet with WAGE exhibited an around-8.8~pp drop in accuracy.
                Shin \emph{et al.} explored treating quantisation resolution as a trainable parameter for both CNNs and RNNs~\cite{R_CNN_FXP_STEP_SIZE}.
                With a tunable quantisation granularity, a four-bit CNN classifying the SVHN dataset and a six-bit RNN performing language modelling each achieved less than 0.1~pp of accuracy loss.
                
                While all of the previously mentioned works featured weights quantised using fixed-point formats, Lai \emph{et al.} implemented CNN inferencing with floating-point weights and fixed-point activations~\cite{NR_CNN_FXP_ACTIVATIONS}.
                Experiments with AlexNet showed that the use of seven-bit floating-point weights could achieve the same accuracy as 11-bit fixed-point representation with ImageNet.
                The authors suggested that weight range is more important than precision in preserving accuracy. 
                This observation laid the foundations for logarithmic quantisation (Section~\ref{sec:exponential}), which trades off precision for range.
                
				The authors of Adaptive Quantisation investigated quantisation at a finer granularity than the aforementioned down-to layer-wise methods~\cite{R_CNN_ADAPTIVE_QUANTISATION}. 
                During retraining, networks adapt, with each filter allowed to assume an independent precision.
				Experiments with small-scale datasets and models showed that Adaptive Quantisation, when combined with pruning, is able to achieve accuracies and compression ratios superior to binarised neural networks, for which each datum is represented using only a single bit.
				A framework for implementing low-precision quantisation, DoReFa-Net, supports arbitrary precisions for weights, activations and gradients, from 32-bit fixed point down to binary~\cite{BNN_CNN_DoReFa-Net}.
				Its authors conducted empirical analysis of various data precision combinations, concluding that accuracy deteriorates rapidly when weights and/or activations are quantised to fewer than four bits.

			\subsubsection{Hardware Implementation}
                
\begin{figure*}
                    \centering
                    \begin{tikzpicture}
\begin{axis}[
	width=0.48\textwidth,
	height=0.4\textwidth,
    view/h=230,
    xlabel=Activation precision (b),
    ylabel=Weight precision (b),
    zlabel=Throughput (cl/s),
    xtick={2,4,6,8},
    ytick={2,4,6,8},
    tick scale binop=\times,
]
  \addplot3[patch, opacity=0.5,blue,shader=interp] 
    table {data/surface_precision_FPGA.txt};
    \addlegendentry{Intel Stratix 10 FPGA};
  \addplot3[patch, opacity=0.5,yellow,shader=flat] 
    table {data/surface_precision_GPU.txt};
    \addlegendentry{Nvidia Titan X GPU};
  \addplot3[domain=2.5:8,mark=none,black, opacity=0.5,thick] ({x},{-0.273*x+3.18},{18700});
\end{axis}
\end{tikzpicture}
                    \caption{Throughput comparison of Intel Stratix~10 FPGA and Nvidia Titan~X GPU AlexNet implementations classifying the ImageNet dataset using various fixed-point weight and activation data representations~\cite{FXP_CNN_INTEL_FCCM}.}
                    \label{FPGA_GPU_PRECISION_COMPARE}
                \end{figure*}
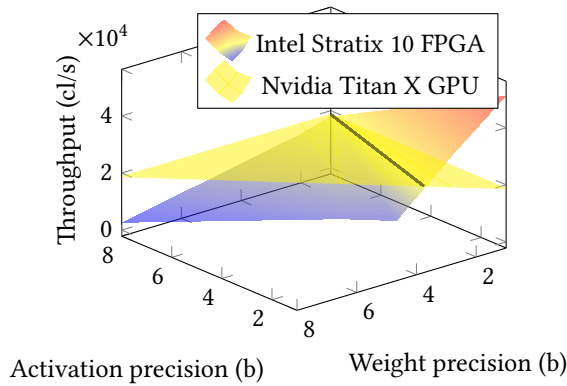

                Nurvitadhi \emph{et al.} conducted experiments to evaluate the performance of Nvidia GPUs and Intel FPGAs for CNN inference using floating- and fixed-point data representations~\cite{FXP_CNN_FPGA_GPU}.
                They concluded that, while their evaluated Stratix-10 FPGA's throughput lagged a Titan~X GPU's with FP32, the FPGA could enable over 50\% greater throughput with six-bit fixed-point data.
                The throughput advantages and energy savings of FPGAs become more significant as precision decreases.
                Colangelo \emph{et al.} presented an Intel FPGA-based inference framework taking advantage of bandwidth and computation savings from low-precision data~\cite{FXP_CNN_INTEL_FCCM}.
                Their experimental results for AlexNet, as presented in Figure~\ref{FPGA_GPU_PRECISION_COMPARE}, showed that, as precision fell, the throughput of their FPGA implementation improved and eventually exceeded that of a GPU of similar scale, supporting the conclusions by Nurvitadhi \emph{et al.}
                The FPGA achieved an order-of-magnitude throughput improvement over the GPU at binary precision.
                Zhang \emph{et al.} showed that a fixed-point-quantised long-term recurrent convolutional network (LRCN) implementation on a Xilinx Virtex~7 VC709 FPGA could achieve a 3.1$\times$ throughput speedup \emph{vs} an Nvidia K80 GPU equivalent~\cite{FXP_RNN_VS_GPU}.
                
                K{\"o}ster \emph{et al.} presented Flexpoint, another BFP variant, for CNN training and inference~\cite{INTEL_FLEXPOINT}.
                Using the ``flex16+5" (16-bit mantissa and five-bit shared exponent) data format, Intel's neural network ASIC, Nervana, was shown to achieve the same accuracy as FP32, while reducing memory bandwidth by around 50\%, for the training of AlexNet and ResNet with ImageNet.
                
                The latest-generation Intel FPGAs can pack up to either one $27\text{-} \times 27$-bit or two $18 \times 19$ MAC(s) per DSP block.
                When using lower precisions on FPGAs, many authors have implemented multipliers using LUTs instead of DSPs to achieve higher resource efficiency.
                Boutros \emph{et al.}~\cite{FXP_CNN_FPL_DSP} proposed the enhancement of DSP blocks to support low-precision MACs with some 12\% area overhead and no drop in achievable frequency.
                One such enhanced DSP can perform one $27 \times 27$ or two $18 \times 19$, four $9 \times 9$ or eight $4 \times 4$ parallel MAC(s).
                The authors implemented AlexNet, VGG-16 and ResNet-50 using the enhanced DSPs.
                On average, they improved the throughput of eight-bit and four-bit DNNs by 1.3$\times$ and 1.6$\times$, respectively, while correspondingly reducing the occupied area by 15\% and 30\% compared to the default use of DSPs in the Intel Arria 10 they targetted.
                
                Sharma \emph{et al.}~\cite{FXP_CNN_BIT_FUSION} and Moons \emph{et al.}~\cite{FXP_CNN_SCALABLE_PRECISION} both introduced variable-precision bit-parallel ASIC implementations.
                Sharma \emph{et al.}'s Bit Fusion consists of an array of bit-level MACs that dynamically fuse to match the precisions of individual DNN layers~\cite{FXP_CNN_BIT_FUSION}.
                Experiments with AlexNet showed that Bit Fusion, while consuming only 900~mW of power, is only 16\% slower than an Nvidia Titan Xp implementation using its native eight-bit vector instructions.
                The Titan Xp can consume up to 250~W of power.
                Moons \emph{et al.} used similar ideas, with their implementation consuming 76~mW to achieve 47~cl/s for AlexNet, outperforming static-precision Eyeriss by 3.9$\times$ in energy efficiency~\cite{FXP_CNN_SCALABLE_PRECISION}. 
                
                Having realised the importance of flexibility of precision in achieving high DNN inference efficiency, GPP manufacturers have recently begun to offer support for low-precision MACs.
                Intel Cascade Lake CPUs provide so-called Vector Neural Network Instructions in 16- and eight-bit formats~\cite{INTEL_CASCADE_LAKE}, while Nvidia Turing GPUs support TensorRT, a deep learning platform integrable with TensorFlow, allowing for low-precision arithmetic down to as few as four bits~\cite{NVIDIA_TURING}.

				We can categorise MACs into two families: bit-parallel and -serial.
				FPGA- and ASIC-based DNN inference architectures with consistent precision generally use bit-parallel MACs for performance and/or simplicity of reuse.
				fpgaConvNet~\cite{NR_CNN_FXP_FPGACONVNET}, Angel-eye~\cite{NR_CNN_FXP_ANGEL-EYE}, ESE~\cite{R_LSTM_FXP_ESE} and works by Chang \emph{et al.}~\cite{NR_LSTM_FXP_Q882} and Shen \emph{et al.}~\cite{FXP_CNN_FPGA_TOWARDS_A_UNIFORM} represent the state-of-the-art in FPGA-based CNN and RNN implementation using low-precision bit-parallel MACs.
                DaDianNao~\cite{FXP_CNN_DADIANNAO}, Cnvlutin~\cite{PRU_ASIC_CNVLUTIN}, NeuFlow~\cite{NEUFLOW} and the TPU~\cite{FXP_CNN_TPU}, meanwhile, are cutting-edge ASIC-based bit-parallel DNN inference platforms.
                For bit-parallel MACs, DNN hardware is typically designed to natively support the maximum precision of an entire network.
                However, as suggested by Khoram \emph{et al.}~\cite{R_CNN_ADAPTIVE_QUANTISATION} and Li \emph{et al.}~\cite{FXP_RNN_MIXED}, since actual precision requirements vary considerably across DNN layers, bit-parallel DNN hardware typically processes an excess of bits per operation.
                Bit-serial alternatives, however, allow precision to be trivially varied at runtime, making their use suitable for fine-grained mixed-precision networks.

                Stripes~\cite{FXP_CNN_STRIPES}, Loom~\cite{FXP_CNN_LOOM} and Bit Pragmatic (PRA)~\cite{FXP_CNN_PRAGMATIC} are ASIC-based DNN accelerators that perform layer-wise mixed-precision inference using bit-serial MACs.
                Among these, experiments showed that Stripes achieved a 1.3$\times$ throughput increase over bit-parallel DaDianNao with VGG-19~\cite{FXP_CNN_STRIPES}.
                Based on Stripes, Albericio \emph{et al.} proposed an ASIC implementation, PRA, which performs bit-serial neuron activations by shifting inputs with respect to the indices of non-zero bits in the weights~\cite{FXP_CNN_PRAGMATIC}.
                Experiments showed that PRA could achieve 2.6$\times$ and 2.0$\times$ increases in throughput and energy efficiency, respectively, \emph{vs} DaDianNao.
                Gudovskiy \emph{et al.} proposed an FPGA implementation, ShiftCNN, using similar ideas to PRA~\cite{EXP_CNN_SHIFTCNN}.
                ShiftCNN was shown to obtain 4.2$\times$ and 3.8$\times$ energy efficiency savings over two baseline CNN platforms using DSP- and LUT-based bit-parallel MACs, respectively.
                Moss \emph{et al.} presented an FPGA-based customisable matrix multiplication framework dedicated to DNN inference~\cite{FXP_CNN_FPGA_HARPv2}.
                Their implementation allows for the runtime switching between static-precision bit-parallel and dynamic-precision bit-serial MAC implementations.
                They observed up-to 50$\times$ throughput increases \emph{vs} FP32 baselines for AlexNet, VGGNet and ResNet.

		\subsection{Binarisation and Ternarisation}
		\label{sec:binary_ternary}

			\subsubsection{Algorithmic Development}

				\emph{Binarisation} is the quantisation of parameters into just two values, typically $\left\{-1,1\right\}$ with a scaling factor.
                Although binary quantisation leads to the incursion of greater error than non-binary fixed-point quantisation, inference operations can be substantially simplified.
				Early works, such as BinaryConnect, focussed on \emph{partial} binarisation, for which only weights are binarised~\cite{BNN_CNN_BinaryConnect}.
				\emph{Full} binarisation of CNNs was proposed in BinaryNet: both weights and activations are binarised~\cite{BNN_CNN_BinaryNet}.
                For binarised training, weights are binarised only during forward propagation; they are not binarised during backward propagation since stochastic gradient descent is sensitive to quantisation and does not work well with very low precisions.
                
                The authors of BinaryConnect and BinaryNet proposed binarisation of two types: deterministic and stochastic.
                For deterministic binarisation, a simple sign function is used, while the stochastic binarisation process is equivalent to stochastic rounding, as was shown in Equation~\ref{eqn:stochastic_rounding}.
                Since the derivative of the sign function is a Dirac delta function with zero everywhere but the origin, rendering the training process impossible, the authors of BinaryNet resorted to using a hard hyperbolic tangent (tanh) function to cope with this problem during backward propagation~\cite{BNN_CNN_BinaryNet}:
                \begin{equation*}
                    \text{tanh}_\text{hard}{\left(x\right)} = \begin{cases}
                        1   & \text{if} \ x > 1             \\
                        x   & \text{if} \ -1 \leq x \leq 1  \\
                        -1  & \text{otherwise}.
                    \end{cases}
                \end{equation*}
                In this way, the gradient of their cost function could be preserved for weights within $\left[-1,1\right]$ during training.
                Clipping was also applied to the real-valued weights to constrain them to $\left[-1,1\right]$.
				Experiments with the MNIST and CIFAR-10 datasets on unidentified networks showed that BinaryConnect achieved around 1--2~pp higher prediction accuracies than FP32 baselines.
                The authors suggested that this was due to stochastic rounding's regularisation effect, whence randomisation is injected into a network in a similar way to ``dropout" in the form of per-neuron binarisation noise~\cite{BG_DROPOUT}.
                Experiments with BinaryNet with MNIST, CIFAR-10 and SVHN---also on unknown networks---showed less-than 1~pp accuracy losses compared to baseline cases.
                However, this regularisation effect was only seen for small datasets.
                For large-scale ones such as ImageNet, although BinaryNet with AlexNet achieved significant memory and computational complexity reductions, this was accompanied by around 30~pp top-one accuracy drops.
                Binarisation's high error inducement outweighed the positives of regularisation in these cases.

                In an effort to improve BinaryNet's data representation, XNOR-Net features trainable filter-wise scaling factors for forward propagation~\cite{BNN_CNN_XNOR-Net}. 
                These scaling factors retain the average magnitudes of weights and activations in order to improve the expressiveness of binarised networks.
                Experiments with XNOR-Net inferencing AlexNet with the ImageNet dataset showed that this method successfully improved top-one accuracy by around 20~pp compared with BinaryNet, while there was still an accuracy drop of over 10~pp \emph{vs} a FP32 baseline.
                XNOR-Net does, however, require averaging operations over input features, adding costly high-precision dividers~\cite{BNN_CNN_REBNET_FCCM}.
				
				ABC-Net alleviates the information loss from binarisation by approximating FP32 parameters and activations as linear combinations of multiple binary values~\cite{BNN_CNN_ABC-Net}.
                Its authors pointed out that, during forward propagation, their $K$-binarisation scheme ($K$ parallel bitwise XNORs) is cheaper than performing $K$-bit fixed-point multiplication, emphasising ABC-Net's superior resource efficiency over conventional fixed-point CNN implementations.
				A five-bit weight/activation ABC-Net achieved a 14~pp top-one accuracy improvement \emph{vs} XNOR-Net with ImageNet on ResNet-18.
				
				Tang \emph{et al.} proposed a number of improvements to the binarised retraining process~\cite{BNN_CNN_BINARY_CONSTRIANED_TRAINING}.
                One of their discoveries was that a low learning rate is preferable in order to avoid frequent parameter oscillation, which leads to prolonged and inefficient training.
				Furthermore, a binary-constrained regulariser was added to their training loss function to encourage more bipolar weight values (closer to $\pm$1).
				This was implemented within the function as
                \begin{equation}
                    \label{eqn:bnn_bipolar_regulariser}
                    \text{loss}_\text{post-reg}{\left(\boldsymbol{W},\boldsymbol{b}\right)} = \text{loss}_\text{task}{\left(\boldsymbol{W},\boldsymbol{b}\right)} + \lambda \sum^L_{l=1}{\sum^{N_l}_{n=1}{\sum^{M_l}_{m=1}{1-\boldsymbol{W}_{lnm}^2}}},
                \end{equation}
                wherein $\boldsymbol{W}$, $\boldsymbol{b}$ and $\lambda$ represent weight, bias and regularisation factor, respectively. 
                $L$ is the network's depth and $M_l$ and $N_l$ are the input and output channel numbers in the $l$\textsuperscript{th} layer.
                $\text{loss}_\text{task}{\left(\boldsymbol{W},\boldsymbol{b}\right)}$ returns the task-related loss based on the original network settings, while $\text{loss}_\text{post-reg}{\left(\boldsymbol{W},\boldsymbol{b}\right)}$ gives the post-regularisation loss.
                Tang \emph{et al.}'s regulariser penalised with respect to the implemented network's overall quantisation loss.
				These optimisations, together with multi-bit activation representation, resulted in a 6.4~pp top-one AlexNet accuracy increase over XNOR-Net for ImageNet.
				
				Going further, HWGQ addressed the problem of mismatching gradients between the binarised forward activation function, sign, and the backward activation function, hard tanh~\cite{BNN_CNN_HWGO}.
				HWGQ uses a half-wave Gaussian-quantised (HWGQ) rectified linear unit (ReLU) for forward propagation and a standard ReLU function for backward propagation.
                The authors' experiments with AlexNet produced a 47\% top-one ImageNet error rate: the lowest achieved for a binary network to date.
                
                Ott \emph{et al.} suggested that RNNs are not amenable to binarisation since the large quantisation losses of near-zero values forced to $\pm$1 get amplified over their recursions~\cite{TNN_LSTM_BENGIO}.
                Nevertheless, Liu \emph{et al.} implemented binarisation in LSTMs targetting English and Chinese language modelling, although they only applied it to input and output embedding layers (those that encode text as vectors)~\cite{BNN_LSTM_LM}.
                The authors reported up to 11$\times$ compression of those layers without accuracy loss.
                Given these seemingly conflicting conclusions, further experiments are required to establish the effectiveness of binarisation in RNNs.

				Adding zero to the binary value set gives \emph{ternary} representation.
				TernaryConnect~\cite{TNN_CNN_BENGIO} and Ott \emph{et al.}'s work~\cite{TNN_LSTM_BENGIO} introduced ternary CNNs and RNNs, respectively, for improved accuracy.
                The accuracies of TernaryConnect exceeded the previous-best results for MNIST, CIFAR-10 and SVHN reported by the authors of BinaryConnect~\cite{BNN_CNN_BinaryConnect}.
				For each layer $l$, Ternary Weight Networks (TWNs) use tunable symmetric thresholds $\pm\delta_l$ to differentiate 0 from $\pm$1~\cite{TNN_CNN_TWN}.
				For an AlexNet implementation classifying ImageNet, TWNs achieved a 46\% top-one error rate: lower than all binarised neural networks reported thus far.
				In Trained Ternary Quantization, parameters are represented in the form $\{w_l^-,0,w_l^+\}$, wherein $w_l^-$ and $w_l^+$ are trainable~\cite{TNN_CNN_TTQ}.
				Compared with TWNs, a further accuracy improvement---around 5~pp---was reported for AlexNet with ImageNet.
                
                Mellempudi \emph{et al.} presented Fine-grained Quantisation (FGQ)~\cite{TNN_CNN_FGQ}, which involves the ternarisation of a pretrained FP32 network into groups, then ternarising each group independently.
                Within a group $g$, the ternary weights can have distinct quantisation levels $\left\{-w_g, 0, w_g\right\}$.
                Although groups can be determined arbitrarily, in this case the authors grouped by channel to promote implementational efficiency.
                Assuming that a network has $G$ such groups, there are $2G + 1$ distinct levels with which to represent weights in total, increasing the model's representation capacity over ternarisation with equal granularity.
                Weights are partitioned along channels for simplicity.
                Experiments with ImageNet showed that an FGQ-quantised AlexNet with ternary weights and four-bit activations suffered 7.8~pp accuracy loss compared to the baseline.
                
                Alemdar \emph{et al.} combined ternarisation with knowledge distillation, in which shallower ``student" networks are used to mimic deeper ``teachers"~\cite{KD_TNN_FPGA}.
                In hardware, ternarisation requires cheaper arithmetic operators than higher-than-two-bit fixed-point quantisation.
                To improve the accuracy of a ternary student network, stochastic rounding (Equation~\ref{eqn:stochastic_rounding}) is used while ternarising during teacher network backward propagation.
                Experiments with MNIST, CIFAR-10 and SVHN on arbitrarily chosen models showed that ASIC implementations of this work achieved 3.1$\times$ greater energy efficiency, on average, than IBM TrueNorth executing the same benchmarks with ternary data~\cite{FXP_CNN_TRUENORTH}.

                While low-precision networks lead to significant network compression, they often require higher numbers of neurons to achieve accuracies comparable to their floating-point counterparts.
                For the CIFAR-10 dataset, for example, binary networks such as FINN and ReBNet require a wider and deeper model, CNV, in order to achieve similar accuracy to an FP32 baseline with CifarNet, a much thinner and shallower model~\cite{TF_SLIM_MODEL_LIBRARY}.
                Zhu \emph{et al.} proposed the Binary Ensemble Neural Network (BENN), in which multiple binarised networks are aggregated by ``boosting" (parallel ensemble with trained weights)~\cite{BNN_CNN_BENN}.
                The authors showed that their network ensembles exhibited lower bias and variance than their individual constituents while also having improved robustness to noise.
                Experiments with AlexNet on the ImageNet dataset showed that the use of BENN, with AdaBoost (adaptive boosting) and an ensemble of six binarised networks, led to only 2.3~pp of top-one accuracy loss \emph{vs} an FP32 baseline.
                The authors of WRPN explored the same phenomenon by gradually reducing network precision and increasing the number of channels of an originally FP32 network, finding that, by increasing model complexity, a low-precision network can eventually match or even surpass the accuracy of its baseline.
                Further research is required to identify models that are particularly amenable to low-precision inference~\cite{BNN_CNN_WRPN}.

			\subsubsection{Hardware Implementation}
            
                For inference, binary networks have several properties that enable elegant mapping to Boolean operations.
                With a set bit representing 1 and an unset bit -1, multiplication becomes an XNOR operation: significantly cheaper to implement than non-binary fixed-point multiplication.
                Furthermore, accumulation becomes a population count (popcount) operation, which, on an FPGA, requires half the LUTs of an equivalent adder tree~\cite{BNN_CNN_FINN}.
                Umuroglu \emph{et al.}~\cite{BNN_CNN_FINN} and Ghasemzadeh \emph{et al.}~\cite{BNN_CNN_REBNET_FCCM} suggested that, during binary inference, operations in batch normalisation can be simplified to binary thresholding, where $y = \text{sign}(\alpha x - b) = \text{sign}(x - \nicefrac{b}{\alpha})$.
                $x$, $\alpha$, $b$ and $y$ are the input, scaling factor, bias and output, respectively.
                A max-, min- or average-pooling layer in a binary network can be efficiently implemented using OR, AND or majority functions.
                
                On GPUs, 32 one-bit activations and weights can be packed into each word to perform bit-wise XNORs.
                On a Titan X Pascal GPU, 32 32-bit popcounts can be issued per cycle per streaming multiprocessor (SM).
                Thus, up to 512 binary MAC operations can be performed per cycle per SM.
                As it can issue up to 128 FP32 MAC instructions per cycle per SM, however, it can be estimated that the theoretical peak throughput gain of a binary network over FP32 for that GPU is only 4$\times$~\cite{FXP_CNN_FPGA_GPU}.

                On FPGAs, binary network inference can show more significant performance gains.
                Many frameworks, including FINN~\cite{BNN_CNN_FINN}, FP-BNN~\cite{BNN_CNN_FP-BNN} and that from Moss \emph{et al.}~\cite{FXP_CNN_FPGA_HARPv2}, have been built to achieve this, resulting in orders of magnitude higher throughput and energy efficiency than floating-point counterparts of comparable scale.
                FINN's authors constructed small binary networks for the MNIST, CIFAR-10 and SVHN datasets targetting the Xilinx Zynq ZC706 FPGA. 
                Experiments with the CNV network (110~Mop/cl) resulted in sustained throughput of 22~kcl/s---the highest throughput at the time of publication---while consuming as little as 25~W of power.
                The authors of FP-BNN implemented AlexNet (2.3~Gop/cl), one of the larger CNNs, on an Intel Stratix~V FPGA, reporting a throughput of 870~cl/s, 2.7$\times$ faster than a 235~W-consuming Tesla K40 GPU executing the same binary network, while drawing only 26~W of power.
                On a smaller custom network designed for CIFAR-10 inference (1.2~Gop/cl), in which arithmetic intensity was higher, FP-BNN achieved a peak throughput of 7.6~kcl/s.
                Moss \emph{et al.} showed that, with binarisation, the HARPv2 heterogeneous platform could achieve a peak throughput of 110~cl/s for VGGNet, with 1.2$\times$ greater energy efficiency than a Titan X Pascal GPU-based alternative~\cite{FXP_CNN_FPGA_HARPv2}.
                
                The authors of ReBNet implemented ``residual binarisation" on FPGAs~\cite{BNN_CNN_REBNET_FCCM}: similar to ABC-NET's aforementioned $K$-binarisation scheme~\cite{BNN_CNN_ABC-Net}.
                They observed accuracy improvements when higher data widths were used, as was the case for ABC-Net. 
                ReBNet's authors reported that their work exposes a continuum between accuracy and area, making it amenable to a wide range of application requirements and hardware constraints.
                
                Prost-Boucle \emph{et al.} implemented ternary CNNs on a Xilinx Virtex-7 VC709 FPGA, presenting both high-performance- and low-power-targetting designs~\cite{TNN_CNN_FPL}.
				Their experiments with the CNV model classifying CIFAR-10 demonstrated a 6.6~pp accuracy improvement compared to FINN's binarised inference.
                In high-performance mode, up to 27~kcl/s was achieved with around 13~W of power consumption while, in low-power mode, 14~kcl/s was obtained for half the power.
                
                The authors of YodaNN introduced a 65~nm ASIC implementation featuring partial binarisation, in which activations and weights are quantised to 12 and one bit(s), respectively~\cite{BNN_CNN_YODANN}.
                Experiments with AlexNet and the ImageNet dataset showed that YodaNN achieved a throughput of 0.50~cl/s and an energy efficiency of 2.0~kcl/J at 0.60~V.

		\subsection{Logarithmic Quantisation}
		\label{sec:exponential}

			\subsubsection{Algorithmic Development}

				In a base-two logarithmic representation, parameters are quantised into powers of two with a scaling factor.
				Suiting the observation that a weight's representation range is more important than its precision in preserving network accuracy, logarithmic representations can cover wide ranges using few bits~\cite{NR_CNN_FXP_ACTIVATIONS}.
				While logarithmic representation can also be used for activations, this has yet to be explored.
                LogNet's authors quantised CNNs with weights encoded in a four-bit logarithmic format, after which they performed retraining to recover some lost accuracy~\cite{EXP_CNN_LOGNET}.
                Their experiments with the ImageNet dataset revealed 4.9~pp and 4.6~pp top-five accuracy drops for AlexNet and VGG16, respectively.
				In Incremental Quantisation (INQ), weights are iteratively quantised into a logarithmic format, with activations left as eight-bit fixed point values~\cite{EXP_CNN_INCR}.
                In each iteration, parameters in each layer are partitioned into two groups using a threshold on absolute parameter values.
				The group with higher absolute values is quantised into powers of two directly, whereas the other is retrained in the following iteration in FP32 to compensate for losses.
				This process repeats until all parameters are quantised.
				Experiments with ImageNet on AlexNet showed a negligible ($\sim$0.1~pp) accuracy loss against the baseline while using only five bits per weight.

			\subsubsection{Hardware Implementation}
                
                For hardware inference, base-two logarithmic representations see multiplications converted into binary shifts for greater area and energy efficiencies as well as speed.
                GPPs perform binary shifts using shifters embedded in arithmetic and logic units, most of which can move their operands by an arbitrary number of bits per operation.
                On an Nvidia Maxwell GPU, the theoretical peak throughput of 32-bit binary shifts is 50\% of that of FP32 MACs~\cite{CUDA_BINARY_SHIFT}.
                
				In custom hardware, a multiplication between an exponentially quantised weight parameter and an activation can be implemented cheaply using a variable-length binary shifter.
				With LogNet, CNN inference is performed on FPGAs with four-bit logarithmic-quantised weights~\cite{EXP_CNN_LOGNET}.
				Experiments with three convolutional layers showed an over-3.0$\times$ energy efficiency improvement \emph{vs} an Nvidia Titan X GPU implementation, while a four-bit logarithmic implementation of AlexNet demonstrated an around-5~pp accuracy loss for ImageNet.
                Wang \emph{et al.} implemented base-two logarithmic quantisation on weights associated with input, output and forget gates in LSTMs while leaving the remaining gates in non-logarithmic eight-bit fixed-point precision~\cite{SM_LSTM_HYBRID_CIRCULAR}.
                In their 90~nm ASIC implementation, multiplications with logarithmic-quantised weights are implemented with shift-and-add operations, which occupy significantly less area than MACs using non-logarithmic fixed-point quantisation.
                Wang \emph{et al.}'s ASIC was able to process a $512\times 512$ LSTM layer within 1.7~$\mu$s at a silicon area cost of 31~mm\textsuperscript{2}.
                
                The implementations mentioned above reuse binary shifters over different groups of weights for scalability.
				For custom hardware, if shift amounts are constant, no logic is required for multiplication: they can be performed in routing alone.
                This means that fixing DNN parameters using constant-length shifts instead of multiplications can result in significant resource and latency savings.
                Server-scale platforms with massive resource availability, such as Microsoft Catapult~\cite{MICROSOFT_CATAPULT} and Amazon Web Services, should be able to benefit hugely from such optimisations.

	\section{Weight Reduction}
	\label{sec:weight_reduction}

		Let us now turn to DNN approximation's second key subject: weight reduction.
		Here, parameters deemed unimportant are eliminated entirely.
        Weight reduction improves the performance of hardware inference by reducing both workload and off-chip memory traffic.

		\subsection{Pruning}
		\label{sec:pruning}

			\subsubsection{Algorithmic Development} 

				Pruning is the process of removing redundant connections in a DNN.
				Inspired by early works including Optimal Brain Damage~\cite{PRU_CNN_OPTIMAL_BRAIN_DAMAGE} and Optimal Brain Surgeon~\cite{PRU_CNN_OPTIMAL_BRAIN_SURGEON}, Srinivas \emph{et al.} proposed a retraining-free method for removing redundant neurons in trained CNNs~\cite{PRU_CNN_DATA_FREE}.
                Similar neurons can be wired together and hence pruned away.
                The authors proposed the similarity evaluation of neurons using a matrix of their squared Euclidean distances.
                This method resulted in 6.7$\times$ and 1.5$\times$ compression for the MNIST and AlexNet networks, respectively.
				Experiments with AlexNet revealed 2.2~pp of ImageNet accuracy loss.
				
				Han \emph{et al.} were the first to propose an iterative pruning process~\cite{PRU_CNN_TRAIN_PRUNE_RETRAIN}.
                In their work, one iteration consists of pruning followed by retraining, allowing the remaining connections to learn to compensate for the pruning loss.
                After many such iterations, lossless compression ratios of 9.0 and 13 were achieved for AlexNet and VGG16, respectively, both classifying the ImageNet dataset.
				The authors attempted to promote sparsity in the networks by penalising non-zero parameters with an $l_1$ or $l_2$ norm-based sparsity regulariser~\cite{GROUP_SPARSITY_REGULARISER} during retraining.
                An $l_2$ norm-based sparsity regulariser can be implemented as
                \begin{equation}
                    \text{loss}'_\text{post-reg}{\left(\boldsymbol{W},\boldsymbol{b}\right)} = \text{loss}_\text{task}{\left(\boldsymbol{W},\boldsymbol{b}\right)} + \lambda \sqrt{\sum^L_{l=1}{\sum^{N_l}_{n=1}{\sum^{M_l}_{m=1}{\boldsymbol{W}_{lnm}^2}}}},
                    \label{eqn:cnn_sparsity_regulariser}
                \end{equation}
                wherein $\boldsymbol{W}$, $\boldsymbol{b}$, $\lambda$, $L$, $M_l$, $N_l$ and $\text{loss}_\text{task}{\left(\boldsymbol{W},\boldsymbol{b}\right)}$ share Equation~\ref{eqn:bnn_bipolar_regulariser}'s definitions, while $\text{loss}'_\text{post-reg}{\left(\boldsymbol{W},\boldsymbol{b}\right)}$ gives the post-regularisation loss.
                During training, this regulariser penalises Han \emph{et al.}'s loss function with respect to the magnitudes of non-zero weights, resulting in more weights near zero.
				See \emph{et al.} implemented a similar strategy for RNNs, finding that their 5.0$\times$-compressed network for neural machine translation actually surpassed the baseline's accuracy for the WMT'14 dataset due to the effect of regularisation~\cite{PRU_LSTM_NMT_TRAIN_PRUNE_RETRAIN}.
                
                Following the idea of incorporating sparsity into training objective functions, Zhou \emph{et al.} implemented low-rank constraints~\cite{TENSOR_LOW_RANK_CONSTRAINTS,PRU_CNN_LESS_IS_MORE}.
                The authors aimed to induce lower average ranks in weight matrices using a group sparsity constraint with a regulariser of the form of Equation~\ref{eqn:cnn_sparsity_regulariser}.
                They achieved an AlexNet compression ratio of 4.3, inducing 0.57~pp of top-one ImageNet accuracy loss.
				
				Inspired by Han \emph{et al.}'s work, the authors of Dynamic Network Surgery (DNS) performed pruning followed by ``splicing," wherein the salience (importance) of the remaining parameters is evaluated; parameters' salience varies when others are removed~\cite{PRU_CNN_DNS}.
				DNS achieved 110$\times$ and 18$\times$ compression for LeNet-5 and AlexNet, respectively.

				The proposals above all see DNNs pruned at element-wise granularity, often referred to as \emph{fine-grained} pruning.
                Although pruning at the finest granularity leads to excellent compression ratios, it can also result in significant irregularities in weight distribution, which, in turn, can make it difficult for the inference hardware to convert compression into increased throughput.
                \emph{Coarse-grained} pruning methods have hence been proposed, which produce larger but denser networks than those resulting from fine-trained pruning.
				Lebedev \emph{et al.} introduced Structured Brain Damage, wherein a group-wise sparsification regulariser (Equation \ref{eqn:cnn_sparsity_regulariser}) shapes each weight matrix's non-zeroes into a regular, dense pattern~\cite{PRU_CNN_GROUP_WISE_BRAIN_DAMAGE}.
				Experiments showed 3.0$\times$ improvements in both compression ratio and throughput with sub-1.5~pp accuracy degradation for AlexNet classifying ImageNet.
				Wen \emph{et al.}~\cite{PRU_CNN_STRUCTURED_SPARSITY}, Li \emph{et al.}~\cite{PRU_CNN_LASSO_ALONG_FILTERS}, He \emph{et al.}~\cite{PRU_CNN_CHANNEL_PRUNING} and Su \emph{et al.}~\cite{PRU_CNN_ALEX_SU} performed structured pruning along channels, filters, layers and shapes (arbitrary groups of parameters) of CNNs.
                All of these works proposed the pruning of groups of redundant parameters based on sums of parameter magnitudes where, intuitively, those with lower values are deemed less important.
				
				The authors of Network Slimming argued that, although sparsity can be realised at different granularities, pruning at the channel level provides a tradeoff between flexibility and ease of hardware implementation~\cite{PRU_CNN_SLIMMING}.
                The output of Network Slimming is simply a ``thinned" version of an unpruned network.
                With every convolutional and fully connected layer followed by a batch-normalisation layer, networks are trained before pruning such that batch normalisation scaling factors represent the relative importance of each channel.
                Layer-wise pruning is then performed by thresholding them.
                An $l_1$ sparsity regulariser is used on the scaling factors, instead of each parameter, in order to promote channel-wise sparsity.
                20$\times$ compression and a 5$\times$ workload reduction were reported against an unpruned baseline for VGGNet.
				Experiments with ImageNet on the VGG-A model demonstrated about-5.8$\times$ compression with less than 0.1~pp of accuracy loss.

                Decisions on whether to prune specific parameters are based on parameter salience. 
                Establishing accurate salience estimations is thus crucial for pruning effectiveness. 
                Molchanov \emph{et al.} proposed and compared various criteria for determining weight salience, including pruning by the magnitude, mutual information (against classification ground truth) and Taylor expansion of quantisation noise~\cite{PRU_CNN_SALIENCY_CRITERIA}.
                Of these, the Taylor expansion-based criterion was found to perform particularly well. 
                Unlike the works above, which all defined parameter salience as the impact on accuracy, Yang \emph{et al.} defined it as the impact on energy efficiency, achieving an energy saving of 3.7$\times$ with ImageNet on AlexNet against an Nvidia Titan~X GPU equivalent~\cite{SALIENCY}.

			\subsubsection{Hardware Implementation}
                
                Coarse-grained pruning produces outputs in structured and dense patterns such that the Basic Linear Algebra Subprograms (BLAS) for GPPs can directly benefit from reductions in workload.
                It is more challenging for GPPs to benefit from fine-grained pruning, however.
                Modern GPUs follow a SIMT execution model, in which threads execute the same sequence of instructions on different data.
                Compute speed is thus bottlenecked by the slowest thread; others remain idle until synchronisation points are reached.
                Checking for zeroes in matrices adds extra instructions to each thread, further reducing computational efficiency.
                An alternative approach is to use linear algebra libraries supporting zero-skipping, such as sparse matrix-vector multiplication (SPMV).
                Monakov \emph{et al.} proposed a matrix storage format that improves locality and enables automatic parameter tuning on GPUs~\cite{PRU_GPU_MONAKOV}.
                Bell \emph{et al.} implemented data structures and algorithms for SPMV on an Nvidia GeForce~GTX~280 GPU, with which they achieved state-of-the-art FP32 performance~\cite{PRU_GPU_BELL}.
                For SPMV to show performance and/or memory storage advantages, however, matrices need to be highly sparse.
                This is often the case for RNNs, which normally have over 80\% sparsity~\cite{R_LSTM_FXP_ESE}, but is not usually true for CNNs (typically only 5--50\% sparsity)~\cite{PRU_CNN_GROUP_WISE_BRAIN_DAMAGE}.

				Custom hardware can handle irregular, sparse data more efficiently than GPPs for fine-grained-pruned DNNs.
                Li \emph{et al.} presented an FPGA design framework for CNN sparsification and acceleration~\cite{PRU_FPGA_SPARSIFICATION_FCCM}.
                Their work features a load balancing-aware sparsification training scheme facilitating efficient parallelism.
                Their FPGA implementation of AlexNet achieved 12$\times$ throughput acceleration over an Intel Xeon CPU-based benchmark.
                Posewsky \emph{et al.} presented an FPGA implementation of high-throughput zero-skipping suiting fine-grained pruning~\cite{PRU_FPGA_POSEWSKY_ZERO_SKIPPING}.
                The authors proposed that, post-pruning, each non-zero weight be encoded as a two-element tuple $\left(w_i, z_i\right)$ containing weight value $w_i$ and number of preceding zeroes $z_i$, where $i$ is the weight's index.
                In this way, when a batch of input activations is buffered on-chip, the hardware will only fetch the weights pointed to by $z_i$, corresponding to non-zeroes only.
                Experiments with an unidentified model showed that their Xilinx Zynq XC7Z020 FPGA implementation surpassed the throughput of ARM Cortex-A9 and Intel Core~i7-5600U CPU equivalents, with $>$ 85\% energy savings.
				
				ESE's authors reported that, with pruning and retraining, more than 90\% of the parameters of an arbitrarily chosen LSTM trained on the TIMIT dataset could be pruned away without harming accuracy~\cite{R_LSTM_FXP_ESE}.
				Its authors proposed ``balance-aware" pruning to shape weight matrices into equal workloads for parallel compute units during retraining.
				On FPGAs, weight matrices are stored and computed in a compressed sparse column format to skip zeroes under this proposal.
				ESE demonstrated 3.0$\times$ throughput acceleration \emph{vs} an Nvidia Pascal~Titan~X GPU implementation.
                
                The authors of Eyeriss~\cite{PRU_CNN_EYERISS}, EIE~\cite{PRU_CNN_EIE}, Cnvlutin~\cite{PRU_ASIC_CNVLUTIN} and Laconic~\cite{PRU_CNN_LACONIC} sought to remove multiplications by zero-valued activations.
                The authors of Cnvlutin achieved this by computing only non-zero inputs and using an ``offset" buffer, alongside the input buffer, to store the indices of each input's corresponding weights after zero-skipping.
                A hardware controller fills the offset buffer on the fly such that it does not consume extra bandwidth.
                To further increase acceleration, Cnvlutin prunes near-zero outputs during inference in order to increase the sparsity of the next layer's input buffer.
                Experiments with several CNNs, including AlexNet, GoogleNet and VGG-19, showed 1.2--1.6$\times$ throughput increases over DaDianNao~\cite{FXP_CNN_DADIANNAO} without any loss in accuracy for ImageNet.
                While Cnvlutin incurred an area overhead of 4.5\% over DaDianNao, it beat it by 1.5$\times$ in terms of energy efficiency for an unnamed model.
                Eyeriss, EIE and Laconic's authors achieved benefits from pruning using similar strategies to those employed by Cnvlutin's.
                
                Unlike the previous proposals, which all prune parameters to achieve throughput speedups, the authors of Eyeriss and Minerva targetted power savings through the elimination of redundant off-chip memory fetches~\cite{PRU_CNN_EYERISS,PRU_CNN_MINERVA}.
                Experiments with Minerva showed that their 40~nm ASIC implementation achieved an 8.1$\times$ energy efficiency reduction---also for an unidentified model---compared with an ASIC baseline.
                
		\subsection{Weight Sharing}
		\label{sec:weight_sharing}

			\subsubsection{Algorithmic Development}

				Weight sharing groups parameters into buckets, reducing network size as well as enabling multiplications to be converted into cheaper table lookups.
				In HashedNets, a low-cost hash function is used to randomly group connection weights, the connections in each of which all share a single value~\cite{PRU_CNN_HASHEDNET}.
                These parameters are then trained to adjust to the weight sharing with standard backward propagation.
                Experiments with the MNIST dataset showed that HashedNets achieved a compression ratio of 64 with an around-0.7~pp accuracy improvement against a five-layer CNN baseline.
                The authors suggested that the accuracy rise could be attributed to the ``virtual" connections created that seemingly increased expressiveness.
				
				Ullrich \emph{et al.} performed retraining using soft weight sharing on pretrained networks in order to fine-tune the centroids used for parameter clustering~\cite{PRU_CNN_SOFT_WEIGHT_SHARING}.
                Soft weight sharing was originally proposed by Nowlan and Hinton, who modelled cluster centroids with a mixture of Gaussians~\cite{SOFT_WEIGHT_SHARING}.
                When retraining with this constraint, weights tend to concentrate very tightly around a number of cluster components, the centroids of which optimise to improve accuracy.
                Experiments showed 160$\times$ compression for MNIST on LeNet-5 with an accuracy loss of $\sim$0.1~pp.
				
				With Deep Compression, weight sharing is performed in several steps~\cite{PRU_CNN_DEEP_COMPRESSION}.
                A network is first pruned with iterative retraining~\cite{PRU_CNN_TRAIN_PRUNE_RETRAIN}, after which weights are quantised via $k$-means clustering.
                The quantised network is then retrained again to fine-tune the remaining connections and update the cluster centroids.
                Finally, the quantised weights are compressed with Huffman coding to save memory. 
                With $k$-means clustering, the spatial complexity of a size-$K$ weight matrix reduces from $\mathcal{O}{\left(K^2\right)}$ to $\mathcal{O}{\left(k\right)}$.
				Using their basket of approximation techniques, the authors of Deep Compression achieved $35\times$ overall compression for AlexNet with no drop in ImageNet accuracy.
                
                The proposals above only encode weights.
                Both LookNN~\cite{PRU_CNN_LOOKNN} and Quantised CNN~\cite{PRU_CNN_QUANTIZED_CNN} follow the ``product quantisation" algorithm~\cite{PRODUCT_QUANTISATION}, which encode both weights and activations.
				Rather than operating element-wise, this method does so on subvectors of weight matrices.
                Experiments with Quantised CNN revealed 19$\times$ AlexNet compression in return for 1.5~pp of ImageNet accuracy loss.

			\subsubsection{Hardware Implementation}
            
                During inference, weight sharing-based implementations require a large number of lookup operations, which can be performed significantly more efficiently on FPGAs than GPPs.
				Samragh \emph{et al.} implemented weight sharing on FPGAs~\cite{PRU_CNN_FCCM}.
				Here, $k$-means cluster centroids are determined with tunable parameters during retraining, eliminating almost all multiplications.
				An up-to 15$\times$ improvement in throughput and compression ratio of 9.0 were reported along with with sub-0.1~pp of accuracy losses for small DNN datasets such as MNIST and ISOLET on unidentified network models.
                
                The authors of PQ-CNN presented a hardware-software framework for compressing and accelerating CNNs on FPGAs using product quantisation~\cite{PRODUCT_QUANTISATION}, adopting a similar idea to that used in Quantised CNN~\cite{PRU_CNN_QUANTIZED_CNN,PRU_FPGA_PQ_CNN_FCCM}.
                Going further, the authors implemented an extra codebook to compress encoding parameters, increasing the compression of the original algorithm.
                During inference, since all possible multiplication outputs with every codeword are precomputed and stored on-chip, PQ-CNN sees dot products for both convolutions and fully connected layers converted into table lookups and accumulations.
                The authors' Amazon~F1 implementation achieved 4.6~kcl/s for the VGG16 model with a sub-0.5~pp drop in top-five accuracy for ImageNet.

		\subsection{Low-rank Factorisation}
		\label{sec:LRF}

			\subsubsection{Algorithmic Development}
			
			    Post-training low-rank factorisation of DNNs can achieve significant network compression and computation reductions for inference.
	            Denton \emph{et al.} analysed the effect of applying several decomposition methods---singular-value decomposition (SVD), canonical polyadic (CP) decomposition and biclustering approximation---on pretrained weight matrices~\cite{LRF_SVD_NIPS}.
                A biclustering approximation performs $k$-means clustering on rows and columns of weight matrices~\cite{PRODUCT_QUANTISATION}. 
                These methods were tested with a 15-layer CNN classifying the ImageNet dataset.
                Among them, SVD achieved the best performance: 13$\times$ compression of the first fully connected layer with 0.84~pp of top-one accuracy loss.
				Tai \emph{et al.} also performed network decomposition using SVD~\cite{LRF_SVD_WITH_OPTIMISER}.
				They achieved up to 5.0$\times$ compression and a 1.8$\times$ throughput speedup for ImageNet on AlexNet, reporting a top-five accuracy reduction below 0.5~pp.

                While post-training decomposition is simple and flexible, many works have shown that training after decomposition can recover compression losses.
				As suggested by Jaderberg \emph{et al.}, weight matrices can be decomposed into several low-rank matrices to enable workload and/or memory reductions~\cite{LRF_LINEAR_COMBINATIONS}.
                The authors proposed the factorisation of each of their four-dimensional layers into a sequence of two regular convolutional layers, each of three dimensions.
                Experiments with various nonstandard scene text character recognition datasets showed that this method achieved, on average, a 4.5$\times$ increase in throughput with around-1~pp falls in accuracy for some unidentified networks.
                This factorisation scheme inspired MobileNet, which uses one three-dimensional ``depthwise" and one two-dimensional ``splitwise" separable convolutional layers to approximate each original layer~\cite{MOBILENET}.
                Assume that a convolutional layer contains $K \times K \times M \times N$ values, where $K$, $M$ and $N$ are the size of the kernel and numbers of input and output channels, respectively.
                In MobileNet, this is factorised into a depthwise convolutional layer with $K \times K \times M \times 1$ values and a pointwise convolutional layer of size $1 \times 1 \times M \times N$.
                This method effectively reduces the complexity of forward propagation from $\mathcal{O}{\left(M D^2 K^2 N\right)}$ to $\mathcal{O}{\left(M D^2\left(K^2+N\right)\right)}$, where $D$ is the size of the input feature map.
                Experiments with ImageNet showed that MobileNet can achieve a 3.0~pp top-one accuracy improvement with 46$\times$ compression for AlexNet.
                
                Ba \emph{et al.} combined low-rank factorisation with knowledge distillation, where a deep and complex neural network is mimicked with a simpler, shallower one~\cite{KD_DO_NEED_DEEP}.
                More detail on knowledge distillation is given in Section~\ref{sec:KD}.
                The authors noticed that learning is very slow for the weight matrices of shallow networks.
                Since there are many highly correlated parameters, gradient descent converges slowly, with the majority of training time spent on matrix-vector multiplication.
                They suggested that forward and backward propagation could be sped up by approximating each large weight matrix as the product of two low-rank matrices. 
                Increases in convergence rate of the network mimicking and reductions in memory space complexity were observed.
                Lebedev \emph{et al.} presented a CP decomposition-based retraining method facilitating greater workload reductions, achieving a 4.5$\times$ throughput boost with $\sim$1~pp of top-five ImageNet accuracy loss for layer two of AlexNet~\cite{LRF_CP_ICLR}.

				Following the logic that learnt weight matrices tend to be structured and can be decomposed using low-rank factorisation, Denil \emph{et al.} suggested the storage of only parts of weight matrices, predicting the remainder using a second learning model~\cite{LRF_PREDICTING_PARAMETERS}.
				They reported that, in the best case---with small-scale datasets---more than 95\% of weights can be predicted without accuracy loss.
				The networks used therein were nonstandard.
				
				Rather than compressing layers individually, Kim \emph{et al.} performed ``one-shot" whole-network compression using Tucker decomposition.
				Here, the post-decomposision ranks of all layers are determined all at once through global Bayesian matrix factorisation.
				Experiments showed that, while this method requires at least 10 retraining epochs for accuracy recovery, the inference of AlexNet on an Nvidia Titan~X GPU achieved 1.8$\times$ speedup, with 1.7~pp of top-five ImageNet accuracy loss, against an FP32 baseline on the same platform.

			\subsubsection{Hardware Implementation}
            
                Low-rank factorisation methods produce structured DNN models which can inference efficiently on GPPs with dense matrix-vector BLAS.
                Li \emph{et al.} presented a CNN compression framework combining coarse-grained pruning using sparsification with low-rank factorisation~\cite{LRF_CNN_GPU_ASP_DAC}.
                Similar to the idea proposed by Jaderberg \emph{et al.}~\cite{LRF_LINEAR_COMBINATIONS}, the authors represented filters as linear combinations of lower-rank basis filters.
                GPU experiments with AlexNet, GoogleNet and VGGNet-A revealed about-2$\times$ throughput speedups without accuracy loss for ImageNet.

                Custom hardware implementations, however, can achieve comparable performance with lower power envelopes.
                Rizakis \emph{et al.} implemented SVD-factorised gates for LSTMs~\cite{LRF_LSTM_SVD_BASED}. 
                In their proposal, SVD is performed on the weights of the four LSTM gates independently. 
                For each gate, the weights associated with both the current input and previous output are concatenated together to form a large weight matrix, which is then SVD-factorised. 
                Pruning is also performed by retaining only rows with a majority of non-zeroes in each weight matrix.
                The authors implemented their design on an FPGA platform, achieving a 6.5$\times$ throughput increase for an arbitrarily chosen LSTM compared with an uncompressed FPGA-based LSTM baseline.

		\subsection{Structured Matrices}
		\label{sec:SM}

			\subsubsection{Algorithmic Development}

				A weight matrix can be represented as a structure of repeated patterns such that it can be expressed with fewer parameters.
				The use of circulant matrices for representing weight matrices $\boldsymbol{W}$ in CNNs and RNNs has proven to be a very popular proposal~\cite{SM_CNN_CIRCULANT_PROJECTIONS,SM_CNN_CIRCULANT_PROJECTIONS_0,SM_CNN_TOEPLITZ_LIKE,SM_LSTM_HYBRID_CIRCULAR,SM_LSTM_HYBRID_TOEPLITS_LIKE}.
				A circulant matrix $\boldsymbol{W}_\text{circ}$ of size $K$ is square, with all rows being a shifted version of the first, $\boldsymbol{w}_{0*}$, thereby reducing spatial complexity from $\mathcal{O}{\left(K^2\right)}$ to $\mathcal{O}{\left(K\right)}$.
				It is constructed as such:
                \begin{equation*}
                    \boldsymbol{W}_\text{circ} =
                    \begin{pmatrix}
                        w_{0}   & w_{K-1}   & \cdots    & w_{1}     \\
                        w_{1}   & w_{0}     & \cdots    & w_{2}     \\
                        \vdots  & \vdots    & \ddots    & \vdots    \\
                        w_{K-1} & w_{K-2}   & \cdots    & w_{0}
                    \end{pmatrix}.
                \end{equation*}
				The multiplication of $\boldsymbol{W}_\text{circ}$ by input vector $\boldsymbol{x}$ can thus be computed using a fast Fourier transform (FFT) of the first row of $\boldsymbol{W}_\text{circ}$, reducing inference time complexity from $\mathcal{O}{\left(K^2\right)}$ to $\mathcal{O}{\left(K \log K\right)}$, as
                \begin{equation*}
                    \boldsymbol{W}_\text{circ} \boldsymbol{x} = \text{ifft}{\left(\text{fft}{\left({\boldsymbol{w}_\text{circ}}_{0*}\right)} \odot \text{fft}{\left(\boldsymbol{x}\right)}\right)}.
                \end{equation*}
                
				While the circulant matrix method has shown outstanding memory and computational complexity reductions, its application also introduces accuracy degradation.
				For example, the AlexNet implementation of a circulant matrix-based framework, CirCNN, achieved compression of 40$\times$ with 16-bit fixed-point quantisation, yet its use also resulted in 2.2~pp of ImageNet accuracy degradation against an FP32 baseline~\cite{SM_CNN_FPGA_CIRCNN}.
				An alternative transformation, the Adaptive Fastfood transform (AFT), achieved a compression ratio of 3.7, but only about 0.1~pp of accuracy loss with ImageNet, for AlexNet~\cite{SM_CNN_ADAPTIVE_FASTFOOD_TRANSFORM}.
				In an AFT, a weight matrix $\boldsymbol{W}$ is approximated as
                \begin{equation*}
                    \boldsymbol{W}_\text{AFT} = \boldsymbol{SHG\Pi HB},
                \end{equation*} 
                in which $\boldsymbol{S}$, $\boldsymbol{G}$ and $\boldsymbol{B}$ are trainable diagonal matrices, $\boldsymbol{H}$ a Hadamard matrix and $\boldsymbol{\Pi} \in \left\{0,1\right\}^{K \times K}$ a trainable permutation matrix.
				This and the circulant method have equal complexities.

                For both of the aforementioned structures, generality is not guaranteed when dealing with classification tasks of varying scales. 
                Sindhwani \emph{et al.} proposed structured transformations characterised by the notion of a \emph{displacement rank} parameter~\cite{SM_CNN_TOEPLITZ_LIKE}. 
                With different displacement ranks, a continuum is exposed from fully structured to completely unstructured. 
                With displacement rank less than or equal to two, weight matrices become Toeplitz matrices, which have the form
                \begin{equation*}
                    \label{eqn:toeplitz}
                    \boldsymbol{W}_\text{Top} =
                    \begin{pmatrix}
                        w_{0}   & w_{-1}    & \cdots    & w_{-\left(K-1\right)} \\
                        w_{1}   & w_{0}     & \cdots    & w_{-\left(K-2\right)} \\
                        \vdots  & \vdots    & \ddots    & \vdots                \\
                        w_{K-1} & w_{K-2}   & \cdots    & w_{0}
                    \end{pmatrix}.
                \end{equation*}
                Different to a circulant matrix, a Toeplitz matrix $\boldsymbol{W}_\text{Top}$ of size $K$ has element values $w_{-\left(K-1\right)}$ to $w_{K-1}$.
                Matrix-vector multiplications can still take advantage of FFTs by embedding Toeplitz matrices into larger circulant matrices, as in
                \begin{equation*}
                    \boldsymbol{W}_\text{circ,\ Top} =
                    \begin{pmatrix}[cccc|cccc]
                        w_{0}					& w_{-1}	            & \cdots	& w_{-\left(K-1\right)} & 0						& w_{K-1}	            & \cdots	& w_{1}					        \\
                        w_{1}					& w_{0}		            & \cdots	& w_{-\left(K-2\right)} & w_{-\left(K-1\right)}	& 0                     & \cdots	& w_{2}					        \\
                        \vdots					& \vdots	            & \ddots	& \vdots				& \vdots				& \vdots                & \ddots    & \vdots				    \\
                        w_{K-1}					& w_{K-2}	            & \cdots	& w_{0}					& w_{-1}				& w_{-2}	            & \cdots	& 0						        \\
                        \hline
                        0						& w_{K-1}	            & \cdots	& w_{1}					& w_{0}					& w_{-1}	            & \cdots	& w_{-\left(K-1\right)}	\\
                        w_{-\left(K-1\right)}	& 0                     & \cdots	& w_{2}					& w_{1}					& w_{0}		            & \cdots    & w_{-\left(K-2\right)}   \\
                        \vdots					& \vdots                & \ddots    & \vdots				& w_{2}					& \vdots                & \ddots    & \vdots			    	\\
                        w_{-1}					& w_{-2}	            & \cdots	& 0						& w_{K-1}				& w_{K-2}	            & \cdots	& w_{0}
                    \end{pmatrix},
                \end{equation*}
                and exploiting the relationship
                \begin{equation*}
                    \label{eqn:toeplitz_mvmult}
                    \boldsymbol{W}_\text{Top} \boldsymbol{x} =
                    \begin{pmatrix}
                        \boldsymbol{I}_K    & \boldsymbol{0}_{K \times K}
                    \end{pmatrix}
                    \boldsymbol{W}_\text{circ,\ Top}
                    \begin{pmatrix}
                        \boldsymbol{x}              \\
                        \boldsymbol{0}_{K \times K}
                    \end{pmatrix},
                \end{equation*}
                wherein $\boldsymbol{I}$ and $\boldsymbol{0}$ are identity and zero matrices, respectively~\cite{TOEPLITZ_CIRCULAR_REVIEW}.

                A family of Toeplitz-like matrices can be generated by increasing rank beyond two. 
                With rank $K$, a matrix becomes unstructured and uncompressed.
                Lu \emph{et al.} applied Toeplitz-like matrices in LSTMs, with weight matrices of gates trained in Toeplitz-like structures of various ranks~\cite{SM_LSTM_HYBRID_TOEPLITS_LIKE}. 
                The authors compressed the first two layers of an unidentified five-layer LSTM into structures of rank five, achieving a compression ratio of around 1.7 with $\sim$0.3~pp loss in speech recognition accuracy for a dataset consisting of some 300 hours of English utterances.
                
                While the authors of the works mentioned above reported that the use of circulant matrix-based methods resulted in the incursion of at-least 2~pp accuracy drops for large-scale CNN image classifications, their accuracies for RNN tasks are significantly superior.
                Wang \emph{et al.} implemented circulant matrices together with non-linear function approximation and quantisation for LSTMs~\cite{SM_LSTM_HYBRID_CIRCULAR}.
                Language modelling and speech recognition were performed by their 90~nm ASIC, achieving more than 20$\times$ compression with a 2.8~pp loss in accuracy for classification of the AN4 speech database.
				
				C-LSTM features block-circulant matrices, each of which consists of circulant submatrices of arbitrary size~\cite{SM_CNN_FPGA_C-LSTM}.
                Tunable block size facilitates a tradeoff between storage requirements and accuracy.
                Experiments with the Google LSTM architecture revealed a linear relationship between block size and compression ratio, as well as a clear tradeoff between block size and TIMIT phone error rate (PER) increase.
                For an LSTM model with block size of eight on the TIMIT dataset, C-LSTM exhibited 7.6$\times$ compression and a 2.6$\times$ workload reduction while incurring a 0.32~pp PER rise.

			\subsubsection{Hardware Implementation}
            
                Convolutions on GPPs are normally performed after unrolling, flattening four-dimensional inputs and kernels into two-dimensional matrices.
                This converts four-dimensional tensor operations into two-dimensional matrix multiplications, trading off memory use for performance.
                For block-circulant matrix methods, since each two-dimensional slice of a kernel is circulant, the two-dimensional unrolled version of that kernel is also block-circulant.
                Time complexity reductions from the FFT-based method for block-circulant matrix inference are hence achievable for DNN inference performed on both GPPs and in custom hardware.
                Despite this, custom hardware implementations still excel in terms of energy efficiency~\cite{SM_CNN_FPGA_CIRCNN}.
                
				Combined with 16-bit fixed-point quantisation, FPGA-based C-LSTM~\cite{SM_CNN_FPGA_C-LSTM} achieved a 10$\times$ throughput speedup and 34$\times$ energy efficiency improvement over ESE for the Google LSTM, the prior state of the art.
                Ding \emph{et al.} presented implementations using similar methods for CNNs and RNNs on both FPGAs and ASICs~\cite{SM_FPGA_ASIC_BLOCK_CIRCULANT}.
                Using Intel Cyclone~V FPGAs, the authors achieved at-least 150$\times$ and 72$\times$ improvements in performance and energy efficiency, respectively, over IBM TrueNorth implementations~\cite{FXP_CNN_TRUENORTH} of some unidentified networks.
                For Xilinx Kintex~UltraScale FPGA LSTM implementation, the proposed architecture achieved up-to 21$\times$ and 34$\times$ improvements in throughput and energy efficiency, respectively, over ESE for the Google LSTM~\cite{R_LSTM_FXP_ESE}.
                The authors also experimented with a LeNet-5 ASIC implementation, achieving a throughput of 1.1~Mcl/s and energy efficiency of 8.1~Mcl/J.
                Wang \emph{et al.} presented a circulant matrix-based LSTM inference implementation in 90~nm ASIC technology~\cite{SM_LSTM_HYBRID_CIRCULAR}.
                They adopted a hybrid strategy in their work, also exploiting fixed-point quantisation and activation function approximation.
                With a 520~KiB on-chip memory allocation, the authors were able to process a $512\times 512$ compressed layer of an arbitrarily chosen LSTM in 1.7~{\textmu}s: equivalent to 580~kcl/s.
                
				Fox \emph{et al.} implemented AFTs for accelerating matrix-vector multiplication on FPGAs~\cite{SM_FPGA_FASTFOOD}.
				Although their work was not presented in the context of DNN inference, its results on matrix-vector multiplication are still relevant.
				The authors concluded that the AFT's small memory complexity allows for the processing of input matrices some 1000$\times$ larger than previous online kernel methods with the same area occupancy.

		\subsection{Knowledge Distillation}
		\label{sec:KD}

			\subsubsection{Algorithmic Development}

				Knowledge distillation mimics large, complex DNNs using simpler and shallower networks in order to achieve network compression.
                In one of the earliest works in this field, Hinton \emph{et al.} suggested that knowledge could be distilled from an ensemble of models (teachers) into a simple model (student) by training the student model with outputs from the teachers~\cite{KD_HINTON}.
                Ba \emph{et al.} provided empirical evidence showing that, in simple machine learning tasks, a student network can mimic teacher networks with comparable performance~\cite{KD_DO_NEED_DEEP}. 
				In FITNet, intermediate outputs of these teacher models are used as ``hints" for training the student model to improve its accuracy~\cite{KD_FITNETS}.
				Experiments with the CIFAR-10 dataset showed that a FITNet trained from an unidentified 9M-parameter teacher CNN could achieve 10$\times$ compression and a 1.4~pp accuracy improvement \emph{vs} the teacher network.
                The authors explained that a reduction in network complexity from teacher to student led to less overfitting, causing the accuracy increase.
                
                Chen \emph{et al.} proposed various optimisations for improving the performance of network mimicking~\cite{KD_WEIGHTED_CROSS_ENTROPY}. 
                Unlike carefully selected image classification datasets with uniform class distributions, object detection problems need to deal with dominant background classes.
                Class-weighted cross-entropy can be introduced to handle such scenarios, wherein a background class is assigned an appropriate scaling factor to correct for class imbalances.
                When teacher overfitting occurs, hints from a teacher network may ``mislead" a student into even more severe overfitting.
                In an effort to avoid this, Chen \emph{et al.} used their teacher network's original regression curve as an upper bound for student network training.
                Experiments with the PASCAL, KITTI and COCO datasets showed that these optimisations improved accuracies by 3--5~pp.
                
                Alemdar \emph{et al.} introduced a framework for knowledge distillation in which ternary student networks were trained from a ternarised teacher~\cite{KD_TNN_FPGA}.
                During ternarisation, two thresholds for each weight index $i$, $\left\{\delta_i^-, \delta_i^+\right\}$, are used to differentiate quantisation levels $\left\{w_i^-, w_i^+\right\}$ from zero.
                The authors suggested that the use of well selected thresholds should result in outputs from the student network perfectly matching those of the teacher network.
                A greedy method was proposed to search for thresholds by minimising the difference between the probability distribution functions of layer-wise outputs from the student and teacher networks.
                Experiments with MNIST, CIFAR-10 and SVHN showed that this work achieved higher accuracies than IBM TrueNorth classifying the same datasets on VGG-like models with ternary data~\cite{FXP_CNN_TRUENORTH}.

			\subsubsection{Hardware Implementation}

                Knowledge distillation essentially converts deep DNNs into shallow ones, which, from a hardware perspective, allows the replacement of deep, sequential processing with parallel, distributed processing.
                This structural conversion greatly facilitates the acceleration of DNN training and inference using GPPs.
                Ba \emph{et al.} even observed that some shallow, mimicked models reached similar accuracies for TIMIT to deep models about 8$\times$ more quickly~\cite{KD_DO_NEED_DEEP}.
                
                While some acceleration can be achieved with knowledge distillation on GPPs, further benefit can be realised given the flexibility of custom hardware by taking advantage of additional approximation.
				Alemdar \emph{et al.} presented a hardware mapping framework in which student networks trained through network mimicking are translated into hardware descriptions for FPGA or ASIC implementation~\cite{KD_TNN_FPGA}.
				Their Xilinx Virtex-7 FPGA prototype achieved an over-30$\times$ throughput improvement and comparable energy efficiency \emph{vs} IBM TrueNorth~\cite{FXP_CNN_TRUENORTH} executing VGG-like models.
                A 28nm~ASIC implementation was also presented and compared against a state-of-the-art ASIC implementation, EIE~\cite{PRU_CNN_EIE}.
                While their ASIC did not beat EIE in terms of throughput, it did achieve 1.2$\times$ energy efficiency and 2.9$\times$ area occupancy improvements for an unidentified network model.

	\section{Input-dependent Computation}
	\label{sec:RuntimeMethods}
            
		\subsection{Algorithmic Development}

			Different regions of a DNN's input data may have differing levels of contribution to its output.
			Input-dependent computation exploits this observation by assigning compute proportionally to the input data's relative importance.
			Stochastic Times Smooth units mask CNN input frames with a pretrained binary decision matrix to facilitate conditional computation, which was shown to give 10$\times$ compression for a nonstandard CNN classifying the MNIST dataset with a 0.2~pp accuracy improvement~\cite{DR_CNN_STS}.
			Karpathy \emph{et al.} allocated more resources to the centres of CNN input frames for improved video classification accuracy~\cite{DR_CNN_VIDEO}.
            Their implementation consists of two CNNs in parallel, with a ``context stream" CNN processing entire frames and a ``fovea stream" CNN processing only the centre of each. 
            The authors reported a 65\% prediction accuracy on the UCF-101 video prediction dataset: state-of-the-art performance at the time of publication.
            
            Low-rank approximation has not just been studied in the parameter space; it has been used for input compression as well.
            In Deep3~\cite{LRF_CNN_DEEP3} and DeLight~\cite{LRF_CNN_DELIGHT}, input data matrices are factorised into lower-rank matrices using an ``embedding matrix."
            These are iteratively updated to reduce the Frobenius norm of factorisation errors.
            Experiments with Deep3 on GPUs on various deep learning tasks, including audio classification, demonstrated up-to 11$\times$ inference speedups compared to a TensorFlow baseline running the same models~\cite{LRF_CNN_DEEP3}.

			While the aforementioned \emph{static} computation allocation schemes can achieve significant resource savings and/or accuracy improvements, recent research, such as Dynamic Capacity Networks, has introduced \emph{dynamic} input-dependent allocation, guided at runtime by additional pretrained subnetworks~\cite{DR_CNN_DCN}.
			In Bengio \emph{et al.}~\cite{DR_CNN_REINFORCEMENT} and Liu \emph{et al.}'s~\cite{DR_CNN_DDNN} proposals, and Runtime Neural Pruning (RNP)~\cite{DR_CNN_RUNTIME_PRUNING}, partial execution of DNNs is performed using pretrained Markov decision process reinforcement learning.
			RNP was shown to achieve a 10$\times$ workload reduction and 5.9$\times$ latency reduction for VGG16 with ImageNet in return for a 4.9~pp drop in top-five accuracy.
			Runtime methods achieve superior accuracy to their static counterparts at the expense of an extra network.
            
			The works discussed above all targetted CNNs, for which computation is dependent upon the spatial features of their inputs.
			The authors of DeltaRNN, on the other hand, reduced RNN workload based on inputs' temporal behaviour~\cite{DR_GRU_FPGA_DeltaRNN}.
			DeltaRNN updates the output of an RNN only when its input changes by more than some threshold.
            They reported 9.8$\times$ throughput and 130$\times$ efficiency improvements for an arbitrarily chosen network, with a 1.5~pp accuracy drop, against their baseline classifying the TIDIGITs dataset.

		\subsection{Hardware Implementation}
        
			Since input-dependent computation involves frequent dynamic branching during inference, these implementations are not likely to pipeline efficiently, especially for deep CNNs.
            Hence, for CNN implementations exploiting this method, throughput is not their greatest advantage.
            They are instead focussed more on latency-critical applications, which generally do not require high throughput.
            Custom hardware, unlike GPPs, allows for specially designed dynamic branching mechanisms which can inference fine-grained, irregular data patterns more efficiently.
            
            The authors of CascasdeCNN presented the input-dependent computation of CNN inference on FPGAs~\cite{DR_CNN_FPGA_CASCADECNN}.
            Similar to Dynamic Capacity Networks~\cite{DR_CNN_DCN}, CascadeCNN features a high-precision subnetwork in addition to a low-precision main network.
            The former is activated when there is a potential misclassification in the latter, \emph{i.e.} when the confidence of the main network's best guess is low.
            Experiments showed that CascadeCNN achieved latency reductions of up to 55\% for VGG-16 and 48\% for AlexNet over the baseline design for the same resource budget and accuracy for ImageNet.
            The FPGA implementation of DeltaRNN on an LSTM requiring 5.6~Gop/cl demonstrated reduced off-chip memory bandwidth, achieving a throughput of 220~cl/s and an energy efficiency of 29~cl/J: state-of-the-art performance for RNN inference at the time~\cite{DR_GRU_FPGA_DeltaRNN}.

	\section{Activation Function Approximation}
	\label{sec:Other}
	
	   \subsection{Algorithmic Development} 

    		With non-linear functions such as sigmoid and tanh, computations including exponentiation and division are expensive to perform.
    		Piecewise Linear Approximation of Non-linear Functions (PLAN) simplifies such functions into serieses of table lookups~\cite{PLAN}.
    		In turn, this leads to the quantisation of activations in subsequent layers, reducing both memory requirements and numbers of arithmetic operations to perform.
            PLAN appears more often in RNN implementations than CNNs; mainstream CNNs use ReLU as the activation function, which can be cheaply implemented by comparing outputs with zero.
            In RNNs, on the other hand, empirical analysis suggests that sigmoid and tanh provide better performance, whereas ReLU not only performs poorly but also diverges frequently, partly because it is positively unbounded~\cite{ACTIVATION_FUNCTIONS}.
            
      \subsection{Hardware Implementation}  
    
    		PLAN can be efficiently implemented in custom hardware.
            Guan \emph{et al.} implemented PLAN within an FPGA-based inference framework for unidentified LSTMs, and their experiments showed that its use introduced only 0.63~pp of TIMIT accuracy degradation~\cite{PLAN_LSTM_FPGA}.
    		Li \emph{et al.}~\cite{FXP_RNN_MIXED} and the authors of ESE~\cite{R_LSTM_FXP_ESE}, C-LSTM~\cite{SM_CNN_FPGA_C-LSTM} and DeltaRNN~\cite{DR_GRU_FPGA_DeltaRNN} implemented arbitrarily chosen RNNs on FPGAs with PLAN, reporting increases in throughput with negligible accuracy losses for the same dataset.

	\section{Tradeoffs and Current Trends}
	\label{sec:WDiscussion}
    
        Thus far, we have detailed DNN approximation techniques and their hardware implementations on different platforms.
        Performance evaluations were made against benchmarks and baseline implementations of their authors' choosing, which are inconsistent and often not particularly useful when attempting to perform comparisons.
        We now quantitatively evaluate the hardware and software performance of those works using common DNN models and datasets as benchmarks.
        By doing so, we analyse the compression-accuracy tradeoffs of the approximation techniques and their design-space exploration for custom hardware, from which we explain current research trends.

		\subsection{Compression \emph{vs} Accuracy}
        \label{sec:com_acc}

\begin{figure*}
				\centering
				\begin{subfigure}[t]{0.49\columnwidth}
					\centering
					\begin{tikzpicture}

	\begin{axis}[
		width=\textwidth,
		height=\textwidth,
		xlabel={Compression \emph{vs} baseline},
		ylabel={Top-one error rate (\%)},
		xmode=log,
		log basis x=2,
		log ticks with fixed point,
		every node near coord/.append style={font=\scriptsize, color=black, anchor=\myanchor},
		every axis plot/.append style={scatter, only marks, nodes near coords, point meta=explicit symbolic, visualization depends on={value \thisrow{anchor} \as \myanchor}, thick}
	]

		\addplot[
			mark=triangle,
			color=red,
			discard if not={category}{base}
		]
		table[x=compRatio, y=errorRate, meta=method]{data/algo_quant.txt};
		\label{plt:algo_quant_base}
		
		\addplot[
			mark=+,
			color=green,
			discard if not={category}{fixed32}
		]
		table[x=compRatio, y=errorRate, meta=method]{data/algo_quant.txt};
		\label{plt:algo_quant_fixed32}
		
		\addplot[
			mark=diamond,
			color=blue,
			discard if not={category}{exp}
		]
		table[x=compRatio, y=errorRate, meta=method]{data/algo_quant.txt};
		\label{plt:algo_quant_exp}
		
		\addplot[
			mark=*,
			color=cyan,
			discard if not={category}{tern}
		]
		table[x=compRatio, y=errorRate, meta=method]{data/algo_quant.txt};
		\label{plt:algo_quant_tern}
		
		\addplot[
			mark=x,
			color=magenta,
			discard if not={category}{bin}
		]
		table[x=compRatio, y=errorRate, meta=method]{data/algo_quant.txt};
		\label{plt:algo_quant_bin}
	
	\end{axis}

\end{tikzpicture}
	
					\caption{Quantisation methods: baseline~(\ref{plt:algo_quant_base}), eight-bit fixed point~(\ref{plt:algo_quant_fixed32}), logarithmic~(\ref{plt:algo_quant_exp}), ternary~(\ref{plt:algo_quant_tern}) and binary~(\ref{plt:algo_quant_bin}).}
					\label{ALGO_QUANT}
				\end{subfigure}%
				\hfill%
				\begin{subfigure}[t]{0.49\columnwidth}
					\centering
					\begin{tikzpicture}

	\begin{axis}[
		width=\textwidth,
		height=\textwidth,
		xlabel={Compression \emph{vs} baseline},
		ylabel={Top-one error rate (\%)},
		xmode=log,
		log basis x=2,
		log ticks with fixed point,
		every node near coord/.append style={font=\scriptsize, color=black, anchor=\myanchor},
		every axis plot/.append style={scatter, only marks, nodes near coords, point meta=explicit symbolic, visualization depends on={value \thisrow{anchor} \as \myanchor}, thick}
	]

		\addplot[
			mark=triangle,
			color=red,
			discard if not={category}{base}
		]
		table[x=compRatio, y=errorRate, meta=method]{data/algo_prune.txt};
		\label{plt:algo_prune_base}
		
		\addplot[
			mark=+,
			color=green,
			discard if not={category}{hybrid}
		]
		table[x=compRatio, y=errorRate, meta=method]{data/algo_prune.txt};
		\label{plt:algo_prune_hybrid}
		
		\addplot[
			mark=diamond,
			color=blue,
			discard if not={category}{share}
		]
		table[x=compRatio, y=errorRate, meta=method]{data/algo_prune.txt};
		\label{plt:algo_prune_share}
		
		\addplot[
			mark=*,
			color=cyan,
			discard if not={category}{prune}
		]
		table[x=compRatio, y=errorRate, meta=method]{data/algo_prune.txt};
		\label{plt:algo_prune_prune}
		
		\addplot[
			mark=x,
			color=magenta,
			discard if not={category}{struct}
		]
		table[x=compRatio, y=errorRate, meta=method]{data/algo_prune.txt};
		\label{plt:algo_prune_struct}
		
		\addplot[
			mark=o,
			color=purple,
			discard if not={category}{fact}
		]
		table[x=compRatio, y=errorRate, meta=method]{data/algo_prune.txt};
		\label{plt:algo_prune_fact}
	
	\end{axis}

\end{tikzpicture}
	
					\caption{Weight-reduction methods: baseline~(\ref{plt:algo_prune_base}), hybrid~(\ref{plt:algo_prune_hybrid}), weight sharing~(\ref{plt:algo_prune_share}), pruning~(\ref{plt:algo_prune_prune}), structured matrix~(\ref{plt:algo_prune_struct}) and factorisation~(\ref{plt:algo_prune_fact}).}
					\label{ALGO_PRUNE}
				\end{subfigure}
				\caption{Comparison of reported top-one error rates for implementations of AlexNet classifying ImageNet.}
				\label{fig:ALGO_QUANT_PRUNE}
			\end{figure*}
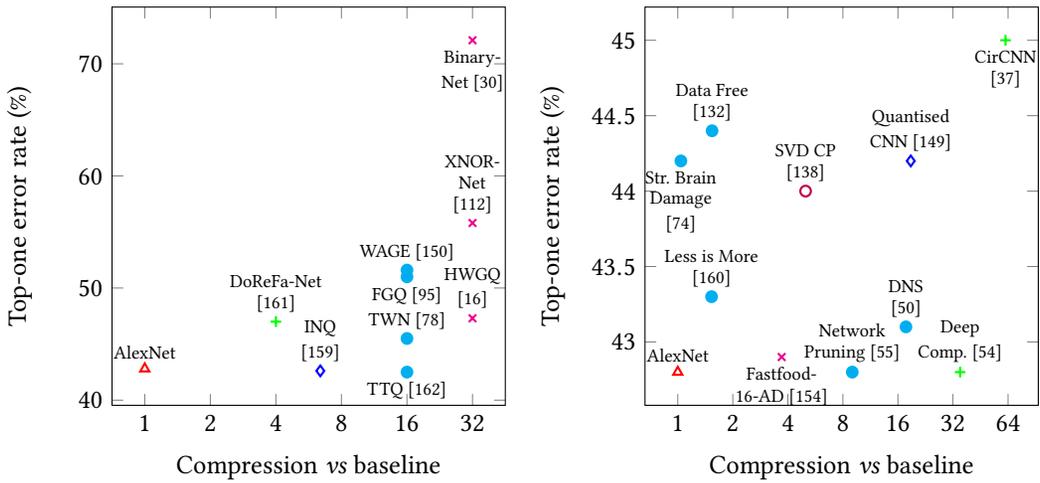

			Fig.~\ref{ALGO_QUANT} compares the compression-accuracy behaviour of key quantisation methods introduced in Section~\ref{sec:Quantisation} for ImageNet on AlexNet, indicating a clear relationship between precision and error rate.
			Among the methods, binary networks exhibit greater accuracy degradations ($\geq$~4.5~pp) than the remainder ($<$~3.0~pp), while also achieving the greatest compression ratios: 32 \emph{vs} an FP32 baseline.
			
			The parameters of trained DNNs usually have Gaussian-like distributions, wherein the majority of data have near-zero values.
			For this reason, binary networks exhibit high quantisation error for values with small magnitudes because they are unable to represent zeroes.
			Compared to binarisation, ternarisation generally results in better accuracy, with compression ratios of 16.
			Among all methods compared, TTQ has the highest accuracy at a reasonably high compression ratio, suggesting that the ability to represent zeroes has significant implications for network performance~\cite{TNN_CNN_TTQ}.
			INQ reached a similar level of accuracy to TTQ, but with a lower compression ratio (6.4)~\cite{EXP_CNN_INCR}.
			The accuracy of INQ is higher than fixed-point-quantised networks with similar precisions, supporting the conclusion by Lai \emph{et al.} that it is weights' representation range, rather than precision, that is crucial to the preservation of accuracy~\cite{NR_CNN_FXP_ACTIVATIONS}.

			Fig.~\ref{ALGO_PRUNE} facilitates comparison of the compression-accuracy tradeoffs, also for ImageNet on AlexNet, of the key weight-reduction methods introduced in Section~\ref{sec:weight_reduction}.
			It shows that the reported compression ratios for weight-sharing methods, such as Deep Compression~\cite{PRU_CNN_DEEP_COMPRESSION} and Quantised CNN~\cite{PRU_CNN_QUANTIZED_CNN}, and structured matrices, \emph{e.g.} CirCNN~\cite{SM_CNN_FPGA_CIRCNN}, are higher than the alternatives.
			This observation supports the theoretical analysis in Sections~\ref{sec:weight_sharing} and \ref{sec:SM} that these methods have good memory complexity reduction capabilities.
            
			Structured matrix methods induce significant accuracy degradation in CNNs~\cite{SM_CNN_FPGA_CIRCNN}, but not so much in LSTMs~\cite{SM_CNN_FPGA_CIRCNN,SM_CNN_FPGA_C-LSTM}.
			This phenomenon is not yet well understood.
            
            Pruning-based methods also lead to the obtainment of good accuracies at high compression ratios.
            Among them, fine-grained methods (DNS~\cite{PRU_CNN_DNS} and Network Pruning~\cite{PRU_CNN_TRAIN_PRUNE_RETRAIN}) show more promising tradeoffs than coarse-grained alternatives (Structured Brain Damage~\cite{PRU_CNN_GROUP_WISE_BRAIN_DAMAGE} and Less is More~\cite{PRU_CNN_LESS_IS_MORE}).
            This suggests that higher pruning granularities, despite inducing significant irregularity, possess greater potential for network compression and memory transfer reductions.
            
			Deep Compression exhibited both outstanding accuracy and compression~\cite{PRU_CNN_DEEP_COMPRESSION}.
			As a hybrid strategy, multiple quantisation methods work together to provide high compression.

			We can conclude that (re)training has proven to be effective in compensating for accuracy losses incurred due to approximation~\cite{PRU_LSTM_NMT_TRAIN_PRUNE_RETRAIN,PRU_CNN_TRAIN_PRUNE_RETRAIN}.
			The authors of methods exploiting binarisation, ternarisation, structured matrices, low-rank factorisation and knowledge distillation trained their networks from scratch, while the remaining methods---apart from Data Free~\cite{PRU_CNN_DATA_FREE}---use post-approximation retraining.
			Although Data Free featured pruning of similar neurons without the employment of retraining, it was used for all of the implementations in Fig.~\ref{fig:ALGO_QUANT_PRUNE}, suggesting that retraining has become a standard accuracy-recovery approach in state-of-the-art proposals.

		\subsection{Design-space Exploration}
        \label{sec:hardware_dse}

			Table~\ref{tab:THROUGHPUT_CHECKLIST} shows how each approximation method contributes to DNN inference acceleration in custom hardware.
            Increases in parallelism and reductions in model memory use increase compute bounds and arithmetic intensities, respectively, which, in turn, increase throughput.
			
			\begin{table}
			    \centering
			    \caption{How each approximation method contributes to DNN inference acceleration in custom hardware.}
			    \begin{footnotesize}
					\begin{tabular}{ccccc}
						\toprule
                                                                                        &                                   		& \shortstack{Cheaper arithmetic\\operations}  & \shortstack{Memory\\reduction}    & \shortstack{Workload\\reduction}											\\
                        \midrule
                        \multirow{4}{*}{\rotatebox{90}{\shortstack{Quant-\\isation}}}   & Fixed-point representation        		& \yup                                          & \yup                             & \nope																		                                                                        \\
                                                                                        & Binarisation and ternarisation    		& \yup                                          & \yup                             & \nope																		        \\
                                                                                        & \multirow{2}{*}{Logarithmic quantisation}	& \multirow{2}{*}{\yup}                         & \multirow{2}{*}{\yup}            & \multirow{2}{*}{\shortstack{\yup\\(if shift lengths are constant)}}		    \\
																																																																					 \\
                        \midrule
                        \multirow{6}{*}{\rotatebox{90}{\shortstack{Weight\\reduction}}} & Pruning                           		& \nope                                         & \yup                             & \yup																		                                                                        \\
                                                                                        & \multirow{2}{*}{Weight sharing}			& \multirow{2}{*}{\nope}                        & \multirow{2}{*}{\yup}            & \multirow{2}{*}{\shortstack{\yup\\(if multiplications are precomputed)}}    \\
																																																																					 \\
                                                                                        & Low-rank factorisation					& \nope                                         & \yup                             & \yup																		        \\
                                                                                        & Structured matrices						& \nope                                         & \yup                             & \yup																		        \\
                                                                                        & Knowledge distillation					& \nope                                         & \yup                             & \yup																		        \\
                        \midrule
                                                                                        & Input-dependent computation				& \nope                                         & \nope                            & \yup																		        \\
                        \midrule
                                                                                        & Activation function approximation			& \nope                                         & \nope                            & \yup																		        \\
                        \midrule
                                                                                        & Hybrid strategies							& \yup	                                        & \yup                             & \yup																		        \\
						\bottomrule
					\end{tabular}
					\label{tab:THROUGHPUT_CHECKLIST}
				\end{footnotesize}
			\end{table}
			
            Quantisation-based methods allow for increased parallelism through the use of cheaper arithmetic units.
            They also facilitate memory transfer reductions.
            With extremely low-precision quantisation, it becomes feasible to fix parameters in hardware such that weights do not need to be stored in, or fetched from, off-chip memory.
            Weight-reduction methods reduce numbers of parameters, saving memory while simultaneously decreasing workload.
            Weight sharing is slightly different from the other weight-reduction methods because it does not necessarily cause a reduction in workload.
            The number of operations to be performed per classification can be reduced if results are precomputed and stored on-chip, such as in PQ-CNN, however~\cite{PRU_FPGA_PQ_CNN_FCCM}.
            Unlike weight-reduction methods, input-dependent methods reduce workload without decreasing memory occupancy.
            Through precomputation, activation function approximation only reduces workload.
            Hybrid strategies have been commonly adopted recently; these can benefit from all three factors, achieving greater performance than could be realised through the use of any single method.

			\subsubsection{Throughput}

				Table~\ref{tab:VGG16_HARDWARE} details the performance of state-of-the-art FPGA-based DNN inference engines targetting the CIFAR-10 (CNN), ImageNet (CNN) and TIMIT (RNN) datasets.
                Implementations are ordered according to power consumption, thus platforms of similar scales are adjacent.
                While categorised with respect to their target datasets, frameworks accelerating the inference of the same dataset may have been benchmarked using different DNN models and hence with dissimilar workloads.
                Some works did not report full-network workload information, making it impossible for us to quantify their throughputs.
                We thus detail arithmetic performance, which captures raw computational speed, as well.
                
                
                In general, custom hardware implementations exhibit up-to orders-of-magnitude higher throughput than GPP equivalents of similar scales, corresponding to the conclusions drawn in Section~\ref{sec:motivation}.
                Among the custom hardware implementations, the throughput of ASIC platforms is higher than other works with similar power consumption, largely due to their higher clock frequencies.
				
				By comparing Wang \emph{et al.}~\cite{FXP_CNN_PYNQ_ERWEI} and Zhao \emph{et al.}'s~\cite{BNN_CNN_FPGA17} CIFAR-10-targetting CNN implementations with the Going Deeper~\cite{NR_CNN_FXP_GOING_DEEPER}, fpgaConvNet~\cite{NR_CNN_FXP_FPGACONVNET} and FP-BNN~\cite{BNN_CNN_FP-BNN} ImageNet CNNs, all of which used FPGAs of similar scales, we can observe that, as precision is reduced, linear or even superlinear throughput increases can be achieved.
				Superlinear increases can be explained using the roofline modelling in Section~\ref{sec:motivation}.
				With quantisation on FPGAs, the use of cheaper fixed-point processing units allows for increased parallel-computing capability via area savings, in turn leading to increases in compute bounds.
                Arithmetic intensity can also be increased as model size decreases due to the opportunities presented by on-chip caching.
				The combined effect of these factors allows inference throughput to increase linearly if the baseline is memory bound, or superlinearly if compute bound.
				The accuracy-throughput tradeoff exposed through quantisation makes it possible for embedded-scale custom hardware implementations to beat even high-end GPPs in terms of inference throughput.
				This is evident throughout Table~\ref{tab:VGG16_HARDWARE}, in which the performance of schemes employing binarisation on custom hardware can be seen to have achieved either superior or comparable throughput to that of popular high-performance GPPs.
				
				EIE~\cite{PRU_CNN_EIE} and Li \emph{et al.}'s work~\cite{PRU_FPGA_SPARSIFICATION_FCCM} used pruning with fixed-point quantisation in ASICs and FPGAs, respectively, for CNN weight reduction.
				Comparing these against other works listed that used the same platform but without pruning, NeuFlow in ASICs~\cite{NEUFLOW} and Going Deeper in FPGAs~\cite{NR_CNN_FXP_GOING_DEEPER}, significantly superior arithmetic performance was obtained.
				This supports the other conclusion drawn from the roofline modelling in Section~\ref{sec:motivation}: with network compression, operational intensity increases due to reduced off-chip memory traffic, facilitating speedups.
				EIE, using fine-grained pruning with runtime zero-skipping, achieved a 19$\times$ improvement in arithmetic performance over NeuFlow, whereas Li \emph{et al.}'s work, using coarse-grained pruning, achieved only 2$\times$ improvement over Going Deeper.
                This seems to support the conclusion in Sections~\ref{sec:motivation} and \ref{sec:com_acc} that fine-grained pruning results in more workload reduction than coarse-grained, and that custom hardware allows for the design of efficient mechanisms to convert these reductions into speedups.
				
				\begin{sidewaystable}
					\centering
					\begin{threeparttable}
    					\caption{Comparison of large-scale DNN inference performance. Implementations are ordered by power consumption, lowest first.}
    					\begin{footnotesize}
    						\begin{tabular}[c]{cccccSSSSSc}
								\toprule
																								&															& \multicolumn{2}{c}{Quantisation(s)\tnote{1}}	& \multirow{2}[4]{*}{Platform}	& {\multirow{2}[4]{*}{\shortstack{Frequency\\(MHz)}}}	& {\multirow{2}[4]{*}{\shortstack{Throughput\\(cl/s)}}} & {\multirow{2}[4]{*}{\shortstack{Workload\\(Gop/cl)}}}	& {\multirow{2}[4]{*}{\shortstack{Arithmetic\\perf. (Gop/s)}}}	& {\multirow{2}[4]{*}{\shortstack{Efficiency\\(cl/J)}}}	& \multirow{2}[4]{*}{\shortstack{Approximation\\method(s)\tnote{1}}}	\\
																																							  \cmidrule{3-4}
																								&															& Weights		& Acts	\\
								\midrule
								\multirow{8}{*}{\rotatebox{90}{\shortstack{CNN\\(CIFAR-10)}}}	& Wang \emph{et al.}~\cite{FXP_CNN_PYNQ_ERWEI}				& FXP8			& FXP8							& Xilinx Zynq XC7Z020			& 100													& 103													& 0.0248												& 2.56															& 54.4													& FXP																	\\ 
																								& Zhao \emph{et al.}~\cite{BNN_CNN_FPGA17}					& BIN			& BIN							& Xilinx Zynq XC7Z020			& 143													& 168													& 1.24													& 208															& 35.6													& BIN																	\\ 
																								& CaffePresso~\cite{CAFFEPRESSO}							& FXP32			& FXP32							& Adapteva Parallella			& {--}													& 95.9													& 0.0146												& 1.40															& 14.2													& {--}																	\\ 
																								& FINN~\cite{BNN_CNN_FINN}									& BIN			& BIN							& Xilinx Zynq XC7Z045			& 200													& 21900													& 0.113													& 2500															& 3160													& BIN																	\\ 
																								& CaffePresso~\cite{CAFFEPRESSO}							& FXP16			& FXP16							& TI Keystone-II				& {--}													& 1000													& 0.0146												& 146															& 1000													& FXP																	\\ 
																								& FP-BNN~\cite{BNN_CNN_FP-BNN}								& BIN			& BIN							& Intel Stratix~V 5SGSD8		& 150													& 7640													& 1.23													& 9400															& 292													& BIN																	\\ 
																								& CPU~\cite{BNN_CNN_FP-BNN}									& FP32			& FP32							& Intel Xeon E5-2640			& 2500													& 147													& 1.23													& 181															& 1.55													& {--}																	\\ 
																								& GPU~\cite{BNN_CNN_FP-BNN}									& FP32			& FP32							& Nvidia Tesla K40				& 745													& 1510													& 1.23													& 1850															& 6.41													& {--}																	\\ 
								\midrule
								\multirow{14}{*}{\rotatebox{90}{\shortstack{CNN\\(ImageNet)}}}	& YodaNN~(0.60V)~\cite{BNN_CNN_YODANN}						& BIN			& FXP12							& 65~nm ASIC					& {--}													& 4.50													& 3.60													& 16.2															& 13400													& BIN																	\\ 
																								& DaDianNao~\cite{FXP_CNN_DADIANNAO}\tnote{2}				& FXP16			& FXP16							& 28~nm ASIC					& 606													& {--}													& {--}													& 452													        & {--}													& FXP																	\\ 
																								& EIE~\cite{PRU_CNN_EIE}\tnote{2}							& FXP4			& FXP16							& 45~nm ASIC					& 800													& {--}													& {--}													& 3000      													& {--}													& FXP, PRU, WS															\\ 
																								& NeuFlow~\cite{NEUFLOW}\tnote{2}							& FXP16			& FXP16							& 45~nm ASIC					& 400													& {--}													& {--}													& 160       													& {--}													& FXP																	\\ 
																								& fpgaConvNet~\cite{NR_CNN_FXP_FPGACONVNET}					& FXP16			& FXP16							& Xilinx Zynq XC7Z045			& 125													& 5.07													& 30.7													& 156															& 0.726													& FXP																	\\ 
																								& Angel-eye~\cite{NR_CNN_FXP_ANGEL-EYE}						& BFP8			& BFP8							& Xilinx Zynq XC7Z045			& 150													& 6.12													& 30.7													& 188															& 0.635													& BFP																	\\ 
																								& Going Deeper~\cite{NR_CNN_FXP_GOING_DEEPER}				& FXP16			& FXP16							& Xilinx Zynq XC7Z045			& 150													& 4.46													& 30.7													& 137															& 0.463													& FXP																	\\ 
																								& Li \emph{et al.}~\cite{PRU_FPGA_SPARSIFICATION_FCCM}		& {--}			& {--}							& Xilinx Zynq XC7Z045			& {--}													& 205													& 1.33													& 272															& {--}													& PRU																	\\ 
																								& Shen \emph{et al.}~\cite{FXP_CNN_FPGA_TOWARDS_A_UNIFORM}	& FXP16			& FXP16							& Xilinx Virtex US VCU440		& 200													& 26.7													& 30.7													& 821															& 1.03													& FXP																	\\ 
																								& FP-BNN~\cite{BNN_CNN_FP-BNN}								& BIN			& BIN							& Intel Stratix~V 5SGSD8		& 150													& 863													& 2.27													& 1960															& 33.0													& BIN																	\\ 
																								& TPU~\cite{FXP_CNN_TPU}\tnote{2}							& FXP8			& FXP8							& 28~nm ASIC					& 700													& {--}													& {--}													& 92000         												& {--}													& FXP																	\\ 
																								& HARPv2~\cite{FXP_CNN_FPGA_HARPv2}							& BIN			& BIN							& Intel HARPv2					& {--}													& 114													& 30.7													& 3500															& 2.37													& BIN																	\\ 
																								& GPU~\cite{FXP_CNN_FPGA_HARPv2}							& FP32			& FP32							& Nvidia Titan X				& {--}													& 121													& 30.7													& 3710															& 1.76													& {--}																	\\ 
																								& Brainwave~\cite{MICROSOFT_BRAINWAVE}						& BFP5			& BFP5							& Intel Arria 10		        & 300													& 559													& 7.80													& 4360															& 4.47													& BFP															        \\ 
								\midrule
								\multirow{7}{*}{\rotatebox{90}{\shortstack{RNN\\(TIDIGITs)}}}	& Wang \emph{et al.}~\cite{SM_LSTM_HYBRID_CIRCULAR}			& LOG8          & FXP8							& 90~nm ASIC					& 600													& 585000												& 0.00421												& 2460															& 580000												& FXP, LOG, ACT, STR													\\ 
																								& DeltaRNN~\cite{DR_GRU_FPGA_DeltaRNN}						& FXP16			& FXP16							& Xilinx Zynq XC7Z100			& 125													& 2650000													& 0.000453  											& 1200															& 362000													& FXP, ACT, IDC															\\ 
																								& C-LSTM FFT8~\cite{SM_CNN_FPGA_C-LSTM}						& FXP16			& FXP16							& Xilinx Kintex US XCKU060		& 200													& 195000												& 0.208													& 40600															& 8130													& FXP, ACT, STR															\\ 
																								& ESE~\cite{R_LSTM_FXP_ESE}									& FXP12			& FXP16							& Xilinx Kintex US XCKU060		& 200													& 12100													& 0.208													& 2520															& 296													& FXP, ACT, PRU															\\ 
																								& CPU~\cite{R_LSTM_FXP_ESE}							        & FP32			& FP32							& Intel i7-5930K				& {--}													& 166												& 0.208													& 34.6															& 1.50												& {--}																	\\ 
																								& Brainwave~\cite{MICROSOFT_BRAINWAVE}						& BFP5			& BFP5							& Intel Stratix 10		        & 250													& 13500													& 1.67													& 22600															& 108													& BFP															        \\ 
																								& GPU~\cite{R_LSTM_FXP_ESE}							        & FP32			& FP32							& Nvidia Titan X			& {--}													& 4160													& 0.208													& 866															& 20.6													& {--}																	\\ 
								\bottomrule
    						\end{tabular}
    						\begin{tablenotes}
                                \item[1] FXP: fixed point. BFP: block floating point. BIN: binary. LOG: logarithmic. ACT: activation function approximation. PRU: pruning. STR: structured matrix. WS: weight sharing. IDC: input-dependent computation.
                                \item[2] Reported arithmetic performance is a ``peak" value, not that for any particular network, since the authors did not report the latter.
                            \end{tablenotes}
    					\end{footnotesize}
    					\label{tab:VGG16_HARDWARE}
    				\end{threeparttable}
				\end{sidewaystable}

				As mentioned in Section~\ref{sec:SM}, circulant matrix-based methods do not work well with CNNs due to their significant accuracy losses, yet they provide exceptionally good accuracy and compression for RNNs.
				This is reflected in Table~\ref{tab:VGG16_HARDWARE}, in which it is shown that C-LSTM exhibited 47$\times$ and 390$\times$ gains in throughput and efficiency, respectively, compared to a GPU implementation~\cite{SM_CNN_FPGA_C-LSTM}.
				Among all RNN implementations listed, those that employed block-circulant matrices or input-dependent computation achieved superior throughputs and efficiencies \emph{vs} the remainder since the use of these methods resulted in the greatest workload reductions.

				Almost all of the listed RNN FPGA frameworks made use of hybrid strategies, featuring processing elements tailored to low-precision computation along with weight reduction, achieving significant throughput improvements compared to GPU alternatives.

			\subsubsection{Latency}
            
				While the majority of existing works in the field are throughput- or energy-oriented, some DNN applications prioritise latency instead.
                Some implementations simultaneously achieved good throughput and latency performance.
                Ma \emph{et al.} implemented VGG-16 on FPGAs with fixed-point quantisation for ImageNet classification~\cite{NR_CNN_FXP_OPTIM_LOOP}.
                Tradeoffs between resource consumption and throughput were systematically analysed, with high performance achieved by balancing memory traffic and computation.
                The authors reported throughput of 21~cl/s and latency of 48~ms, both of which are 4.7$\times$ higher than the previous state of the art, Going Deeper~\cite{NR_CNN_FXP_GOING_DEEPER}.

				The earliest version of fpgaConvNet was throughput-oriented~\cite{NR_CNN_FXP_FPGACONVNET}.
				The authors later extended their design-space exploration tool to optimise for latency in addition to throughput, demonstrating outstanding latency-critical application performance \emph{vs} alternative embedded implementations~\cite{FPGA_LATENCY_BOUGANIS_FPL}.
                Zhang \emph{et al.} also presented an FPGA-based RNN/CNN inference framework, providing highly configurable layer templates and a design-space exploration engine for resource allocation management facilitating design optimisation for resource-constrained latency minimisation~\cite{FXP_RNN_VS_GPU}.
                
                Hardware implementations of input-dependent computation methods have an intrinsic emphasis on latency.
                Due to their conditional computation nature, pipeline stalls happen frequently, reducing throughput.
                This is not a problem for latency-driven applications, however, in which the inference batch size is normally one.
                Implementations based on input-dependent methods, \emph{e.g.} CascadeCNN~\cite{DR_CNN_FPGA_CASCADECNN}, are able to achieve significant latency reductions. 

			\subsubsection{Energy Efficiency}

				Table~\ref{tab:VGG16_HARDWARE} also facilitates the energy efficiency comparison of DNN inference implementations.
                Given a constant power budget, higher throughput translates to higher energy efficiency.
                Thus, approximation methods leading to higher parallelism and workload and/or off-chip memory transfer reductions, such as binarisation~\cite{BNN_CNN_YODANN}, logarithmic quantisation~\cite{SM_LSTM_HYBRID_CIRCULAR} and block-circulant matrices~\cite{SM_CNN_FPGA_C-LSTM}, tend to result in higher energy efficiencies over alternative techniques with comparable network topologies and power consumptions.
                
                When comparing platforms with similar throughput, the efficiency of power-hungry high-end GPPs tends to be lower than custom hardware implementations'.
                These facilitate parallelism at low precisions, achieving high throughput when running at a few hundred MHz, while CPUs and GPUs tend to operate at speeds on the order of GHz.
                For example, a binary HARPv2 implementation can provide comparable throughput to a Titan~X~Pascal GPU's, but is 24\% more energy efficient~\cite{FXP_CNN_FPGA_HARPv2}.
                
                The ASIC implementations achieve the highest energy efficiencies, primarily because they are not configurable and thus have lower capacitive loading than FPGA equivalents.
                Due to hardware overheads allowing for arbitrary logic and routing configurations and their lack of clock tree customisability, FPGAs can never compete with ASICs in terms of energy efficiency, yet FPGA implementations are still significantly more efficient than GPPs~\cite{BG_FPGA_ASIC_ENERGY}.
                Memory hierarchy customisability also facilitates efficiency improvements, as was shown for YodaNN~\cite{BNN_CNN_YODANN}.
                
         \subsection{Application-specific Considerations}
         
            \subsubsection{Retraining Time and Parameter Fine-tuning}
             
                Fixed-point and logarithmic quantisation, pruning and input-dependent compute methods require post-approximation retraining.
    			The majority of the pruning methods captured in Fig.~\ref{ALGO_PRUNE} use $l_1$ and $l_2$ regularisers.
    			Their employment, however, tends to result in more iterations being required to achieve convergence, increasing training time.
    			Ullrich \emph{et al.} reported that training of networks exploiting the soft weight-sharing method is very slow for large-scale datasets~\cite{PRU_CNN_SOFT_WEIGHT_SHARING}.
    			Furthermore, the search for so-called \emph{hyper-parameters}, such as pruning thresholds and quantisation precisions, can be cumbersome and expensive~\cite{PRU_CNN_TRAIN_PRUNE_RETRAIN,R_CNN_ADAPTIVE_QUANTISATION}. 
    
    			The use of low-rank factorisation tends to necessitate more retraining iterations for convergence than alternative methods since layer-wise factorisation results in increased network depth, exacerbating the problem of vanishing gradients in DNNs.
    			Factorisation is also compute-intensive.
             
             \subsubsection{Parameterisation}
             
                 During hardware design-space exploration, ASIC designs and some early FPGA-based works were only optimised for a single design metric: usually throughput.
                 Many recent FPGA-based works have introduced general-purpose DNN accelerator frameworks which can cater to different design considerations based on desired application requirements.
                 As a follow-up to FPGA-based framework fpgaConvNet~\cite{NR_CNN_FXP_FPGACONVNET}, Stylianos \emph{et al.} extended their automatic design-space exploration algorithm to also support area and latency optimisation~\cite{FPGA_LATENCY_BOUGANIS_FPL}.
             
             \subsubsection{Hardware Design and Turnaround}
             
                 Due to the rapidly evolving landscape of DNN algorithmic development, the flexibility of the hardware design process becomes a practical issue.
                 With a time- and resource-consuming process, an inference platform could well become obsolete before it is manufactured.
                 The design, fabrication and validation of ASICs normally take months, if not years, to complete.
                 Such slow turnarounds expose DNN application designers to high risks in terms of time and monetary investment.
                 GPPs, on the other hand, are well supported by full-stack DNN design frameworks using high-level front ends, with which approximation methods can be prototyped in weeks.
                 Compared with these two families of platforms, FPGAs provide a useful tradeoff between performance and design costs.
                 High-level synthesis tools reduce design difficulty and lead time while allowing the obtainment of high throughput and energy efficiency. 
             
             \subsubsection{Regularisation}
             \label{sec:regularisation}

                The authors of works exploiting many approximation methods, including low-precision quantisation~\cite{BNN_CNN_BinaryConnect,TNN_LSTM_BENGIO,TNN_CNN_GRAD_NOISE}, pruning~\cite{PRU_CNN_TRAIN_PRUNE_RETRAIN,PRU_LSTM_NMT_TRAIN_PRUNE_RETRAIN} and weight sharing~\cite{PRU_CNN_HASHEDNET}, reported accuracies greater than FP32 baselines after their application.
                Courbariaux \emph{et al.} explained that low-precision quantisation limits network capacity, forcing networks to leave local minima and find broader minima instead, improving generalisability by avoiding overfitting~\cite{BNN_CNN_BinaryConnect}.
                Similarly, in FITNet, the student network achieved 10$\times$ compression but a 1.4~pp accuracy improvement over its teacher due to the regularisation effect from reduced network complexity~\cite{KD_FITNETS}.
                The authors of HashedNets explained that the random ``virtual" connections generated by their parameter hashing increased network expressiveness~\cite{PRU_CNN_HASHEDNET}.
                Similar to dropout layers in DNN training, the introduction of randomness from approximation, in the form of either quantisation noise or connections, creates regularisation that improves the accuracy of smaller networks.

	\section{Future Directions}
	\label{sec:trends}

		Now that we have evaluated the current trends in the field of DNN approximation algorithms and their implementations, we are in a position to propose some promising future research directions.
		  
		\subsection{Evaluation Methodologies}\label{sec:evaluation_methodologies}

            In the development of throughput-oriented DNN algorithm implementations, being able to identify bottlenecks is crucial to the efficiency of research.
            A misidentification of a bottleneck's source usually leads to wasted design effort.
            In many publications to date, authors have employed \emph{ad hoc} evaluation methodologies, reporting improvements against seemingly arbitrary DNN benchmarks without systematically determining their baselines' bottlenecks, how the characteristics of the selected models affect those bottlenecks or how far away design points are from theoretical maxima.
    
            One of the major issues with DNN evaluation is the emphasis currently placed by many authors on peak arithmetic performance (in op/s).
            For example, the authors of the TPU stated that their architecture can achieve 92~Top/s~\cite{FXP_CNN_TPU}.
            When tested with real DNN layers, however, that actually achieved was below 15~Top/s due to memory bandwidth limits for all cases but one with a particularly high operational intensity.
            A focus on peak op/s can potentially lead to ignorance of the importance of microarchitectural design, making post-deployment accelerator efficiency underwhelming.
                
            In Section~\ref{sec:motivation}, we compared the acceleration potential of DNN inference platforms using roofline modelling.
            For cross-platform evaluation, such models are useful since they present major bottlenecks in uniform and comparable formats, allowing the relative strengths and weaknesses of those platforms to be contrasted.
            Some authors have extended roofline modelling in order to capture other metrics.
            For example, in an attempt to analyse the tradeoff between energy efficiency and performance, Sayed \emph{et al.} added frequency as a third axis, allowing power draw estimation~\cite{EXTENDED_ROOFLINE_MODEL}.
                
            For comparison of \emph{implementations}, however---particularly those on the same platform---we are of the opinion that the use of roofline modelling is misguided.
            While points showing achieved arithmetic performance could be added to roofline plots, showing how much of their compute and memory bandwidth potential particular implementations achieve, the methodology's inherent orientation to arithmetic performance obscures other factors affecting analysis: chiefly workload.
            Two otherwise identical implementations with different levels of pruning, for example, may well exhibit negatively correlated op/s and cl/s, potentially making comparison of arithmetic performance misleading.
            In an attempt to tackle this, metrics including ``equivalent throughput" (the arithmetic performance of a post-pruned network using the pre-pruning workload) have been introduced and are unfortunately now commonplace~\cite{PRU_CNN_EIE,SM_CNN_FPGA_CIRCNN}.
            We consider these to be unmeaningful and to needlessly distract from consideration of fundamental measures, particularly classification rate.
                
            We encourage the community to report sustained throughput (in cl/s or similar) for standard, up-to-date models and datasets in preference to (peak) arithmetic performance.
            In conducting the research for this article, we encountered many issues with performance comparison owing to authors evaluating their works very differently, with some of the benchmarks used unpopular or even obsolete.
            Emerging benchmark suites such as MLPerf and DeepBench, which provide selections of widely accepted and current test cases, should be used for comprehensive evaluation, thereby also facilitating apples-to-apples comparison.
            
        \subsection{Research Objectives}

    
    
    

            \subsubsection{Convergence Guarantees and Optimal Design Choices}

    			Many approximation methods do not yet have mathematical proofs of guaranteed convergence, meaning that existing methods may not be applicable to new DNN models.
    			We are therefore of the opinion that theoretical investigation into each such method's convergence would be a very useful endeavour.
    			As a counterexample, Li \emph{et al.} provided derivations for quantised DNNs' convergence criteria~\cite{TRAINING_QUANTIZED_NETS_CONVERGENCE_GUARANTEE}.
    			Sakr \emph{et al.} also investigated analytical guarantees on the numerical precision of DNNs with fixed-point quantisation~\cite{ANALYTICAL_GUARANTEES_ON_NUMERICAL_PRECISION_OF_DNNS}.
    
    			It would also be interesting to prove the existence of optimal design choices for each method.
    			For example, Tai \emph{et al.}~\cite{LRF_SVD_WITH_OPTIMISER} suggested that the CP decomposition proposed by Lebedev \emph{et al.}~\cite{LRF_CP_ICLR} does not guarantee an optimal rank-$r$ factorisation since the problem of finding the best low-rank CP factorisation is ill-posed~\cite{LRF_SVD_CP_OPTIMISER_ILLPOSED}.
    			Similarly, for circulant matrix methods, we can clearly observe a difference in accuracy degradation between CNNs and RNNs, but it is not yet possible to explain this discrepancy mathematically.
                A good understanding of the convergence and applicability of the various approximation methods would be beneficial to allow for their generalisation. 

		    \subsubsection{Self-adaptive Hyper-parameter Fine-tuning}

    			During quantisation and pruning, many hyper-parameters need to be determined through extensive manual fine-tuning with a validation dataset.
    			This will become infeasible as networks deepen.
    			Those with dynamic fine-tuning mechanisms are therefore potentially more scalable than those requiring manual intervention.
    			As examples of the former, Bengio \emph{et al.}~\cite{DR_CNN_REINFORCEMENT} and Lin \emph{et al.}~\cite{DR_CNN_RUNTIME_PRUNING} made pruning decisions using a Markov decision process, Liu \emph{et al.} performed filter pruning using trainable scaling factors~\cite{PRU_CNN_SLIMMING}, Shin \emph{et al.} learnt quantisation granularities via retraining~\cite{R_CNN_FXP_STEP_SIZE} and Yang \emph{et al.} removed filters to meet resource constraints~\cite{PRU_CNN_NETADAPT}.
                If self-adaptive network fine tuning can be generalised to different hyper-parameters and network models, the latency of DNN application design could be significantly reduced.
            
            \subsubsection{FPGA-ASIC Heterogeneous Systems}
        
                From Table~\ref{tab:VGG16_HARDWARE}, we can conclude that, while FPGAs are extremely flexible, ASICs offer the greatest performance. 
                Instead of focussing on purely FPGA- or ASIC-only solutions, Nurvitadhi \emph{et al.} proposed the single-package, heterogeneous integration of FPGAs and ASICs using Intel's Embedded Multi-die Interconnect Bridge~\cite{FPGA_ASIC_EMIB_FPL}.
                In their system, the ASIC components, called TensorTiles, execute typical DNN operations such as matrix-vector MACs at eight-bit or lower precision, while the FPGA enables the application-specific optimisation of data management and scheduling.
                With two TensorTiles and one FPGA, this design demonstrated 3.3$\times$ and 4.0$\times$ improvements in energy efficiency and throughput, respectively, with AlexNet against an FPGA-only implementation on an Intel Stratix~10.
                This work proved that such heterogeneous systems are promising platforms for DNN applications and thus deserve particular attention.
                Xilinx's recently announced Adaptive Compute Acceleration Platform, featuring a hardened array of processors suited to neural network compute interfaced with soft logic through a network on chip, was designed to simultaneously achieve high performance and flexibility~\cite{XILINX_EVEREST}.

    

		    \subsubsection{Hardware Inference of Irregular Data Patterns}

    			While fine-grained pruning can lead to high compression, it also produces data distribution irregularity, making conversion of compression into speedups challenging~\cite{PRU_CNN_TRAIN_PRUNE_RETRAIN,PRU_LSTM_NMT_TRAIN_PRUNE_RETRAIN,R_LSTM_FXP_ESE}.
    			For example, for AlexNet on GPUs with structured pruning, a compression ratio of 3.0 led to 3.0$\times$ greater throughput~\cite{PRU_CNN_GROUP_WISE_BRAIN_DAMAGE}, while, in contrast, element-wise pruning resulted in superior compression (9.0$\times$) but the same throughput~\cite{PRU_CNN_TRAIN_PRUNE_RETRAIN}.
                In this context, there is an emerging need for hardware accelerators to support compressed and sparse networks to become competitive high-performance, low-power GPP alternatives.
                Works based on custom hardware, such as ESE~\cite{R_LSTM_FXP_ESE} on FPGAs and Cnvlutin~\cite{PRU_ASIC_CNVLUTIN} and Minerva~\cite{PRU_CNN_MINERVA} on ASICs, featured fast and dynamic arithmetic operation avoidance suiting fine-grained pruning, achieving superior throughput and energy efficiency to GPP implementations.
                Future works should explore the further use of design flexibility to realise more acceleration from sparsity.
			
            \subsubsection{Parameter Hardening}
         
                Almost all works exploiting existing approximation still see the storage of parameters in DRAM for hardware reusability and scalability.
                With the large memory transfer reductions achievable through the use of aggressive methods including binarisation, logarithmic quantisation and weight sharing, however, smaller-sized parameters can fit on-chip more easily.
                It has thus become increasingly sensible to harden parameters into logic, reducing off-chip memory fetches.
                In some cases, memory fetching can be eliminated entirely.
                With base-two logarithmic quantisation, for example, multiplications are converted into binary shifts, which, when hardened, can be implemented without consuming any logic.
                Industrial firms such as Microsoft and Google have focussed their efforts on the optimisation of datacentre-scale DNN inference with custom ASIC~\cite{FXP_CNN_TPU} and FPGA~\cite{MICROSOFT_BRAINWAVE} designs.
                Their huge throughput and energy efficiency requirements justify the use of extremely large and specialised accelerators employing loop unrolling and parameter hardening.
                Future research can explore the feasibility of this approach, showing how it trades off design reusability and scalability for throughput and efficiency.
                
	\section{Summary}

		In this article, we discussed the past, present and future of DNN approximation for custom hardware.
		With a roofline model analysis, we explained why DNNs' algorithmic advancement favours custom implementations, demonstrating how FPGAs and ASICs can offer performance superior to that of alternative platforms through the exploitation of approximation.
		With a comprehensive selection of state-of-the-art publications, we presented in-depth evaluations and comparisons of DNN approximation algorithms along with their respective hardware implementations.
		We summarised the current trends in the field, based on which we proposed several research questions which are yet to be sufficiently answered.
		Through this work, we hope to inspire new and exciting developments in DNN approximation that tap into the full potential offered by custom hardware platforms.

    \bibliographystyle{ACM-Reference-Format}
    \bibliography{survey_bibliography}


\begin{thebibliography}{00}


\ifx \showCODEN    \undefined \def \showCODEN     #1{\unskip}     \fi
\ifx \showDOI      \undefined \def \showDOI       #1{{\tt DOI:}\penalty0{#1}\ }
  \fi
\ifx \showISBNx    \undefined \def \showISBNx     #1{\unskip}     \fi
\ifx \showISBNxiii \undefined \def \showISBNxiii  #1{\unskip}     \fi
\ifx \showISSN     \undefined \def \showISSN      #1{\unskip}     \fi
\ifx \showLCCN     \undefined \def \showLCCN      #1{\unskip}     \fi
\ifx \shownote     \undefined \def \shownote      #1{#1}          \fi
\ifx \showarticletitle \undefined \def \showarticletitle #1{#1}   \fi
\ifx \showURL      \undefined \def \showURL       #1{#1}          \fi
\providecommand\bibfield[2]{#2}
\providecommand\bibinfo[2]{#2}
\providecommand\natexlab[1]{#1}

\bibitem[\protect\citeauthoryear{AI}{AI}{2017}]%
        {INTEL_NERVANA}
\bibfield{author}{\bibinfo{person}{Intel AI}.} \bibinfo{year}{2017}\natexlab{}.
\newblock \bibinfo{title}{{Intel Nervana Neural Network Processors (NNP)
  Redefine AI Silicon}}.
\newblock   (\bibinfo{year}{2017}).
\newblock
\showURL{%
\url{https://ai.intel.com/intel-nervana-neural-network-processors-nnp-redefine-ai-silicon/}}


\bibitem[\protect\citeauthoryear{Akopyan, Sawada, Cassidy, Alvarez-Icaza,
  Arthur, Merolla, Imam, Nakamura, Datta, and Nam}{Akopyan
  et~al\mbox{.}}{2015}]%
        {FXP_CNN_TRUENORTH}
\bibfield{author}{\bibinfo{person}{Filipp Akopyan}, \bibinfo{person}{Jun
  Sawada}, \bibinfo{person}{Andrew Cassidy}, \bibinfo{person}{Rodrigo
  Alvarez-Icaza}, \bibinfo{person}{John Arthur}, \bibinfo{person}{Paul
  Merolla}, \bibinfo{person}{Nabil Imam}, \bibinfo{person}{Yutaka Nakamura},
  \bibinfo{person}{Pallab Datta}, {and} \bibinfo{person}{Gi-Joon Nam}.}
  \bibinfo{year}{2015}\natexlab{}.
\newblock \showarticletitle{{TrueNorth: Design and Tool Flow of a 65 mW 1
  Million Neuron Programmable Neurosynaptic Chip}}.
\newblock \bibinfo{journal}{{\em IEEE Transactions on Computer-aided Design of
  Integrated Circuits and Systems\/}} \bibinfo{volume}{{34}, 10}
  (\bibinfo{year}{2015}).
\newblock


\bibitem[\protect\citeauthoryear{Albericio, Delm{\'a}s, Judd, Sharify, O'Leary,
  Genov, and Moshovos}{Albericio et~al\mbox{.}}{2017}]%
        {FXP_CNN_PRAGMATIC}
\bibfield{author}{\bibinfo{person}{Jorge Albericio}, \bibinfo{person}{Alberto
  Delm{\'a}s}, \bibinfo{person}{Patrick Judd}, \bibinfo{person}{Sayeh Sharify},
  \bibinfo{person}{Gerard O'Leary}, \bibinfo{person}{Roman Genov}, {and}
  \bibinfo{person}{Andreas Moshovos}.} \bibinfo{year}{2017}\natexlab{}.
\newblock \showarticletitle{{Bit-pragmatic Deep Neural Network Computing}}. In
  \bibinfo{booktitle}{{\em IEEE/ACM International Symposium on
  Microarchitecture}}.
\newblock


\bibitem[\protect\citeauthoryear{Albericio, Judd, Hetherington, Aamodt, Jerger,
  and Moshovos}{Albericio et~al\mbox{.}}{2016}]%
        {PRU_ASIC_CNVLUTIN}
\bibfield{author}{\bibinfo{person}{Jorge Albericio}, \bibinfo{person}{Patrick
  Judd}, \bibinfo{person}{Tayler Hetherington}, \bibinfo{person}{Tor Aamodt},
  \bibinfo{person}{Natalie~E. Jerger}, {and} \bibinfo{person}{Andreas
  Moshovos}.} \bibinfo{year}{2016}\natexlab{}.
\newblock \showarticletitle{{Cnvlutin: Ineffectual-neuron-free Deep Neural
  Network Computing}}. In \bibinfo{booktitle}{{\em ACM SIGARCH Computer
  Architecture News}}.
\newblock


\bibitem[\protect\citeauthoryear{Alemdar, Leroy, Prost-Boucle, and
  P{\'e}trot}{Alemdar et~al\mbox{.}}{2017}]%
        {KD_TNN_FPGA}
\bibfield{author}{\bibinfo{person}{Hande Alemdar}, \bibinfo{person}{Vincent
  Leroy}, \bibinfo{person}{Adrien Prost-Boucle}, {and}
  \bibinfo{person}{Fr{\'e}d{\'e}ric P{\'e}trot}.}
  \bibinfo{year}{2017}\natexlab{}.
\newblock \showarticletitle{{Ternary Neural Networks for Resource-efficient AI
  Applications}}. In \bibinfo{booktitle}{{\em International Joint Conference on
  Neural Networks}}.
\newblock


\bibitem[\protect\citeauthoryear{Almahairi, Ballas, Cooijmans, Zheng,
  Larochelle, and Courville}{Almahairi et~al\mbox{.}}{2016}]%
        {DR_CNN_DCN}
\bibfield{author}{\bibinfo{person}{Amjad Almahairi}, \bibinfo{person}{Nicolas
  Ballas}, \bibinfo{person}{Tim Cooijmans}, \bibinfo{person}{Yin Zheng},
  \bibinfo{person}{Hugo Larochelle}, {and} \bibinfo{person}{Aaron Courville}.}
  \bibinfo{year}{2016}\natexlab{}.
\newblock \showarticletitle{{Dynamic Capacity Networks}}. In
  \bibinfo{booktitle}{{\em International Conference on Machine Learning}}.
\newblock


\bibitem[\protect\citeauthoryear{Amara, Amiel, and Ea}{Amara
  et~al\mbox{.}}{2006}]%
        {BG_FPGA_ASIC_ENERGY}
\bibfield{author}{\bibinfo{person}{Amara Amara}, \bibinfo{person}{Frederic
  Amiel}, {and} \bibinfo{person}{Thomas Ea}.} \bibinfo{year}{2006}\natexlab{}.
\newblock \showarticletitle{{FPGA vs. ASIC for Low Power Applications}}.
\newblock \bibinfo{journal}{{\em Microelectronics Journal\/}}
  \bibinfo{volume}{{37}, 8} (\bibinfo{year}{2006}).
\newblock


\bibitem[\protect\citeauthoryear{Amin, Curtis, and Hayes-Gill}{Amin
  et~al\mbox{.}}{1997}]%
        {PLAN}
\bibfield{author}{\bibinfo{person}{Hesham Amin}, \bibinfo{person}{K.~Mervyn
  Curtis}, {and} \bibinfo{person}{Barrie~R. Hayes-Gill}.}
  \bibinfo{year}{1997}\natexlab{}.
\newblock \showarticletitle{{Piecewise Linear Approximation Applied to
  Nonlinear Function of a Neural Network}}.
\newblock \bibinfo{journal}{{\em IEE Proceedings -- Circuits, Devices and
  Systems\/}} \bibinfo{volume}{{144}, 6} (\bibinfo{year}{1997}).
\newblock


\bibitem[\protect\citeauthoryear{Andri, Cavigelli, Rossi, and Benini}{Andri
  et~al\mbox{.}}{2018}]%
        {BNN_CNN_YODANN}
\bibfield{author}{\bibinfo{person}{Renzo Andri}, \bibinfo{person}{Lukas
  Cavigelli}, \bibinfo{person}{Davide Rossi}, {and} \bibinfo{person}{Luca
  Benini}.} \bibinfo{year}{2018}\natexlab{}.
\newblock \showarticletitle{{YodaNN: An Architecture for Ultra-low Power
  Binary-weight CNN Acceleration}}.
\newblock \bibinfo{journal}{{\em IEEE Transactions on Computer-aided Design of
  Integrated Circuits and Systems\/}} \bibinfo{volume}{{37}, 1}
  (\bibinfo{year}{2018}).
\newblock


\bibitem[\protect\citeauthoryear{Ayat, Khalil-Hani, and Ab~Rahman}{Ayat
  et~al\mbox{.}}{2018}]%
        {EXTENDED_ROOFLINE_MODEL}
\bibfield{author}{\bibinfo{person}{Sayed~O. Ayat}, \bibinfo{person}{Mohamed
  Khalil-Hani}, {and} \bibinfo{person}{Ab~Al-Hadi Ab~Rahman}.}
  \bibinfo{year}{2018}\natexlab{}.
\newblock \showarticletitle{{Optimizing FPGA-based CNN Accelerator for Energy
  Efficiency with an Extended Roofline Model}}.
\newblock \bibinfo{journal}{{\em Turkish Journal of Electrical Engineering \&
  Computer Sciences\/}} \bibinfo{volume}{{26}, 2} (\bibinfo{year}{2018}).
\newblock


\bibitem[\protect\citeauthoryear{Ba and Caruana}{Ba and Caruana}{2014}]%
        {KD_DO_NEED_DEEP}
\bibfield{author}{\bibinfo{person}{Jimmy Ba} {and} \bibinfo{person}{Rich
  Caruana}.} \bibinfo{year}{2014}\natexlab{}.
\newblock \showarticletitle{{Do Deep Nets Really Need to be Deep?}}. In
  \bibinfo{booktitle}{{\em Conference on Neural Information Processing
  Systems}}.
\newblock


\bibitem[\protect\citeauthoryear{Bell and Garland}{Bell and Garland}{2009}]%
        {PRU_GPU_BELL}
\bibfield{author}{\bibinfo{person}{Nathan Bell} {and} \bibinfo{person}{Michael
  Garland}.} \bibinfo{year}{2009}\natexlab{}.
\newblock \showarticletitle{{Implementing Sparse Matrix-vector Multiplication
  on Throughput-oriented Processors}}. In \bibinfo{booktitle}{{\em Conference
  on High Performance Computing Networking, Storage and Analysis}}.
\newblock


\bibitem[\protect\citeauthoryear{Bengio, Bacon, Pineau, and Precup}{Bengio
  et~al\mbox{.}}{2015}]%
        {DR_CNN_REINFORCEMENT}
\bibfield{author}{\bibinfo{person}{Emmanuel Bengio},
  \bibinfo{person}{Pierre-Luc Bacon}, \bibinfo{person}{Joelle Pineau}, {and}
  \bibinfo{person}{Doina Precup}.} \bibinfo{year}{2015}\natexlab{}.
\newblock \showarticletitle{{Conditional Computation in Neural Networks for
  Faster Models}}. In \bibinfo{booktitle}{{\em International Conference on
  Learning Representations}}.
\newblock


\bibitem[\protect\citeauthoryear{Bengio, L{\'e}onard, and Courville}{Bengio
  et~al\mbox{.}}{2013}]%
        {DR_CNN_STS}
\bibfield{author}{\bibinfo{person}{Yoshua Bengio}, \bibinfo{person}{Nicholas
  L{\'e}onard}, {and} \bibinfo{person}{Aaron Courville}.}
  \bibinfo{year}{2013}\natexlab{}.
\newblock \showarticletitle{{Estimating or Propagating Gradients Through
  Stochastic Neurons for Conditional Computation}}.
\newblock \bibinfo{journal}{{\em arXiv preprint arXiv:1308.3432\/}}
  (\bibinfo{year}{2013}).
\newblock


\bibitem[\protect\citeauthoryear{Boutros, Yazdanshenas, and Betz}{Boutros
  et~al\mbox{.}}{2018}]%
        {FXP_CNN_FPL_DSP}
\bibfield{author}{\bibinfo{person}{Andrew Boutros}, \bibinfo{person}{Sadegh
  Yazdanshenas}, {and} \bibinfo{person}{Vaughn Betz}.}
  \bibinfo{year}{2018}\natexlab{}.
\newblock \showarticletitle{{Embracing Diversity: Enhanced DSP Blocks for
  Low-precision Deep Learning on FPGAs}}. In \bibinfo{booktitle}{{\em
  International Conference on Field-programmable Logic and Applications}}.
\newblock


\bibitem[\protect\citeauthoryear{Cai, He, Sun, and Vasconcelos}{Cai
  et~al\mbox{.}}{2017}]%
        {BNN_CNN_HWGO}
\bibfield{author}{\bibinfo{person}{Zhaowei Cai}, \bibinfo{person}{Xiaodong He},
  \bibinfo{person}{Jian Sun}, {and} \bibinfo{person}{Nuno Vasconcelos}.}
  \bibinfo{year}{2017}\natexlab{}.
\newblock \showarticletitle{{Deep Learning with Low Precision by Half-wave
  Gaussian Quantization}}. In \bibinfo{booktitle}{{\em IEEE Conference on
  Computer Vision and Pattern Recognition}}.
\newblock


\bibitem[\protect\citeauthoryear{Caulfield, Chung, Putnam, Angepat, Fowers,
  Haselman, Heil, Humphrey, Kaur, Kim, Lo, Massengill, Ovtcharov, Papamichael,
  Woods, Lanka, Chiou, and Burger}{Caulfield et~al\mbox{.}}{2016}]%
        {MICROSOFT_CATAPULT}
\bibfield{author}{\bibinfo{person}{Adrian Caulfield}, \bibinfo{person}{Eric
  Chung}, \bibinfo{person}{Andrew Putnam}, \bibinfo{person}{Hari Angepat},
  \bibinfo{person}{Jeremy Fowers}, \bibinfo{person}{Michael Haselman},
  \bibinfo{person}{Stephen Heil}, \bibinfo{person}{Matt Humphrey},
  \bibinfo{person}{Puneet Kaur}, \bibinfo{person}{Joo-Young Kim},
  \bibinfo{person}{Daniel Lo}, \bibinfo{person}{Todd Massengill},
  \bibinfo{person}{Kalin Ovtcharov}, \bibinfo{person}{Michael Papamichael},
  \bibinfo{person}{Lisa Woods}, \bibinfo{person}{Sitaram Lanka},
  \bibinfo{person}{Derek Chiou}, {and} \bibinfo{person}{Doug Burger}.}
  \bibinfo{year}{2016}\natexlab{}.
\newblock \showarticletitle{{A Cloud-scale Acceleration Architecture}}. In
  \bibinfo{booktitle}{{\em International Symposium on Microarchitecture}}.
\newblock


\bibitem[\protect\citeauthoryear{Chang and Culurciello}{Chang and
  Culurciello}{2017}]%
        {NR_LSTM_FXP_Q882}
\bibfield{author}{\bibinfo{person}{Andre Xian~Ming Chang} {and}
  \bibinfo{person}{Eugenio Culurciello}.} \bibinfo{year}{2017}\natexlab{}.
\newblock \showarticletitle{{Hardware Accelerators for Recurrent Neural
  Networks on FPGA}}. In \bibinfo{booktitle}{{\em International Symposium on
  Circuits and Systems}}.
\newblock


\bibitem[\protect\citeauthoryear{Chen, Seff, Kornhauser, and Xiao}{Chen
  et~al\mbox{.}}{2015a}]%
        {BG_AUTO_DRIVING_1}
\bibfield{author}{\bibinfo{person}{Chenyi Chen}, \bibinfo{person}{Ari Seff},
  \bibinfo{person}{Alain Kornhauser}, {and} \bibinfo{person}{Jianxiong Xiao}.}
  \bibinfo{year}{2015}\natexlab{a}.
\newblock \showarticletitle{{Deepdriving: Learning Affordance for Direct
  Perception in Autonomous Driving}}. In \bibinfo{booktitle}{{\em IEEE
  International Conference on Computer Vision}}.
\newblock


\bibitem[\protect\citeauthoryear{Chen, Choi, Yu, Han, and Chandraker}{Chen
  et~al\mbox{.}}{2017a}]%
        {KD_WEIGHTED_CROSS_ENTROPY}
\bibfield{author}{\bibinfo{person}{Guobin Chen}, \bibinfo{person}{Wongun Choi},
  \bibinfo{person}{Xiang Yu}, \bibinfo{person}{Tony Han}, {and}
  \bibinfo{person}{Manmohan Chandraker}.} \bibinfo{year}{2017}\natexlab{a}.
\newblock \showarticletitle{{Learning Efficient Object Detection Models with
  Knowledge Distillation}}. In \bibinfo{booktitle}{{\em Conference on Neural
  Information Processing Systems}}.
\newblock


\bibitem[\protect\citeauthoryear{Chen, Wilson, Tyree, Weinberger, and
  Chen}{Chen et~al\mbox{.}}{2015b}]%
        {PRU_CNN_HASHEDNET}
\bibfield{author}{\bibinfo{person}{Wenlin Chen}, \bibinfo{person}{James
  Wilson}, \bibinfo{person}{Stephen Tyree}, \bibinfo{person}{Kilian
  Weinberger}, {and} \bibinfo{person}{Yixin Chen}.}
  \bibinfo{year}{2015}\natexlab{b}.
\newblock \showarticletitle{{Compressing Neural Networks with the Hashing
  Trick}}. In \bibinfo{booktitle}{{\em International Conference on Machine
  Learning}}.
\newblock


\bibitem[\protect\citeauthoryear{Chen, Luo, Liu, Zhang, He, Wang, Li, Chen, Xu,
  and Sun}{Chen et~al\mbox{.}}{2014}]%
        {FXP_CNN_DADIANNAO}
\bibfield{author}{\bibinfo{person}{Yunji Chen}, \bibinfo{person}{Tao Luo},
  \bibinfo{person}{Shaoli Liu}, \bibinfo{person}{Shijin Zhang},
  \bibinfo{person}{Liqiang He}, \bibinfo{person}{Jia Wang},
  \bibinfo{person}{Ling Li}, \bibinfo{person}{Tianshi Chen},
  \bibinfo{person}{Zhiwei Xu}, {and} \bibinfo{person}{Ninghui Sun}.}
  \bibinfo{year}{2014}\natexlab{}.
\newblock \showarticletitle{{DaDianNao: A Machine-learning Supercomputer}}. In
  \bibinfo{booktitle}{{\em IEEE/ACM International Symposium on
  Microarchitecture}}.
\newblock


\bibitem[\protect\citeauthoryear{Chen, Krishna, Emer, and Sze}{Chen
  et~al\mbox{.}}{2017b}]%
        {PRU_CNN_EYERISS}
\bibfield{author}{\bibinfo{person}{Yu-Hsin Chen}, \bibinfo{person}{Tushar
  Krishna}, \bibinfo{person}{Joel~S. Emer}, {and} \bibinfo{person}{Vivienne
  Sze}.} \bibinfo{year}{2017}\natexlab{b}.
\newblock \showarticletitle{{Eyeriss: An Energy-efficient Reconfigurable
  Accelerator for Deep Convolutional Neural Networks}}.
\newblock \bibinfo{journal}{{\em IEEE Journal of Solid-state Circuits\/}}
  \bibinfo{volume}{{52}, 1} (\bibinfo{year}{2017}).
\newblock


\bibitem[\protect\citeauthoryear{Cheng, Wang, Li, Hu, and Lu}{Cheng
  et~al\mbox{.}}{2018a}]%
        {SURV_CNN_FPGA}
\bibfield{author}{\bibinfo{person}{Jian Cheng}, \bibinfo{person}{Peisong Wang},
  \bibinfo{person}{Gang Li}, \bibinfo{person}{Qinghao Hu}, {and}
  \bibinfo{person}{Hanqing Lu}.} \bibinfo{year}{2018}\natexlab{a}.
\newblock \showarticletitle{{Recent Advances in Efficient Computation of Deep
  Convolutional Neural Networks}}.
\newblock \bibinfo{journal}{{\em Frontiers of Information Technology \&
  Electronic Engineering\/}} \bibinfo{volume}{{19}, 1} (\bibinfo{year}{2018}).
\newblock


\bibitem[\protect\citeauthoryear{Cheng, Wang, Zhou, and Zhang}{Cheng
  et~al\mbox{.}}{2018b}]%
        {SURV_CNN_ALGO}
\bibfield{author}{\bibinfo{person}{Yu Cheng}, \bibinfo{person}{Duo Wang},
  \bibinfo{person}{Pan Zhou}, {and} \bibinfo{person}{Tao Zhang}.}
  \bibinfo{year}{2018}\natexlab{b}.
\newblock \showarticletitle{{Model Compression and Acceleration for Deep Neural
  Networks: The Principles, Progress, and Challenges}}.
\newblock \bibinfo{journal}{{\em IEEE Signal Processing Magazine\/}}
  \bibinfo{volume}{{35}, 1} (\bibinfo{year}{2018}).
\newblock


\bibitem[\protect\citeauthoryear{Cheng, Yu, Feris, Kumar, Choudhary, and
  Chang}{Cheng et~al\mbox{.}}{2015a}]%
        {SM_CNN_CIRCULANT_PROJECTIONS}
\bibfield{author}{\bibinfo{person}{Yu Cheng}, \bibinfo{person}{Felix~X. Yu},
  \bibinfo{person}{Rogerio~S. Feris}, \bibinfo{person}{Sanjiv Kumar},
  \bibinfo{person}{Alok Choudhary}, {and} \bibinfo{person}{Shi-Fu Chang}.}
  \bibinfo{year}{2015}\natexlab{a}.
\newblock \showarticletitle{{An Exploration of Parameter Redundancy in Deep
  Networks with Circulant Projections}}. In \bibinfo{booktitle}{{\em
  International Conference on Computer Vision}}.
\newblock


\bibitem[\protect\citeauthoryear{Cheng, Yu, Feris, Kumar, Choudhary, and
  Chang}{Cheng et~al\mbox{.}}{2015b}]%
        {SM_CNN_CIRCULANT_PROJECTIONS_0}
\bibfield{author}{\bibinfo{person}{Yu Cheng}, \bibinfo{person}{Felix~X. Yu},
  \bibinfo{person}{Rogerio~S. Feris}, \bibinfo{person}{Sanjiv Kumar},
  \bibinfo{person}{Alok Choudhary}, {and} \bibinfo{person}{Shih-Fu Chang}.}
  \bibinfo{year}{2015}\natexlab{b}.
\newblock \showarticletitle{{Fast Neural Networks with Circulant Projections}}.
\newblock \bibinfo{journal}{{\em arXiv preprint arXiv:1502.03436\/}}
  (\bibinfo{year}{2015}).
\newblock


\bibitem[\protect\citeauthoryear{Chung, Fowers, Ovtcharov, Papamichael,
  Caulfield, Massengil, Liu, Lo, Alkalay, Haselman, Boehn, Firestein, Forin,
  Gatlin, Ghandi, Heil, Holohan, Juhasz, Kovvuri, Lanka, van Megen, Mukhortov,
  Patel, Reinhardt, Sapek, Seera, Sridharan, Woods, Yi-Xiao, Zhao, and
  Burger}{Chung et~al\mbox{.}}{2017}]%
        {MICROSOFT_BRAINWAVE}
\bibfield{author}{\bibinfo{person}{Eric Chung}, \bibinfo{person}{Jeremy
  Fowers}, \bibinfo{person}{Kalin Ovtcharov}, \bibinfo{person}{Michael
  Papamichael}, \bibinfo{person}{Adrian Caulfield}, \bibinfo{person}{Todd
  Massengil}, \bibinfo{person}{Ming Liu}, \bibinfo{person}{Daniel Lo},
  \bibinfo{person}{Shlomi Alkalay}, \bibinfo{person}{Michael Haselman},
  \bibinfo{person}{Christian Boehn}, \bibinfo{person}{Oren Firestein},
  \bibinfo{person}{Alessandro Forin}, \bibinfo{person}{Kang~S. Gatlin},
  \bibinfo{person}{Mahdi Ghandi}, \bibinfo{person}{Stephen Heil},
  \bibinfo{person}{Kyle Holohan}, \bibinfo{person}{Tamas Juhasz},
  \bibinfo{person}{Ratna~K. Kovvuri}, \bibinfo{person}{Sitaram Lanka},
  \bibinfo{person}{Friedel van Megen}, \bibinfo{person}{Dima Mukhortov},
  \bibinfo{person}{Prerak Patel}, \bibinfo{person}{Steve Reinhardt},
  \bibinfo{person}{Adam Sapek}, \bibinfo{person}{Raja Seera},
  \bibinfo{person}{Balaji Sridharan}, \bibinfo{person}{Lisa Woods},
  \bibinfo{person}{Phillip Yi-Xiao}, \bibinfo{person}{Ritchie Zhao}, {and}
  \bibinfo{person}{Doug Burger}.} \bibinfo{year}{2017}\natexlab{}.
\newblock \showarticletitle{{Accelerating Persistent Neural Networks at
  Datacenter Scale}}. In \bibinfo{booktitle}{{\em Hot Chips}}.
\newblock


\bibitem[\protect\citeauthoryear{Colangelo, Nasiri, Nurvitadhi, Mishra,
  Margala, and Nealis}{Colangelo et~al\mbox{.}}{2018}]%
        {FXP_CNN_INTEL_FCCM}
\bibfield{author}{\bibinfo{person}{Philip Colangelo}, \bibinfo{person}{Nasibeh
  Nasiri}, \bibinfo{person}{Eriko Nurvitadhi}, \bibinfo{person}{Asit Mishra},
  \bibinfo{person}{Martin Margala}, {and} \bibinfo{person}{Kevin Nealis}.}
  \bibinfo{year}{2018}\natexlab{}.
\newblock \showarticletitle{{Exploration of Low Numerical Precision Deep
  Learning Inference Using Intel FPGAs}}. In \bibinfo{booktitle}{{\em
  International Symposium on Field-programmable Custom Computing Machines}}.
\newblock


\bibitem[\protect\citeauthoryear{Courbariaux and Bengio}{Courbariaux and
  Bengio}{2016}]%
        {BNN_CNN_BinaryNet}
\bibfield{author}{\bibinfo{person}{Matthieu Courbariaux} {and}
  \bibinfo{person}{Yoshua Bengio}.} \bibinfo{year}{2016}\natexlab{}.
\newblock \showarticletitle{{BinaryNet: Training Deep Neural Networks with
  Weights and Activations Constrained to +1 or -1}}.
\newblock \bibinfo{journal}{{\em arXiv preprint arXiv:1602.02830\/}}
  (\bibinfo{year}{2016}).
\newblock


\bibitem[\protect\citeauthoryear{Courbariaux, Bengio, and David}{Courbariaux
  et~al\mbox{.}}{2015a}]%
        {BNN_CNN_BinaryConnect}
\bibfield{author}{\bibinfo{person}{Matthieu Courbariaux},
  \bibinfo{person}{Yoshua Bengio}, {and} \bibinfo{person}{Jean-Pierre David}.}
  \bibinfo{year}{2015}\natexlab{a}.
\newblock \showarticletitle{{BinaryConnect: Training Deep Neural Networks with
  Binary Weights During Propagations}}. In \bibinfo{booktitle}{{\em Conference
  on Neural Information Processing Systems}}.
\newblock


\bibitem[\protect\citeauthoryear{Courbariaux, David, and Bengio}{Courbariaux
  et~al\mbox{.}}{2015b}]%
        {R_CNN_FXP_DFXP}
\bibfield{author}{\bibinfo{person}{Matthieu Courbariaux},
  \bibinfo{person}{Jean-Pierre David}, {and} \bibinfo{person}{Yoshua Bengio}.}
  \bibinfo{year}{2015}\natexlab{b}.
\newblock \showarticletitle{{Low Precision Storage for Deep Learning}}. In
  \bibinfo{booktitle}{{\em International Conference on Learning
  Representations}}.
\newblock


\bibitem[\protect\citeauthoryear{De~Silva and Lim}{De~Silva and Lim}{2006}]%
        {LRF_SVD_CP_OPTIMISER_ILLPOSED}
\bibfield{author}{\bibinfo{person}{Vin De~Silva} {and}
  \bibinfo{person}{Lek-Heng Lim}.} \bibinfo{year}{2006}\natexlab{}.
\newblock \showarticletitle{{Tensor Rank and the Ill-posedness of the Best
  Low-rank Approximation Problem}}.
\newblock \bibinfo{journal}{{\em {SIAM Journal on Matrix Analysis and
  Applications}\/}} \bibinfo{volume}{{30}, 3} (\bibinfo{year}{2006}).
\newblock


\bibitem[\protect\citeauthoryear{Deng, Yin, and Zhang}{Deng
  et~al\mbox{.}}{2013}]%
        {GROUP_SPARSITY_REGULARISER}
\bibfield{author}{\bibinfo{person}{Wei Deng}, \bibinfo{person}{Wotao Yin},
  {and} \bibinfo{person}{Yin Zhang}.} \bibinfo{year}{2013}\natexlab{}.
\newblock \showarticletitle{{Group Sparse Optimization by Alternating Direction
  Method}}. In \bibinfo{booktitle}{{\em International Society for Optical
  Engineering}}.
\newblock


\bibitem[\protect\citeauthoryear{Denil, Shakibi, Dinh, and De~Freitas}{Denil
  et~al\mbox{.}}{2013}]%
        {LRF_PREDICTING_PARAMETERS}
\bibfield{author}{\bibinfo{person}{Misha Denil}, \bibinfo{person}{Babak
  Shakibi}, \bibinfo{person}{Laurent Dinh}, {and} \bibinfo{person}{Nando
  De~Freitas}.} \bibinfo{year}{2013}\natexlab{}.
\newblock \showarticletitle{{Predicting Parameters in Deep Learning}}. In
  \bibinfo{booktitle}{{\em Conference on Neural Information Processing
  Systems}}.
\newblock


\bibitem[\protect\citeauthoryear{Denton, Zaremba, Bruna, LeCun, and
  Fergus}{Denton et~al\mbox{.}}{2014}]%
        {LRF_SVD_NIPS}
\bibfield{author}{\bibinfo{person}{Emily~L. Denton}, \bibinfo{person}{Wojciech
  Zaremba}, \bibinfo{person}{Joan Bruna}, \bibinfo{person}{Yann LeCun}, {and}
  \bibinfo{person}{Rob Fergus}.} \bibinfo{year}{2014}\natexlab{}.
\newblock \showarticletitle{{Exploiting Linear Structure within Convolutional
  Networks for Efficient Evaluation}}. In \bibinfo{booktitle}{{\em Conference
  on Neural Information Processing Systems}}.
\newblock


\bibitem[\protect\citeauthoryear{Ding, Liao, Wang, Li, Liu, Zhuo, Wang, Qian,
  Bai, and Yuan}{Ding et~al\mbox{.}}{2017}]%
        {SM_CNN_FPGA_CIRCNN}
\bibfield{author}{\bibinfo{person}{Caiwen Ding}, \bibinfo{person}{Siyu Liao},
  \bibinfo{person}{Yanzhi Wang}, \bibinfo{person}{Zhe Li},
  \bibinfo{person}{Ning Liu}, \bibinfo{person}{Youwei Zhuo},
  \bibinfo{person}{Chao Wang}, \bibinfo{person}{Xuehai Qian},
  \bibinfo{person}{Yu Bai}, {and} \bibinfo{person}{Geng Yuan}.}
  \bibinfo{year}{2017}\natexlab{}.
\newblock \showarticletitle{{CirCNN: Accelerating and Compressing Deep Neural
  Networks Using Block-circulant Weight Matrices}}. In \bibinfo{booktitle}{{\em
  IEEE/ACM International Symposium on Microarchitecture}}.
\newblock


\bibitem[\protect\citeauthoryear{Ding, Ren, Yuan, Ma, Li, Liu, Yuan, and
  Wang}{Ding et~al\mbox{.}}{2018}]%
        {SM_FPGA_ASIC_BLOCK_CIRCULANT}
\bibfield{author}{\bibinfo{person}{Caiwen Ding}, \bibinfo{person}{Ao Ren},
  \bibinfo{person}{Geng Yuan}, \bibinfo{person}{Xiaolong Ma},
  \bibinfo{person}{Jiayu Li}, \bibinfo{person}{Ning Liu}, \bibinfo{person}{Bo
  Yuan}, {and} \bibinfo{person}{Yanzhi Wang}.} \bibinfo{year}{2018}\natexlab{}.
\newblock \showarticletitle{{Structured Weight Matrices-based Hardware
  Accelerators in Deep Neural Networks: FPGAs and ASICs}}.
\newblock \bibinfo{journal}{{\em arXiv preprint arXiv:1804.11239\/}}
  (\bibinfo{year}{2018}).
\newblock


\bibitem[\protect\citeauthoryear{Duch and Jankowski}{Duch and
  Jankowski}{1999}]%
        {ACTIVATION_FUNCTIONS}
\bibfield{author}{\bibinfo{person}{Wlodzislaw Duch} {and}
  \bibinfo{person}{Norbert Jankowski}.} \bibinfo{year}{1999}\natexlab{}.
\newblock \showarticletitle{{Survey of Neural Transfer Functions}}.
\newblock \bibinfo{journal}{{\em Neural Computing Surveys\/}}
  \bibinfo{volume}{{2}, 1} (\bibinfo{year}{1999}).
\newblock


\bibitem[\protect\citeauthoryear{Farabet, Martini, Corda, Akselrod,
  Culurciello, and LeCun}{Farabet et~al\mbox{.}}{2011}]%
        {NEUFLOW}
\bibfield{author}{\bibinfo{person}{Cl{\'e}ment Farabet}, \bibinfo{person}{Berin
  Martini}, \bibinfo{person}{Benoit Corda}, \bibinfo{person}{Polina Akselrod},
  \bibinfo{person}{Eugenio Culurciello}, {and} \bibinfo{person}{Yann LeCun}.}
  \bibinfo{year}{2011}\natexlab{}.
\newblock \showarticletitle{{NeuFlow: A Runtime Reconfigurable Dataflow
  Processor for Vision}}. In \bibinfo{booktitle}{{\em IEEE Computer Society
  Computer Vision and Pattern Recognition Workshops}}.
\newblock


\bibitem[\protect\citeauthoryear{Fox, Boland, and Leong}{Fox
  et~al\mbox{.}}{2018}]%
        {SM_FPGA_FASTFOOD}
\bibfield{author}{\bibinfo{person}{Sean Fox}, \bibinfo{person}{David Boland},
  {and} \bibinfo{person}{Philip H.~W. Leong}.} \bibinfo{year}{2018}\natexlab{}.
\newblock \showarticletitle{{FPGA Fastfood -- A High Speed Systolic
  Implementation of a Large Scale Online Kernel Method}}. In
  \bibinfo{booktitle}{{\em ACM/SIGDA International Symposium on
  Field-programmable Gate Arrays}}.
\newblock


\bibitem[\protect\citeauthoryear{Gandhi, Pinto, and Gupta}{Gandhi
  et~al\mbox{.}}{2017}]%
        {BG_DRONE_1}
\bibfield{author}{\bibinfo{person}{Dhiraj Gandhi}, \bibinfo{person}{Lerrel
  Pinto}, {and} \bibinfo{person}{Abhinav Gupta}.}
  \bibinfo{year}{2017}\natexlab{}.
\newblock \showarticletitle{{Learning to Fly by Crashing}}. In
  \bibinfo{booktitle}{{\em IEEE/RSJ International Conference on Intelligent
  Robots and Systems}}.
\newblock


\bibitem[\protect\citeauthoryear{Gao, Neil, Ceolini, Liu, and Delbruck}{Gao
  et~al\mbox{.}}{2018}]%
        {DR_GRU_FPGA_DeltaRNN}
\bibfield{author}{\bibinfo{person}{Chang Gao}, \bibinfo{person}{Daniel Neil},
  \bibinfo{person}{Enea Ceolini}, \bibinfo{person}{Shih-Chii Liu}, {and}
  \bibinfo{person}{Tobi Delbruck}.} \bibinfo{year}{2018}\natexlab{}.
\newblock \showarticletitle{{DeltaRNN: A Power-efficient Recurrent Neural
  Network Accelerator}}. In \bibinfo{booktitle}{{\em ACM/SIGDA International
  Symposium on Field-programmable Gate Arrays}}.
\newblock


\bibitem[\protect\citeauthoryear{Ghasemzadeh, Samragh, and
  Koushanfar}{Ghasemzadeh et~al\mbox{.}}{2018}]%
        {BNN_CNN_REBNET_FCCM}
\bibfield{author}{\bibinfo{person}{Mohammad Ghasemzadeh},
  \bibinfo{person}{Mohammad Samragh}, {and} \bibinfo{person}{Farinaz
  Koushanfar}.} \bibinfo{year}{2018}\natexlab{}.
\newblock \showarticletitle{{ReBNet: Residual Binarized Neural Network}}. In
  \bibinfo{booktitle}{{\em IEEE International Symposium on Field-programmable
  Custom Computing Machines}}.
\newblock


\bibitem[\protect\citeauthoryear{Gray}{Gray}{2006}]%
        {TOEPLITZ_CIRCULAR_REVIEW}
\bibfield{author}{\bibinfo{person}{Robert~M. Gray}.}
  \bibinfo{year}{2006}\natexlab{}.
\newblock \showarticletitle{{Toeplitz and Circulant Matrices: A Review}}.
\newblock \bibinfo{journal}{{\em Foundations and Trends in Communications and
  Information Theory\/}} \bibinfo{volume}{{2}, 3} (\bibinfo{year}{2006}).
\newblock


\bibitem[\protect\citeauthoryear{Guan, Yuan, Sun, and Cong}{Guan
  et~al\mbox{.}}{2017}]%
        {PLAN_LSTM_FPGA}
\bibfield{author}{\bibinfo{person}{Yijin Guan}, \bibinfo{person}{Zhihang Yuan},
  \bibinfo{person}{Guangyu Sun}, {and} \bibinfo{person}{Jason Cong}.}
  \bibinfo{year}{2017}\natexlab{}.
\newblock \showarticletitle{{FPGA-based Accelerator for Long Short-term Memory
  Recurrent Neural Networks}}. In \bibinfo{booktitle}{{\em Asia and South
  Pacific Design Automation Conference}}.
\newblock


\bibitem[\protect\citeauthoryear{Gudovskiy and Rigazio}{Gudovskiy and
  Rigazio}{2017}]%
        {EXP_CNN_SHIFTCNN}
\bibfield{author}{\bibinfo{person}{Denis~A. Gudovskiy} {and}
  \bibinfo{person}{Luca Rigazio}.} \bibinfo{year}{2017}\natexlab{}.
\newblock \showarticletitle{{ShiftCNN: Generalized Low-precision Architecture
  for Inference of Convolutional Neural Networks}}.
\newblock \bibinfo{journal}{{\em arXiv preprint arXiv:1706.02393\/}}
  (\bibinfo{year}{2017}).
\newblock


\bibitem[\protect\citeauthoryear{Guo, Sui, Qiu, Yao, Han, Wang, and Yang}{Guo
  et~al\mbox{.}}{2016a}]%
        {NR_CNN_FXP_ANGEL-EYE}
\bibfield{author}{\bibinfo{person}{Kaiyuan Guo}, \bibinfo{person}{Lingzhi Sui},
  \bibinfo{person}{Jiantao Qiu}, \bibinfo{person}{Song Yao},
  \bibinfo{person}{Song Han}, \bibinfo{person}{Yu Wang}, {and}
  \bibinfo{person}{Huazhong Yang}.} \bibinfo{year}{2016}\natexlab{a}.
\newblock \showarticletitle{{Angel-eye: A Complete Design Flow for Mapping CNN
  onto Customized Hardware}}. In \bibinfo{booktitle}{{\em IEEE Computer Society
  Annual Symposium on VLSI}}.
\newblock


\bibitem[\protect\citeauthoryear{Guo, Zeng, Yu, Wang, and Yang}{Guo
  et~al\mbox{.}}{2017}]%
        {SURV_FPGA_BASED_NEURAL_NETWORK_ACC}
\bibfield{author}{\bibinfo{person}{Kaiyuan Guo}, \bibinfo{person}{Shulin Zeng},
  \bibinfo{person}{Jincheng Yu}, \bibinfo{person}{Yu Wang}, {and}
  \bibinfo{person}{Huazhong Yang}.} \bibinfo{year}{2017}\natexlab{}.
\newblock \showarticletitle{{A Survey of FPGA Based Neural Network
  Accelerator}}.
\newblock \bibinfo{journal}{{\em ACM Transactions on Reconfigurable Technology
  and Systems\/}} \bibinfo{volume}{{9}, 4} (\bibinfo{year}{2017}).
\newblock


\bibitem[\protect\citeauthoryear{Guo, Yao, and Chen}{Guo
  et~al\mbox{.}}{2016b}]%
        {PRU_CNN_DNS}
\bibfield{author}{\bibinfo{person}{Yiwen Guo}, \bibinfo{person}{Anbang Yao},
  {and} \bibinfo{person}{Yurong Chen}.} \bibinfo{year}{2016}\natexlab{b}.
\newblock \showarticletitle{{Dynamic Network Surgery for Efficient DNNs}}. In
  \bibinfo{booktitle}{{\em Conference on Neural Information Processing
  Systems}}.
\newblock


\bibitem[\protect\citeauthoryear{Gupta, Agrawal, Gopalakrishnan, and
  Narayanan}{Gupta et~al\mbox{.}}{2015}]%
        {R_CNN_FXP_STOCH}
\bibfield{author}{\bibinfo{person}{Suyog Gupta}, \bibinfo{person}{Ankur
  Agrawal}, \bibinfo{person}{Kailash Gopalakrishnan}, {and}
  \bibinfo{person}{Pritish Narayanan}.} \bibinfo{year}{2015}\natexlab{}.
\newblock \showarticletitle{{Deep Learning with Limited Numerical Precision}}.
  In \bibinfo{booktitle}{{\em International Conference on Machine Learning}}.
\newblock


\bibitem[\protect\citeauthoryear{Han, Kang, Mao, Hu, Li, Li, Xie, Luo, Yao, and
  Wang}{Han et~al\mbox{.}}{2017}]%
        {R_LSTM_FXP_ESE}
\bibfield{author}{\bibinfo{person}{Song Han}, \bibinfo{person}{Junlong Kang},
  \bibinfo{person}{Huizi Mao}, \bibinfo{person}{Yiming Hu},
  \bibinfo{person}{Xin Li}, \bibinfo{person}{Yubin Li},
  \bibinfo{person}{Dongliang Xie}, \bibinfo{person}{Hong Luo},
  \bibinfo{person}{Song Yao}, {and} \bibinfo{person}{Yu Wang}.}
  \bibinfo{year}{2017}\natexlab{}.
\newblock \showarticletitle{{ESE: Efficient Speech Recognition Engine with
  Sparse LSTM on FPGA}}. In \bibinfo{booktitle}{{\em ACM/SIGDA International
  Symposium on Field-programmable Gate Arrays}}.
\newblock


\bibitem[\protect\citeauthoryear{Han, Liu, Mao, Pu, Pedram, Horowitz, and
  Dally}{Han et~al\mbox{.}}{2016a}]%
        {PRU_CNN_EIE}
\bibfield{author}{\bibinfo{person}{Song Han}, \bibinfo{person}{Xingyu Liu},
  \bibinfo{person}{Huizi Mao}, \bibinfo{person}{Jing Pu},
  \bibinfo{person}{Ardavan Pedram}, \bibinfo{person}{Mark~A. Horowitz}, {and}
  \bibinfo{person}{William~J. Dally}.} \bibinfo{year}{2016}\natexlab{a}.
\newblock \showarticletitle{{EIE: Efficient Inference Engine on Compressed Deep
  Neural Network}}. In \bibinfo{booktitle}{{\em ACM/IEEE International
  Symposium on Computer Architecture}}.
\newblock


\bibitem[\protect\citeauthoryear{Han, Mao, and Dally}{Han
  et~al\mbox{.}}{2016b}]%
        {PRU_CNN_DEEP_COMPRESSION}
\bibfield{author}{\bibinfo{person}{Song Han}, \bibinfo{person}{Huizi Mao},
  {and} \bibinfo{person}{William~J. Dally}.} \bibinfo{year}{2016}\natexlab{b}.
\newblock \showarticletitle{{Deep Compression: Compressing Deep Neural Networks
  with Pruning, Trained Quantization and Huffman Coding}}. In
  \bibinfo{booktitle}{{\em International Conference on Learning
  Representations}}.
\newblock


\bibitem[\protect\citeauthoryear{Han, Pool, Tran, and Dally}{Han
  et~al\mbox{.}}{2015}]%
        {PRU_CNN_TRAIN_PRUNE_RETRAIN}
\bibfield{author}{\bibinfo{person}{Song Han}, \bibinfo{person}{Jeff Pool},
  \bibinfo{person}{John Tran}, {and} \bibinfo{person}{William~J. Dally}.}
  \bibinfo{year}{2015}\natexlab{}.
\newblock \showarticletitle{{Learning Both Weights and Connections for
  Efficient Neural Network}}. In \bibinfo{booktitle}{{\em Conference on Neural
  Information Processing Systems}}.
\newblock


\bibitem[\protect\citeauthoryear{Hassibi and Stork}{Hassibi and Stork}{1993}]%
        {PRU_CNN_OPTIMAL_BRAIN_SURGEON}
\bibfield{author}{\bibinfo{person}{Babak Hassibi} {and}
  \bibinfo{person}{David~G. Stork}.} \bibinfo{year}{1993}\natexlab{}.
\newblock \showarticletitle{{Second Order Derivatives for Network Pruning:
  Optimal Brain Surgeon}}. In \bibinfo{booktitle}{{\em Conference on Neural
  Information Processing Systems}}.
\newblock


\bibitem[\protect\citeauthoryear{He, Zhang, and Sun}{He et~al\mbox{.}}{2017}]%
        {PRU_CNN_CHANNEL_PRUNING}
\bibfield{author}{\bibinfo{person}{Yihui He}, \bibinfo{person}{Xiangyu Zhang},
  {and} \bibinfo{person}{Jian Sun}.} \bibinfo{year}{2017}\natexlab{}.
\newblock \showarticletitle{{Channel Pruning for Accelerating Very Deep Neural
  Networks}}. In \bibinfo{booktitle}{{\em International Conference on Computer
  Vision}}.
\newblock


\bibitem[\protect\citeauthoryear{Hegde and Kapre}{Hegde and Kapre}{2018}]%
        {CAFFEPRESSO}
\bibfield{author}{\bibinfo{person}{Gopalakrishna Hegde} {and}
  \bibinfo{person}{Nachiket Kapre}.} \bibinfo{year}{2018}\natexlab{}.
\newblock \showarticletitle{{CaffePresso: Accelerating Convolutional Networks
  on Embedded SoCs}}.
\newblock \bibinfo{journal}{{\em ACM Transactions on Embedded Computing
  Systems\/}} \bibinfo{volume}{{17}, 1} (\bibinfo{year}{2018}).
\newblock


\bibitem[\protect\citeauthoryear{Hinton, Vinyals, and Dean}{Hinton
  et~al\mbox{.}}{2015}]%
        {KD_HINTON}
\bibfield{author}{\bibinfo{person}{Geoffrey Hinton}, \bibinfo{person}{Oriol
  Vinyals}, {and} \bibinfo{person}{Jeff Dean}.}
  \bibinfo{year}{2015}\natexlab{}.
\newblock \showarticletitle{{Distilling the Knowledge in a Neural Network}}.
\newblock \bibinfo{journal}{{\em arXiv preprint arXiv:1503.02531\/}}
  (\bibinfo{year}{2015}).
\newblock


\bibitem[\protect\citeauthoryear{Howard, Zhu, Chen, Kalenichenko, Wang, Weyand,
  Andreetto, and Adam}{Howard et~al\mbox{.}}{2017}]%
        {MOBILENET}
\bibfield{author}{\bibinfo{person}{Andrew~G. Howard}, \bibinfo{person}{Menglong
  Zhu}, \bibinfo{person}{Bo Chen}, \bibinfo{person}{Dmitry Kalenichenko},
  \bibinfo{person}{Weijun Wang}, \bibinfo{person}{Tobias Weyand},
  \bibinfo{person}{Marco Andreetto}, {and} \bibinfo{person}{Hartwig Adam}.}
  \bibinfo{year}{2017}\natexlab{}.
\newblock \showarticletitle{{MobileNets: Efficient Convolutional Neural
  Networks for Mobile Vision Applications}}.
\newblock \bibinfo{journal}{{\em arXiv preprint arXiv:1704.04861\/}}
  (\bibinfo{year}{2017}).
\newblock


\bibitem[\protect\citeauthoryear{Intel}{Intel}{2018}]%
        {INTEL_CASCADE_LAKE}
\bibfield{author}{\bibinfo{person}{Intel}.} \bibinfo{year}{2018}\natexlab{}.
\newblock \bibinfo{title}{{Intel at Hot Chips 2018: Showing the Ankle of
  Cascade Lake}}.
\newblock   (\bibinfo{year}{2018}).
\newblock
\showURL{%
\url{https://www.anandtech.com/show/13239/intel-at-hot-chips-2018-showing-the-ankle-of-cascade-lake}}


\bibitem[\protect\citeauthoryear{Jacob, Kligys, Chen, Zhu, Tang, Howard, Adam,
  and Kalenichenko}{Jacob et~al\mbox{.}}{2017}]%
        {FXP_CNN_GOOGLE_MOBILENET}
\bibfield{author}{\bibinfo{person}{Benoit Jacob}, \bibinfo{person}{Skirmantas
  Kligys}, \bibinfo{person}{Bo Chen}, \bibinfo{person}{Menglong Zhu},
  \bibinfo{person}{Matthew Tang}, \bibinfo{person}{Andrew Howard},
  \bibinfo{person}{Hartwig Adam}, {and} \bibinfo{person}{Dmitry Kalenichenko}.}
  \bibinfo{year}{2017}\natexlab{}.
\newblock \showarticletitle{{Quantization and Training of Neural Networks for
  Efficient Integer-arithmetic-only Inference}}.
\newblock \bibinfo{journal}{{\em arXiv preprint arXiv:1712.05877\/}}
  (\bibinfo{year}{2017}).
\newblock


\bibitem[\protect\citeauthoryear{Jaderberg, Vedaldi, and Zisserman}{Jaderberg
  et~al\mbox{.}}{2014}]%
        {LRF_LINEAR_COMBINATIONS}
\bibfield{author}{\bibinfo{person}{Max Jaderberg}, \bibinfo{person}{Andrea
  Vedaldi}, {and} \bibinfo{person}{Andrew Zisserman}.}
  \bibinfo{year}{2014}\natexlab{}.
\newblock \showarticletitle{{Speeding up Convolutional Neural Networks with Low
  Rank Expansions}}. In \bibinfo{booktitle}{{\em British Machine Vision
  Conference}}.
\newblock


\bibitem[\protect\citeauthoryear{Jegou, Douze, and Schmid}{Jegou
  et~al\mbox{.}}{2011}]%
        {PRODUCT_QUANTISATION}
\bibfield{author}{\bibinfo{person}{Herve Jegou}, \bibinfo{person}{Matthijs
  Douze}, {and} \bibinfo{person}{Cordelia Schmid}.}
  \bibinfo{year}{2011}\natexlab{}.
\newblock \showarticletitle{{Product Quantization for Nearest Neighbor
  Search}}.
\newblock \bibinfo{journal}{{\em IEEE Transactions on Pattern Analysis and
  Machine Intelligence\/}} \bibinfo{volume}{{33}, 1} (\bibinfo{year}{2011}).
\newblock


\bibitem[\protect\citeauthoryear{Jouppi, Young, Patil, and Patterson}{Jouppi
  et~al\mbox{.}}{2018}]%
        {FXP_CNN_TPU_JOURNAL}
\bibfield{author}{\bibinfo{person}{Norman~P. Jouppi}, \bibinfo{person}{Cliff
  Young}, \bibinfo{person}{Nishant Patil}, {and} \bibinfo{person}{David
  Patterson}.} \bibinfo{year}{2018}\natexlab{}.
\newblock \showarticletitle{{A Domain-specific Architecture for Deep Neural
  Networks}}.
\newblock \bibinfo{journal}{{\em {Communications of the ACM}\/}}
  \bibinfo{volume}{{61}, 9} (\bibinfo{year}{2018}).
\newblock


\bibitem[\protect\citeauthoryear{Jouppi, Young, Patil, Patterson, Agrawal,
  Bajwa, Bates, Bhatia, Boden, and Borchers}{Jouppi et~al\mbox{.}}{2017}]%
        {FXP_CNN_TPU}
\bibfield{author}{\bibinfo{person}{Norman~P. Jouppi}, \bibinfo{person}{Cliff
  Young}, \bibinfo{person}{Nishant Patil}, \bibinfo{person}{David Patterson},
  \bibinfo{person}{Gaurav Agrawal}, \bibinfo{person}{Raminder Bajwa},
  \bibinfo{person}{Sarah Bates}, \bibinfo{person}{Suresh Bhatia},
  \bibinfo{person}{Nan Boden}, {and} \bibinfo{person}{Al Borchers}.}
  \bibinfo{year}{2017}\natexlab{}.
\newblock \showarticletitle{{In-datacenter Performance Analysis of a Tensor
  Processing Unit}}. In \bibinfo{booktitle}{{\em International Symposium on
  Computer Architecture}}.
\newblock


\bibitem[\protect\citeauthoryear{Judd, Albericio, Hetherington, Aamodt, and
  Moshovos}{Judd et~al\mbox{.}}{2016}]%
        {FXP_CNN_STRIPES}
\bibfield{author}{\bibinfo{person}{Patrick Judd}, \bibinfo{person}{Jorge
  Albericio}, \bibinfo{person}{Tayler Hetherington}, \bibinfo{person}{Tor~M.
  Aamodt}, {and} \bibinfo{person}{Andreas Moshovos}.}
  \bibinfo{year}{2016}\natexlab{}.
\newblock \showarticletitle{{Stripes: Bit-serial Deep Neural Network
  Computing}}. In \bibinfo{booktitle}{{\em IEEE/ACM International Symposium on
  Microarchitecture}}.
\newblock


\bibitem[\protect\citeauthoryear{Karpathy, Toderici, Shetty, Leung, Sukthankar,
  and Fei-Fei}{Karpathy et~al\mbox{.}}{2014}]%
        {DR_CNN_VIDEO}
\bibfield{author}{\bibinfo{person}{Andrej Karpathy}, \bibinfo{person}{George
  Toderici}, \bibinfo{person}{Sanketh Shetty}, \bibinfo{person}{Thomas Leung},
  \bibinfo{person}{Rahul Sukthankar}, {and} \bibinfo{person}{Li Fei-Fei}.}
  \bibinfo{year}{2014}\natexlab{}.
\newblock \showarticletitle{{Large-scale Video Classification with
  Convolutional Neural Networks}}. In \bibinfo{booktitle}{{\em International
  Conference on Computer Vision}}.
\newblock


\bibitem[\protect\citeauthoryear{Khoram and Li}{Khoram and Li}{2018}]%
        {R_CNN_ADAPTIVE_QUANTISATION}
\bibfield{author}{\bibinfo{person}{Soroosh Khoram} {and} \bibinfo{person}{Jing
  Li}.} \bibinfo{year}{2018}\natexlab{}.
\newblock \showarticletitle{{Adaptive Quantization of Neural Networks}}. In
  \bibinfo{booktitle}{{\em International Conference on Learning
  Representations}}.
\newblock


\bibitem[\protect\citeauthoryear{K{\"o}ster, Webb, Wang, Nassar, Bansal,
  Constable, Elibol, Gray, Hall, Hornof, Khosrowshahi, Carey, Pai, and
  Rao}{K{\"o}ster et~al\mbox{.}}{2017}]%
        {INTEL_FLEXPOINT}
\bibfield{author}{\bibinfo{person}{Urs K{\"o}ster}, \bibinfo{person}{Tristan
  Webb}, \bibinfo{person}{Xin Wang}, \bibinfo{person}{Marcel Nassar},
  \bibinfo{person}{Arjun~K. Bansal}, \bibinfo{person}{William Constable},
  \bibinfo{person}{Oguz Elibol}, \bibinfo{person}{Scott Gray},
  \bibinfo{person}{Stewart Hall}, \bibinfo{person}{Luke Hornof},
  \bibinfo{person}{Amir Khosrowshahi}, \bibinfo{person}{Kloss Carey},
  \bibinfo{person}{Ruby~J. Pai}, {and} \bibinfo{person}{Naveen Rao}.}
  \bibinfo{year}{2017}\natexlab{}.
\newblock \showarticletitle{{Flexpoint: An Adaptive Numerical Format for
  Efficient Training of Deep Neural Networks}}. In \bibinfo{booktitle}{{\em
  Conference on Neural Information Processing Systems}}.
\newblock


\bibitem[\protect\citeauthoryear{Kouris, Venieris, and Bouganis}{Kouris
  et~al\mbox{.}}{2018}]%
        {DR_CNN_FPGA_CASCADECNN}
\bibfield{author}{\bibinfo{person}{Alexandros Kouris},
  \bibinfo{person}{Stylianos~I. Venieris}, {and}
  \bibinfo{person}{Christos-Savvas Bouganis}.} \bibinfo{year}{2018}\natexlab{}.
\newblock \showarticletitle{{Cascade CNN: Pushing the Performance Limits of
  Quantisation in Convolutional Neural Networks}}. In \bibinfo{booktitle}{{\em
  International Conference on Field-programmable Logic and Applications}}.
\newblock


\bibitem[\protect\citeauthoryear{Lai, Suda, and Chandra}{Lai
  et~al\mbox{.}}{2017}]%
        {NR_CNN_FXP_ACTIVATIONS}
\bibfield{author}{\bibinfo{person}{Liangzhen Lai}, \bibinfo{person}{Naveen
  Suda}, {and} \bibinfo{person}{Vikas Chandra}.}
  \bibinfo{year}{2017}\natexlab{}.
\newblock \showarticletitle{{Deep Convolutional Neural Network Inference with
  Floating-point Weights and Fixed-point Activations}}. In
  \bibinfo{booktitle}{{\em International Conference on Machine Learning}}.
\newblock


\bibitem[\protect\citeauthoryear{Lebedev, Ganin, Rakhuba, Oseledets, and
  Lempitsky}{Lebedev et~al\mbox{.}}{2015}]%
        {LRF_CP_ICLR}
\bibfield{author}{\bibinfo{person}{Vadim Lebedev}, \bibinfo{person}{Yaroslav
  Ganin}, \bibinfo{person}{Maksim Rakhuba}, \bibinfo{person}{Ivan Oseledets},
  {and} \bibinfo{person}{Victor Lempitsky}.} \bibinfo{year}{2015}\natexlab{}.
\newblock \showarticletitle{{Speeding-up Convolutional Neural Networks Using
  Fine-tuned CP-decomposition}}. In \bibinfo{booktitle}{{\em International
  Conference on Learning Representations}}.
\newblock


\bibitem[\protect\citeauthoryear{Lebedev and Lempitsky}{Lebedev and
  Lempitsky}{2016}]%
        {PRU_CNN_GROUP_WISE_BRAIN_DAMAGE}
\bibfield{author}{\bibinfo{person}{Vadim Lebedev} {and} \bibinfo{person}{Victor
  Lempitsky}.} \bibinfo{year}{2016}\natexlab{}.
\newblock \showarticletitle{{Fast Convnets using Group-wise Brain Damage}}. In
  \bibinfo{booktitle}{{\em IEEE Conference on Computer Vision and Pattern
  Recognition}}.
\newblock


\bibitem[\protect\citeauthoryear{LeCun, Denker, and Solla}{LeCun
  et~al\mbox{.}}{1990}]%
        {PRU_CNN_OPTIMAL_BRAIN_DAMAGE}
\bibfield{author}{\bibinfo{person}{Yann LeCun}, \bibinfo{person}{John~S.
  Denker}, {and} \bibinfo{person}{Sara~A. Solla}.}
  \bibinfo{year}{1990}\natexlab{}.
\newblock \showarticletitle{{Optimal Brain Damage}}. In
  \bibinfo{booktitle}{{\em Conference on Neural Information Processing
  Systems}}.
\newblock


\bibitem[\protect\citeauthoryear{Lee, Miyashita, Chai, Murmann, and Wong}{Lee
  et~al\mbox{.}}{2017}]%
        {EXP_CNN_LOGNET}
\bibfield{author}{\bibinfo{person}{Edward~H. Lee}, \bibinfo{person}{Daisuke
  Miyashita}, \bibinfo{person}{Elaina Chai}, \bibinfo{person}{Boris Murmann},
  {and} \bibinfo{person}{Simon~S. Wong}.} \bibinfo{year}{2017}\natexlab{}.
\newblock \showarticletitle{{LogNet: Energy-efficient Neural Networks Using
  Logarithmic Computation}}. In \bibinfo{booktitle}{{\em IEEE International
  Conference on Acoustics, Speech and Signal Processing}}.
\newblock


\bibitem[\protect\citeauthoryear{Li, Wen, Mao, Li, Chen, and Li}{Li
  et~al\mbox{.}}{2018}]%
        {LRF_CNN_GPU_ASP_DAC}
\bibfield{author}{\bibinfo{person}{Bing Li}, \bibinfo{person}{Wei Wen},
  \bibinfo{person}{Jiachen Mao}, \bibinfo{person}{Sicheng Li},
  \bibinfo{person}{Yiran Chen}, {and} \bibinfo{person}{Hai Li}.}
  \bibinfo{year}{2018}\natexlab{}.
\newblock \showarticletitle{{Running Sparse and Low-precision Neural Network:
  When Algorithm Meets Hardware}}. In \bibinfo{booktitle}{{\em Asia and South
  Pacific Design Automation Conference}}.
\newblock


\bibitem[\protect\citeauthoryear{Li and Liu}{Li and Liu}{2016}]%
        {TNN_CNN_TWN}
\bibfield{author}{\bibinfo{person}{Fengfu Li} {and} \bibinfo{person}{Bin Liu}.}
  \bibinfo{year}{2016}\natexlab{}.
\newblock \showarticletitle{{Ternary Weight Networks}}. In
  \bibinfo{booktitle}{{\em Conference on Neural Information Processing
  Systems}}.
\newblock


\bibitem[\protect\citeauthoryear{Li, De, Xu, Studer, Samet, and Goldstein}{Li
  et~al\mbox{.}}{2017a}]%
        {TRAINING_QUANTIZED_NETS_CONVERGENCE_GUARANTEE}
\bibfield{author}{\bibinfo{person}{Hao Li}, \bibinfo{person}{Soham De},
  \bibinfo{person}{Zheng Xu}, \bibinfo{person}{Christoph Studer},
  \bibinfo{person}{Hanan Samet}, {and} \bibinfo{person}{Tom Goldstein}.}
  \bibinfo{year}{2017}\natexlab{a}.
\newblock \showarticletitle{{Training Quantized Nets: A Deeper Understanding}}.
  In \bibinfo{booktitle}{{\em Conference on Neural Information Processing
  Systems}}.
\newblock


\bibitem[\protect\citeauthoryear{Li, Kadav, Durdanovic, Samet, and Graf}{Li
  et~al\mbox{.}}{2017b}]%
        {PRU_CNN_LASSO_ALONG_FILTERS}
\bibfield{author}{\bibinfo{person}{Hao Li}, \bibinfo{person}{Asim Kadav},
  \bibinfo{person}{Igor Durdanovic}, \bibinfo{person}{Hanan Samet}, {and}
  \bibinfo{person}{Hans~P. Graf}.} \bibinfo{year}{2017}\natexlab{b}.
\newblock \showarticletitle{{Pruning Filters for Efficient Convnets}}. In
  \bibinfo{booktitle}{{\em International Conference on Learning
  Representations}}.
\newblock


\bibitem[\protect\citeauthoryear{Li, Wen, Wang, Han, Chen, and Li}{Li
  et~al\mbox{.}}{2017}]%
        {PRU_FPGA_SPARSIFICATION_FCCM}
\bibfield{author}{\bibinfo{person}{Sicheng Li}, \bibinfo{person}{Wei Wen},
  \bibinfo{person}{Yu Wang}, \bibinfo{person}{Song Han}, \bibinfo{person}{Yiran
  Chen}, {and} \bibinfo{person}{Hai Li}.} \bibinfo{year}{2017}\natexlab{}.
\newblock \showarticletitle{{An FPGA Design Framework for CNN Sparsification
  and Acceleration}}. In \bibinfo{booktitle}{{\em IEEE International Symposium
  on Field-programmable Custom Computing Machines}}.
\newblock


\bibitem[\protect\citeauthoryear{Li, Wu, Li, Li, Wang, and Qiu}{Li
  et~al\mbox{.}}{2015}]%
        {FXP_RNN_MIXED}
\bibfield{author}{\bibinfo{person}{Sicheng Li}, \bibinfo{person}{Chunpeng Wu},
  \bibinfo{person}{Hai Li}, \bibinfo{person}{Boxun Li}, \bibinfo{person}{Yu
  Wang}, {and} \bibinfo{person}{Qinru Qiu}.} \bibinfo{year}{2015}\natexlab{}.
\newblock \showarticletitle{{FPGA Acceleration of Recurrent Neural Network
  Based Language Model}}. In \bibinfo{booktitle}{{\em IEEE International
  Symposium on Field-programmable Custom Computing Machines}}.
\newblock


\bibitem[\protect\citeauthoryear{Liang, Yin, Liu, Luk, and Wei}{Liang
  et~al\mbox{.}}{2018}]%
        {BNN_CNN_FP-BNN}
\bibfield{author}{\bibinfo{person}{Shuang Liang}, \bibinfo{person}{Shouyi Yin},
  \bibinfo{person}{Leibo Liu}, \bibinfo{person}{Wayne Luk}, {and}
  \bibinfo{person}{Shaojun Wei}.} \bibinfo{year}{2018}\natexlab{}.
\newblock \showarticletitle{{FP-BNN: Binarized Neural Network on FPGA}}.
\newblock \bibinfo{journal}{{\em Neurocomputing\/}} \bibinfo{volume}{{275}, C}
  (\bibinfo{year}{2018}).
\newblock


\bibitem[\protect\citeauthoryear{Lin, Talathi, and Annapureddy}{Lin
  et~al\mbox{.}}{2016}]%
        {NR_CNN_FXP_SQNR}
\bibfield{author}{\bibinfo{person}{Darryl Lin}, \bibinfo{person}{Sachin
  Talathi}, {and} \bibinfo{person}{Sreekanth Annapureddy}.}
  \bibinfo{year}{2016}\natexlab{}.
\newblock \showarticletitle{{Fixed Point Quantization of Deep Convolutional
  Networks}}. In \bibinfo{booktitle}{{\em International Conference on Machine
  Learning}}.
\newblock


\bibitem[\protect\citeauthoryear{Lin, Rao, Lu, and Zhou}{Lin
  et~al\mbox{.}}{2017}]%
        {DR_CNN_RUNTIME_PRUNING}
\bibfield{author}{\bibinfo{person}{Ji Lin}, \bibinfo{person}{Yongming Rao},
  \bibinfo{person}{Jiwen Lu}, {and} \bibinfo{person}{Jie Zhou}.}
  \bibinfo{year}{2017}\natexlab{}.
\newblock \showarticletitle{{Runtime Neural Pruning}}. In
  \bibinfo{booktitle}{{\em Conference on Neural Information Processing
  Systems}}.
\newblock


\bibitem[\protect\citeauthoryear{Lin, Zhao, and Pan}{Lin et~al\mbox{.}}{2017}]%
        {BNN_CNN_ABC-Net}
\bibfield{author}{\bibinfo{person}{Xiaofan Lin}, \bibinfo{person}{Cong Zhao},
  {and} \bibinfo{person}{Wei Pan}.} \bibinfo{year}{2017}\natexlab{}.
\newblock \showarticletitle{{Towards Accurate Binary Convolutional Neural
  Network}}. In \bibinfo{booktitle}{{\em Conference on Neural Information
  Processing Systems}}.
\newblock


\bibitem[\protect\citeauthoryear{Lin, Courbariaux, Memisevic, and Bengio}{Lin
  et~al\mbox{.}}{2015}]%
        {TNN_CNN_BENGIO}
\bibfield{author}{\bibinfo{person}{Zhouhan Lin}, \bibinfo{person}{Matthieu
  Courbariaux}, \bibinfo{person}{Roland Memisevic}, {and}
  \bibinfo{person}{Yoshua Bengio}.} \bibinfo{year}{2015}\natexlab{}.
\newblock \showarticletitle{{Neural Networks with Few Multiplications}}. In
  \bibinfo{booktitle}{{\em International Conference on Learning
  Representations}}.
\newblock


\bibitem[\protect\citeauthoryear{Liu, Musialski, Wonka, and Ye}{Liu
  et~al\mbox{.}}{2013}]%
        {TENSOR_LOW_RANK_CONSTRAINTS}
\bibfield{author}{\bibinfo{person}{Ji Liu}, \bibinfo{person}{Przemyslaw
  Musialski}, \bibinfo{person}{Peter Wonka}, {and} \bibinfo{person}{Jieping
  Ye}.} \bibinfo{year}{2013}\natexlab{}.
\newblock \showarticletitle{{Tensor Completion for Estimating Missing Values in
  Visual Data}}.
\newblock \bibinfo{journal}{{\em IEEE Transactions on Pattern Analysis and
  Machine Intelligence\/}} \bibinfo{volume}{{35}, 1} (\bibinfo{year}{2013}).
\newblock


\bibitem[\protect\citeauthoryear{Liu and Deng}{Liu and Deng}{2017}]%
        {DR_CNN_DDNN}
\bibfield{author}{\bibinfo{person}{Lanlan Liu} {and} \bibinfo{person}{Jia
  Deng}.} \bibinfo{year}{2017}\natexlab{}.
\newblock \showarticletitle{{Dynamic Deep Neural Networks: Optimizing
  Accuracy-efficiency Trade-offs by Selective Execution}}.
\newblock \bibinfo{journal}{{\em arXiv preprint arXiv:1701.00299\/}}
  (\bibinfo{year}{2017}).
\newblock


\bibitem[\protect\citeauthoryear{Liu, Anguelov, Erhan, Szegedy, Reed, Fu, and
  Berg}{Liu et~al\mbox{.}}{2016}]%
        {BG_OBJECT_DETECTION_1}
\bibfield{author}{\bibinfo{person}{Wei Liu}, \bibinfo{person}{Dragomir
  Anguelov}, \bibinfo{person}{Dumitru Erhan}, \bibinfo{person}{Christian
  Szegedy}, \bibinfo{person}{Scott Reed}, \bibinfo{person}{Cheng-Yang Fu},
  {and} \bibinfo{person}{Alexander~C. Berg}.} \bibinfo{year}{2016}\natexlab{}.
\newblock \showarticletitle{{SSD: Single Shot Multibox Detector}}. In
  \bibinfo{booktitle}{{\em European Conference on Computer Vision}}.
\newblock


\bibitem[\protect\citeauthoryear{Liu, Cao, and Yu}{Liu et~al\mbox{.}}{2018}]%
        {BNN_LSTM_LM}
\bibfield{author}{\bibinfo{person}{Xuan Liu}, \bibinfo{person}{Di Cao}, {and}
  \bibinfo{person}{Kai Yu}.} \bibinfo{year}{2018}\natexlab{}.
\newblock \showarticletitle{{Binarized LSTM Language Model}}. In
  \bibinfo{booktitle}{{\em Conference of the North American Chapter of the
  Association for Computational Linguistics}}.
\newblock


\bibitem[\protect\citeauthoryear{Liu, Li, Shen, Huang, Yan, and Zhang}{Liu
  et~al\mbox{.}}{2017}]%
        {PRU_CNN_SLIMMING}
\bibfield{author}{\bibinfo{person}{Zhuang Liu}, \bibinfo{person}{Jianguo Li},
  \bibinfo{person}{Zhiqiang Shen}, \bibinfo{person}{Gao Huang},
  \bibinfo{person}{Shoumeng Yan}, {and} \bibinfo{person}{Changshui Zhang}.}
  \bibinfo{year}{2017}\natexlab{}.
\newblock \showarticletitle{{Learning Efficient Convolutional Networks Through
  Network Slimming}}. In \bibinfo{booktitle}{{\em International Conference on
  Computer Vision}}.
\newblock


\bibitem[\protect\citeauthoryear{Lu, Sindhwani, and Sainath}{Lu
  et~al\mbox{.}}{2016}]%
        {SM_LSTM_HYBRID_TOEPLITS_LIKE}
\bibfield{author}{\bibinfo{person}{Zhiyun Lu}, \bibinfo{person}{Vikas
  Sindhwani}, {and} \bibinfo{person}{Tara~N. Sainath}.}
  \bibinfo{year}{2016}\natexlab{}.
\newblock \showarticletitle{{Learning Compact Recurrent Neural Networks}}. In
  \bibinfo{booktitle}{{\em IEEE International Conference on Acoustics, Speech
  and Signal Processing}}.
\newblock


\bibitem[\protect\citeauthoryear{Ma, Cao, Vrudhula, and Seo}{Ma
  et~al\mbox{.}}{2017}]%
        {NR_CNN_FXP_OPTIM_LOOP}
\bibfield{author}{\bibinfo{person}{Yufei Ma}, \bibinfo{person}{Yu Cao},
  \bibinfo{person}{Sarma Vrudhula}, {and} \bibinfo{person}{Jae-Sun Seo}.}
  \bibinfo{year}{2017}\natexlab{}.
\newblock \showarticletitle{{Optimizing Loop Operation and Dataflow in FPGA
  Acceleration of Deep Convolutional Neural Networks}}. In
  \bibinfo{booktitle}{{\em ACM/SIGDA International Symposium on
  Field-programmable Gate Arrays}}.
\newblock


\bibitem[\protect\citeauthoryear{Mellempudi, Kundu, Mudigere, Das, Kaul, and
  Dubey}{Mellempudi et~al\mbox{.}}{2017}]%
        {TNN_CNN_FGQ}
\bibfield{author}{\bibinfo{person}{Naveen Mellempudi}, \bibinfo{person}{Abhisek
  Kundu}, \bibinfo{person}{Dheevatsa Mudigere}, \bibinfo{person}{Dipankar Das},
  \bibinfo{person}{Bharat Kaul}, {and} \bibinfo{person}{Pradeep Dubey}.}
  \bibinfo{year}{2017}\natexlab{}.
\newblock \showarticletitle{{Ternary Neural Networks with Fine-grained
  Quantization}}.
\newblock \bibinfo{journal}{{\em arXiv preprint arXiv:1705.01462\/}}
  (\bibinfo{year}{2017}).
\newblock


\bibitem[\protect\citeauthoryear{Mishra, Nurvitadhi, Cook, and Marr}{Mishra
  et~al\mbox{.}}{2018}]%
        {BNN_CNN_WRPN}
\bibfield{author}{\bibinfo{person}{Asit Mishra}, \bibinfo{person}{Eriko
  Nurvitadhi}, \bibinfo{person}{Jeffrey~J. Cook}, {and} \bibinfo{person}{Debbie
  Marr}.} \bibinfo{year}{2018}\natexlab{}.
\newblock \showarticletitle{{WRPN: Wide Reduced-precision Networks}}. In
  \bibinfo{booktitle}{{\em International Conference on Learning
  Representations}}.
\newblock


\bibitem[\protect\citeauthoryear{Molchanov, Tyree, Karras, Aila, and
  Kautz}{Molchanov et~al\mbox{.}}{2017}]%
        {PRU_CNN_SALIENCY_CRITERIA}
\bibfield{author}{\bibinfo{person}{Pavlo Molchanov}, \bibinfo{person}{Stephen
  Tyree}, \bibinfo{person}{Tero Karras}, \bibinfo{person}{Timo Aila}, {and}
  \bibinfo{person}{Jan Kautz}.} \bibinfo{year}{2017}\natexlab{}.
\newblock \showarticletitle{{Pruning Convolutional Neural Networks for Resource
  Efficient Inference}}. In \bibinfo{booktitle}{{\em International Conference
  on Learning Representations}}.
\newblock


\bibitem[\protect\citeauthoryear{Monakov, Lokhmotov, and Avetisyan}{Monakov
  et~al\mbox{.}}{2010}]%
        {PRU_GPU_MONAKOV}
\bibfield{author}{\bibinfo{person}{Alexander Monakov}, \bibinfo{person}{Anton
  Lokhmotov}, {and} \bibinfo{person}{Arutyun Avetisyan}.}
  \bibinfo{year}{2010}\natexlab{}.
\newblock \showarticletitle{{Automatically Tuning Sparse Matrix-vector
  Multiplication for GPU Architectures}}. In \bibinfo{booktitle}{{\em
  International Conference on High-performance Embedded Architectures and
  Compilers}}.
\newblock


\bibitem[\protect\citeauthoryear{Moons and Verhelst}{Moons and
  Verhelst}{2016}]%
        {FXP_CNN_SCALABLE_PRECISION}
\bibfield{author}{\bibinfo{person}{Bert Moons} {and} \bibinfo{person}{Marian
  Verhelst}.} \bibinfo{year}{2016}\natexlab{}.
\newblock \showarticletitle{{A 0.3--2.6 TOPS/W Precision-scalable Processor for
  Real-time Large-scale ConvNets}}. In \bibinfo{booktitle}{{\em IEEE Symposium
  on VLSI Circuits}}.
\newblock


\bibitem[\protect\citeauthoryear{Moss, Krishnan, Nurvitadhi, Ratuszniak,
  Johnson, Sim, Mishra, Marr, Subhaschandra, and Leong}{Moss
  et~al\mbox{.}}{2018}]%
        {FXP_CNN_FPGA_HARPv2}
\bibfield{author}{\bibinfo{person}{Duncan Moss}, \bibinfo{person}{Srivatsan
  Krishnan}, \bibinfo{person}{Eriko Nurvitadhi}, \bibinfo{person}{Piotr
  Ratuszniak}, \bibinfo{person}{Chris Johnson}, \bibinfo{person}{Jaewoong Sim},
  \bibinfo{person}{Asit Mishra}, \bibinfo{person}{Debbie Marr},
  \bibinfo{person}{Suchit Subhaschandra}, {and} \bibinfo{person}{Philip H.~W.
  Leong}.} \bibinfo{year}{2018}\natexlab{}.
\newblock \showarticletitle{{A Customizable Matrix Multiplication Framework for
  the Intel HARPv2 Xeon + FPGA Platform}}. In \bibinfo{booktitle}{{\em
  ACM/SIGDA International Symposium on Field-programmable Gate Arrays}}.
\newblock


\bibitem[\protect\citeauthoryear{Neelakantan, Vilnis, Le, Sutskever, Kaiser,
  Kurach, and Martens}{Neelakantan et~al\mbox{.}}{2015}]%
        {TNN_CNN_GRAD_NOISE}
\bibfield{author}{\bibinfo{person}{Arvind Neelakantan}, \bibinfo{person}{Luke
  Vilnis}, \bibinfo{person}{Quoc~V. Le}, \bibinfo{person}{Ilya Sutskever},
  \bibinfo{person}{Lukasz Kaiser}, \bibinfo{person}{Karol Kurach}, {and}
  \bibinfo{person}{James Martens}.} \bibinfo{year}{2015}\natexlab{}.
\newblock \showarticletitle{{Adding Gradient Noise Improves Learning for Very
  Deep Networks}}. In \bibinfo{booktitle}{{\em International Conference on
  Learning Representations}}.
\newblock


\bibitem[\protect\citeauthoryear{Nowlan and Hinton}{Nowlan and Hinton}{1992}]%
        {SOFT_WEIGHT_SHARING}
\bibfield{author}{\bibinfo{person}{Steven~J. Nowlan} {and}
  \bibinfo{person}{Geoffrey~E. Hinton}.} \bibinfo{year}{1992}\natexlab{}.
\newblock \showarticletitle{{Simplifying Neural Networks by Soft
  Weight-sharing}}.
\newblock \bibinfo{journal}{{\em Neural Computation\/}} \bibinfo{volume}{{4},
  4} (\bibinfo{year}{1992}).
\newblock


\bibitem[\protect\citeauthoryear{Nurvitadhi, Cook, Mishra, Marr, Nealis,
  Colangelo, Ling, Capalija, Aydonat, Shumarayev, and Dasu}{Nurvitadhi
  et~al\mbox{.}}{2018}]%
        {FPGA_ASIC_EMIB_FPL}
\bibfield{author}{\bibinfo{person}{Eriko Nurvitadhi}, \bibinfo{person}{Jeff
  Cook}, \bibinfo{person}{Asit Mishra}, \bibinfo{person}{Debbie Marr},
  \bibinfo{person}{Kevin Nealis}, \bibinfo{person}{Philip Colangelo},
  \bibinfo{person}{Andrew Ling}, \bibinfo{person}{Davor Capalija},
  \bibinfo{person}{Utku Aydonat}, \bibinfo{person}{Sergey Shumarayev}, {and}
  \bibinfo{person}{Aravind Dasu}.} \bibinfo{year}{2018}\natexlab{}.
\newblock \showarticletitle{{In-package Domain-specific ASICs for Intel Stratix
  10 FPGAs: A Case Study of Accelerating Deep Learning Using TensorTile ASIC}}.
  In \bibinfo{booktitle}{{\em International Conference on Field-programmable
  Logic and Applications}}.
\newblock


\bibitem[\protect\citeauthoryear{Nurvitadhi, Venkatesh, Sim, Marr, Huang, Hock,
  Liew, Srivatsan, Moss, and Subhaschandra}{Nurvitadhi et~al\mbox{.}}{2017}]%
        {FXP_CNN_FPGA_GPU}
\bibfield{author}{\bibinfo{person}{Eriko Nurvitadhi}, \bibinfo{person}{Ganesh
  Venkatesh}, \bibinfo{person}{Jaewoong Sim}, \bibinfo{person}{Debbie Marr},
  \bibinfo{person}{Randy Huang}, \bibinfo{person}{Jason O.~G. Hock},
  \bibinfo{person}{Yeong~Tat Liew}, \bibinfo{person}{Krishnan Srivatsan},
  \bibinfo{person}{Duncan Moss}, {and} \bibinfo{person}{Suchit Subhaschandra}.}
  \bibinfo{year}{2017}\natexlab{}.
\newblock \showarticletitle{{Can FPGAs Beat GPUs in Accelerating
  Next-generation Deep Neural Networks?}}. In \bibinfo{booktitle}{{\em
  ACM/SIGDA International Symposium on Field-programmable Gate Arrays}}.
\newblock


\bibitem[\protect\citeauthoryear{Nvidia}{Nvidia}{2018a}]%
        {CUDA_BINARY_SHIFT}
\bibfield{author}{\bibinfo{person}{Nvidia}.} \bibinfo{year}{2018}\natexlab{a}.
\newblock \bibinfo{title}{{CUDA C Programming Guide}}.
\newblock   (\bibinfo{year}{2018}).
\newblock
\showURL{%
\url{https://docs.nvidia.com/cuda/cuda-c-programming-guide/index.html\#arithmetic-instructions}}


\bibitem[\protect\citeauthoryear{Nvidia}{Nvidia}{2018b}]%
        {NVIDIA_TURING}
\bibfield{author}{\bibinfo{person}{Nvidia}.} \bibinfo{year}{2018}\natexlab{b}.
\newblock \bibinfo{title}{{NVIDIA Turing Architecture Whitepaper}}.
\newblock   (\bibinfo{year}{2018}).
\newblock
\showURL{%
\url{https://www.nvidia.com/content/dam/en-zz/Solutions/design-visualization/technologies/turing-architecture/NVIDIA-Turing-Architecture-Whitepaper.pdf}}


\bibitem[\protect\citeauthoryear{Ofenbeck, Steinmann, Caparros, Spampinato, and
  Puschel}{Ofenbeck et~al\mbox{.}}{2014}]%
        {BG_ROOFLINE_MODEL}
\bibfield{author}{\bibinfo{person}{Georg Ofenbeck}, \bibinfo{person}{Ruedi
  Steinmann}, \bibinfo{person}{Victoria Caparros}, \bibinfo{person}{Daniele~G.
  Spampinato}, {and} \bibinfo{person}{Markus Puschel}.}
  \bibinfo{year}{2014}\natexlab{}.
\newblock \showarticletitle{{Applying the Roofline Model}}. In
  \bibinfo{booktitle}{{\em IEEE International Symposium on Performance Analysis
  of Systems and Software}}.
\newblock


\bibitem[\protect\citeauthoryear{Ott, Lin, Zhang, Liu, and Bengio}{Ott
  et~al\mbox{.}}{2016}]%
        {TNN_LSTM_BENGIO}
\bibfield{author}{\bibinfo{person}{Joachim Ott}, \bibinfo{person}{Zhouhan Lin},
  \bibinfo{person}{Ying Zhang}, \bibinfo{person}{Shih-Chii Liu}, {and}
  \bibinfo{person}{Yoshua Bengio}.} \bibinfo{year}{2016}\natexlab{}.
\newblock \showarticletitle{{Recurrent Neural Networks with Limited Numerical
  Precision}}.
\newblock \bibinfo{journal}{{\em arXiv preprint arXiv:1608.06902\/}}
  (\bibinfo{year}{2016}).
\newblock


\bibitem[\protect\citeauthoryear{Posewsky and Ziener}{Posewsky and
  Ziener}{2018}]%
        {PRU_FPGA_POSEWSKY_ZERO_SKIPPING}
\bibfield{author}{\bibinfo{person}{Thorbj{\"o}rn Posewsky} {and}
  \bibinfo{person}{Daniel Ziener}.} \bibinfo{year}{2018}\natexlab{}.
\newblock \showarticletitle{{Throughput Optimizations for FPGA-based Deep
  Neural Network Inference}}.
\newblock \bibinfo{journal}{{\em Microprocessors and Microsystems\/}}
  \bibinfo{volume}{60} (\bibinfo{year}{2018}).
\newblock


\bibitem[\protect\citeauthoryear{Prost-Boucle, Bourge, P{\'e}trot, Alemdar,
  Caldwell, and Leroy}{Prost-Boucle et~al\mbox{.}}{2017}]%
        {TNN_CNN_FPL}
\bibfield{author}{\bibinfo{person}{Adrien Prost-Boucle}, \bibinfo{person}{Alban
  Bourge}, \bibinfo{person}{Fr{\'e}d{\'e}ric P{\'e}trot},
  \bibinfo{person}{Hande Alemdar}, \bibinfo{person}{Nicholas Caldwell}, {and}
  \bibinfo{person}{Vincent Leroy}.} \bibinfo{year}{2017}\natexlab{}.
\newblock \showarticletitle{{Scalable High-performance Architecture for
  Convolutional Ternary Neural Networks on FPGA}}. In \bibinfo{booktitle}{{\em
  International Conference on Field-programmable Logic and Applications}}.
\newblock


\bibitem[\protect\citeauthoryear{Qiu, Wang, Yao, Guo, Li, Zhou, Yu, Tang, Xu,
  and Song}{Qiu et~al\mbox{.}}{2016}]%
        {NR_CNN_FXP_GOING_DEEPER}
\bibfield{author}{\bibinfo{person}{Jiantao Qiu}, \bibinfo{person}{Jie Wang},
  \bibinfo{person}{Song Yao}, \bibinfo{person}{Kaiyuan Guo},
  \bibinfo{person}{Boxun Li}, \bibinfo{person}{Erjin Zhou},
  \bibinfo{person}{Jincheng Yu}, \bibinfo{person}{Tianqi Tang},
  \bibinfo{person}{Ningyi Xu}, {and} \bibinfo{person}{Sen Song}.}
  \bibinfo{year}{2016}\natexlab{}.
\newblock \showarticletitle{{Going Deeper with Embedded FPGA Platform for
  Convolutional Neural Network}}. In \bibinfo{booktitle}{{\em ACM/SIGDA
  International Symposium on Field-programmable Gate Arrays}}.
\newblock


\bibitem[\protect\citeauthoryear{Rastegari, Ordonez, Redmon, and
  Farhadi}{Rastegari et~al\mbox{.}}{2016}]%
        {BNN_CNN_XNOR-Net}
\bibfield{author}{\bibinfo{person}{Mohammad Rastegari},
  \bibinfo{person}{Vicente Ordonez}, \bibinfo{person}{Joseph Redmon}, {and}
  \bibinfo{person}{Ali Farhadi}.} \bibinfo{year}{2016}\natexlab{}.
\newblock \showarticletitle{{XNOR-Net: ImageNet Classification Using Binary
  Convolutional Neural Networks}}. In \bibinfo{booktitle}{{\em European
  Conference on Computer Vision}}.
\newblock


\bibitem[\protect\citeauthoryear{Razlighi, Imani, Koushanfar, and
  Rosing}{Razlighi et~al\mbox{.}}{2017}]%
        {PRU_CNN_LOOKNN}
\bibfield{author}{\bibinfo{person}{Mohammad~Samragh Razlighi},
  \bibinfo{person}{Mohsen Imani}, \bibinfo{person}{Farinaz Koushanfar}, {and}
  \bibinfo{person}{Tajana Rosing}.} \bibinfo{year}{2017}\natexlab{}.
\newblock \showarticletitle{{LookNN: Neural Network with No Multiplication}}.
  In \bibinfo{booktitle}{{\em Design, Automation and Test in Europe}}.
\newblock


\bibitem[\protect\citeauthoryear{Reagen, Whatmough, Adolf, Rama, Lee, Lee,
  Hern{\'a}ndez-Lobato, Wei, and Brooks}{Reagen et~al\mbox{.}}{2016}]%
        {PRU_CNN_MINERVA}
\bibfield{author}{\bibinfo{person}{Brandon Reagen}, \bibinfo{person}{Paul
  Whatmough}, \bibinfo{person}{Robert Adolf}, \bibinfo{person}{Saketh Rama},
  \bibinfo{person}{Hyunkwang Lee}, \bibinfo{person}{Sae-Kyu Lee},
  \bibinfo{person}{Jos{\'e}~M. Hern{\'a}ndez-Lobato}, \bibinfo{person}{Gu-Yeon
  Wei}, {and} \bibinfo{person}{David Brooks}.} \bibinfo{year}{2016}\natexlab{}.
\newblock \showarticletitle{{Minerva: Enabling Low-power, Highly-accurate Deep
  Neural Network Accelerators}}. In \bibinfo{booktitle}{{\em ACM SIGARCH
  Computer Architecture News}}.
\newblock


\bibitem[\protect\citeauthoryear{Rizakis, Venieris, Kouris, and
  Bouganis}{Rizakis et~al\mbox{.}}{2018}]%
        {LRF_LSTM_SVD_BASED}
\bibfield{author}{\bibinfo{person}{Michalis Rizakis},
  \bibinfo{person}{Stylianos~I. Venieris}, \bibinfo{person}{Alexandros Kouris},
  {and} \bibinfo{person}{Christos-Savvas Bouganis}.}
  \bibinfo{year}{2018}\natexlab{}.
\newblock \showarticletitle{{Approximate FPGA-based LSTMs under Computation
  Time Constraints}}. In \bibinfo{booktitle}{{\em International Symposium on
  Applied Reconfigurable Computing}}.
\newblock


\bibitem[\protect\citeauthoryear{Romero, Ballas, Kahou, Chassang, Gatta, and
  Bengio}{Romero et~al\mbox{.}}{2015}]%
        {KD_FITNETS}
\bibfield{author}{\bibinfo{person}{Adriana Romero}, \bibinfo{person}{Nicolas
  Ballas}, \bibinfo{person}{Samira~Ebrahimi Kahou}, \bibinfo{person}{Antoine
  Chassang}, \bibinfo{person}{Carlo Gatta}, {and} \bibinfo{person}{Yoshua
  Bengio}.} \bibinfo{year}{2015}\natexlab{}.
\newblock \showarticletitle{{FITNets: Hints for Thin Deep Nets}}. In
  \bibinfo{booktitle}{{\em International Conference on Learning
  Representations}}.
\newblock


\bibitem[\protect\citeauthoryear{Rouhani, Mirhoseini, and Koushanfar}{Rouhani
  et~al\mbox{.}}{2016}]%
        {LRF_CNN_DELIGHT}
\bibfield{author}{\bibinfo{person}{Bita~D. Rouhani}, \bibinfo{person}{Azalia
  Mirhoseini}, {and} \bibinfo{person}{Farinaz Koushanfar}.}
  \bibinfo{year}{2016}\natexlab{}.
\newblock \showarticletitle{{Delight: Adding Energy Dimension to Deep Neural
  Networks}}. In \bibinfo{booktitle}{{\em International Symposium on Low Power
  Electronics and Design}}.
\newblock


\bibitem[\protect\citeauthoryear{Rouhani, Mirhoseini, and Koushanfar}{Rouhani
  et~al\mbox{.}}{2017}]%
        {LRF_CNN_DEEP3}
\bibfield{author}{\bibinfo{person}{Bita~Darvish Rouhani},
  \bibinfo{person}{Azalia Mirhoseini}, {and} \bibinfo{person}{Farinaz
  Koushanfar}.} \bibinfo{year}{2017}\natexlab{}.
\newblock \showarticletitle{{Deep3: Leveraging Three Levels of Parallelism for
  Efficient Deep Learning}}. In \bibinfo{booktitle}{{\em Design Automation
  Conference}}.
\newblock


\bibitem[\protect\citeauthoryear{Sakr, Kim, and Shanbhag}{Sakr
  et~al\mbox{.}}{2017}]%
        {ANALYTICAL_GUARANTEES_ON_NUMERICAL_PRECISION_OF_DNNS}
\bibfield{author}{\bibinfo{person}{Charbel Sakr}, \bibinfo{person}{Yongjune
  Kim}, {and} \bibinfo{person}{Naresh Shanbhag}.}
  \bibinfo{year}{2017}\natexlab{}.
\newblock \showarticletitle{{Analytical Guarantees on Numerical Precision of
  Deep Neural Networks}}. In \bibinfo{booktitle}{{\em International Conference
  on Machine Learning}}.
\newblock


\bibitem[\protect\citeauthoryear{Samragh, Ghasemzadeh, and Koushanfar}{Samragh
  et~al\mbox{.}}{2017}]%
        {PRU_CNN_FCCM}
\bibfield{author}{\bibinfo{person}{Mohammad Samragh}, \bibinfo{person}{Mohammad
  Ghasemzadeh}, {and} \bibinfo{person}{Farinaz Koushanfar}.}
  \bibinfo{year}{2017}\natexlab{}.
\newblock \showarticletitle{{Customizing Neural Networks for Efficient FPGA
  Implementation}}. In \bibinfo{booktitle}{{\em IEEE International Symposium on
  Field-programmable Custom Computing Machines}}.
\newblock


\bibitem[\protect\citeauthoryear{Schurman and Brutlag}{Schurman and
  Brutlag}{2009}]%
        {LATENCY_USER_EXPERIENCE}
\bibfield{author}{\bibinfo{person}{Eric Schurman} {and} \bibinfo{person}{Jake
  Brutlag}.} \bibinfo{year}{2009}\natexlab{}.
\newblock \showarticletitle{{The User and Business Impact of Server Delays,
  Additional Bytes, and HTTP Chunking in Web Search}}. In
  \bibinfo{booktitle}{{\em Velocity}}.
\newblock


\bibitem[\protect\citeauthoryear{See, Luong, and Manning}{See
  et~al\mbox{.}}{2016}]%
        {PRU_LSTM_NMT_TRAIN_PRUNE_RETRAIN}
\bibfield{author}{\bibinfo{person}{Abigail See}, \bibinfo{person}{Minh-Thang
  Luong}, {and} \bibinfo{person}{Christopher~D. Manning}.}
  \bibinfo{year}{2016}\natexlab{}.
\newblock \showarticletitle{{Compression of Neural Machine Translation Models
  via Pruning}}. In \bibinfo{booktitle}{{\em SIGNLL Conference on Computational
  Natural Language Learning}}.
\newblock


\bibitem[\protect\citeauthoryear{Sharify, Delm{\'a}s, Siu, Judd, and
  Moshovos}{Sharify et~al\mbox{.}}{2018a}]%
        {FXP_CNN_LOOM}
\bibfield{author}{\bibinfo{person}{Sayeh Sharify}, \bibinfo{person}{Alberto
  Delm{\'a}s}, \bibinfo{person}{Kevin Siu}, \bibinfo{person}{Patrick Judd},
  {and} \bibinfo{person}{Andreas Moshovos}.} \bibinfo{year}{2018}\natexlab{a}.
\newblock \showarticletitle{{Loom: Exploiting Weight and Activation Precisions
  to Accelerate Convolutional Neural Networks}}. In \bibinfo{booktitle}{{\em
  Design Automation Conference}}.
\newblock


\bibitem[\protect\citeauthoryear{Sharify, Mahmoud, Delm{\'a}s, Nikolic, and
  Moshovos}{Sharify et~al\mbox{.}}{2018b}]%
        {PRU_CNN_LACONIC}
\bibfield{author}{\bibinfo{person}{Sayeh Sharify}, \bibinfo{person}{Mostafa
  Mahmoud}, \bibinfo{person}{Alberto Delm{\'a}s}, \bibinfo{person}{Milos
  Nikolic}, {and} \bibinfo{person}{Andreas Moshovos}.}
  \bibinfo{year}{2018}\natexlab{b}.
\newblock \showarticletitle{{Laconic Deep Learning Computing}}.
\newblock \bibinfo{journal}{{\em arXiv preprint arXiv:1805.04513\/}}
  (\bibinfo{year}{2018}).
\newblock


\bibitem[\protect\citeauthoryear{Sharma, Park, Mahajan, Amaro, Kim, Shao,
  Mishra, and Esmaeilzadeh}{Sharma et~al\mbox{.}}{2016}]%
        {DNNWEAVER}
\bibfield{author}{\bibinfo{person}{Hardik Sharma}, \bibinfo{person}{Jongse
  Park}, \bibinfo{person}{Divya Mahajan}, \bibinfo{person}{Emmanuel Amaro},
  \bibinfo{person}{Joon~K. Kim}, \bibinfo{person}{Chenkai Shao},
  \bibinfo{person}{Asit Mishra}, {and} \bibinfo{person}{Hadi Esmaeilzadeh}.}
  \bibinfo{year}{2016}\natexlab{}.
\newblock \showarticletitle{{From High-level Deep Neural Models to FPGAs}}. In
  \bibinfo{booktitle}{{\em IEEE/ACM International Symposium on
  Microarchitecture}}.
\newblock


\bibitem[\protect\citeauthoryear{Sharma, Park, Suda, Lai, Chau, Chandra, and
  Esmaeilzadeh}{Sharma et~al\mbox{.}}{2018}]%
        {FXP_CNN_BIT_FUSION}
\bibfield{author}{\bibinfo{person}{Hardik Sharma}, \bibinfo{person}{Jongse
  Park}, \bibinfo{person}{Naveen Suda}, \bibinfo{person}{Liangzhen Lai},
  \bibinfo{person}{Benson Chau}, \bibinfo{person}{Vikas Chandra}, {and}
  \bibinfo{person}{Hadi Esmaeilzadeh}.} \bibinfo{year}{2018}\natexlab{}.
\newblock \showarticletitle{{Bit Fusion: Bit-level Dynamically Composable
  Architecture for Accelerating Deep Neural Network}}. In
  \bibinfo{booktitle}{{\em International Symposium on Computer Architecture}}.
\newblock


\bibitem[\protect\citeauthoryear{Shen, Huang, Wang, Qiao, Wen, and Zhang}{Shen
  et~al\mbox{.}}{2018}]%
        {FXP_CNN_FPGA_TOWARDS_A_UNIFORM}
\bibfield{author}{\bibinfo{person}{Junzhong Shen}, \bibinfo{person}{You Huang},
  \bibinfo{person}{Zelong Wang}, \bibinfo{person}{Yuran Qiao},
  \bibinfo{person}{Mei Wen}, {and} \bibinfo{person}{Chunyuan Zhang}.}
  \bibinfo{year}{2018}\natexlab{}.
\newblock \showarticletitle{{Towards a Uniform Template-based Architecture for
  Accelerating 2D and 3D CNNs on FPGA}}. In \bibinfo{booktitle}{{\em ACM/SIGDA
  International Symposium on Field-programmable Gate Arrays}}.
\newblock


\bibitem[\protect\citeauthoryear{Shin, Boo, and Sung}{Shin
  et~al\mbox{.}}{2017}]%
        {R_CNN_FXP_STEP_SIZE}
\bibfield{author}{\bibinfo{person}{Sungho Shin}, \bibinfo{person}{Yoonho Boo},
  {and} \bibinfo{person}{Wonyong Sung}.} \bibinfo{year}{2017}\natexlab{}.
\newblock \showarticletitle{{Fixed-point Optimization of Deep Neural Networks
  with Adaptive Step Size Retraining}}. In \bibinfo{booktitle}{{\em IEEE
  International Conference on Acoustics, Speech and Signal Processing}}.
\newblock


\bibitem[\protect\citeauthoryear{Shin, Hwang, and Sung}{Shin
  et~al\mbox{.}}{2016}]%
        {R_LSTM_FXP_DFXP}
\bibfield{author}{\bibinfo{person}{Sungho Shin}, \bibinfo{person}{Kyuyeon
  Hwang}, {and} \bibinfo{person}{Wonyong Sung}.}
  \bibinfo{year}{2016}\natexlab{}.
\newblock \showarticletitle{{Fixed-point Performance Analysis of Recurrent
  Neural Networks}}. In \bibinfo{booktitle}{{\em IEEE International Conference
  on Acoustics, Speech and Signal Processing}}.
\newblock


\bibitem[\protect\citeauthoryear{Silberman and Guadarrama}{Silberman and
  Guadarrama}{2016}]%
        {TF_SLIM_MODEL_LIBRARY}
\bibfield{author}{\bibinfo{person}{Nathan Silberman} {and}
  \bibinfo{person}{Sergio Guadarrama}.} \bibinfo{year}{2016}\natexlab{}.
\newblock \bibinfo{title}{{TensorFlow-Slim Image Classification Model
  Library}}.
\newblock   (\bibinfo{year}{2016}).
\newblock
\showURL{%
\url{https://github.com/tensorflow/models/tree/master/research/slim}}


\bibitem[\protect\citeauthoryear{Sindhwani, Sainath, and Kumar}{Sindhwani
  et~al\mbox{.}}{2015}]%
        {SM_CNN_TOEPLITZ_LIKE}
\bibfield{author}{\bibinfo{person}{Vikas Sindhwani}, \bibinfo{person}{Tara~N.
  Sainath}, {and} \bibinfo{person}{Sanjiv Kumar}.}
  \bibinfo{year}{2015}\natexlab{}.
\newblock \showarticletitle{{Structured Transforms for Small-footprint Deep
  Learning}}. In \bibinfo{booktitle}{{\em Conference on Neural Information
  Processing Systems}}.
\newblock


\bibitem[\protect\citeauthoryear{Srinivas and Babu}{Srinivas and Babu}{2015}]%
        {PRU_CNN_DATA_FREE}
\bibfield{author}{\bibinfo{person}{Suraj Srinivas} {and}
  \bibinfo{person}{R.~Venkatesh Babu}.} \bibinfo{year}{2015}\natexlab{}.
\newblock \showarticletitle{{Data-free Parameter Pruning for Deep Neural
  Networks}}.
\newblock \bibinfo{journal}{{\em arXiv preprint arXiv:1507.06149\/}}
  (\bibinfo{year}{2015}).
\newblock


\bibitem[\protect\citeauthoryear{Srivastava, Hinton, Krizhevsky, Sutskever, and
  Salakhutdinov}{Srivastava et~al\mbox{.}}{2014}]%
        {BG_DROPOUT}
\bibfield{author}{\bibinfo{person}{Nitish Srivastava},
  \bibinfo{person}{Geoffrey Hinton}, \bibinfo{person}{Alex Krizhevsky},
  \bibinfo{person}{Ilya Sutskever}, {and} \bibinfo{person}{Ruslan
  Salakhutdinov}.} \bibinfo{year}{2014}\natexlab{}.
\newblock \showarticletitle{{Dropout: A Simple Way to Prevent Neural Networks
  from Overfitting}}.
\newblock \bibinfo{journal}{{\em Journal of Machine Learning Research\/}}
  \bibinfo{volume}{{15}, 1} (\bibinfo{year}{2014}).
\newblock


\bibitem[\protect\citeauthoryear{Su, Faraone, Liu, Zhao, Thomas, Leong, and
  Cheung}{Su et~al\mbox{.}}{2018}]%
        {PRU_CNN_ALEX_SU}
\bibfield{author}{\bibinfo{person}{Jiang Su}, \bibinfo{person}{Julian Faraone},
  \bibinfo{person}{Junyi Liu}, \bibinfo{person}{Yiren Zhao},
  \bibinfo{person}{David~B. Thomas}, \bibinfo{person}{Philip H.~W. Leong},
  {and} \bibinfo{person}{Peter Y.~K. Cheung}.} \bibinfo{year}{2018}\natexlab{}.
\newblock \showarticletitle{{Redundancy-reduced MobileNet Acceleration on
  Reconfigurable Logic for ImageNet Classification}}. In
  \bibinfo{booktitle}{{\em International Symposium on Applied Reconfigurable
  Computing}}.
\newblock


\bibitem[\protect\citeauthoryear{Sung and Kum}{Sung and Kum}{1995}]%
        {Q-sensitivity-check}
\bibfield{author}{\bibinfo{person}{Wonyong Sung} {and} \bibinfo{person}{Ki-Il
  Kum}.} \bibinfo{year}{1995}\natexlab{}.
\newblock \showarticletitle{{Simulation-based Word-length Optimization Method
  for Fixed-point Digital Signal Processing Systems}}.
\newblock \bibinfo{journal}{{\em IEEE Transactions on Signal Processing\/}}
  \bibinfo{volume}{{43}, 12} (\bibinfo{year}{1995}).
\newblock


\bibitem[\protect\citeauthoryear{Sze, Chen, Yang, and Emer}{Sze
  et~al\mbox{.}}{2017}]%
        {SURV_EFFICIENT_DNN}
\bibfield{author}{\bibinfo{person}{Vivienne Sze}, \bibinfo{person}{Yu-Hsin
  Chen}, \bibinfo{person}{Tien-Ju Yang}, {and} \bibinfo{person}{Joel~S. Emer}.}
  \bibinfo{year}{2017}\natexlab{}.
\newblock \showarticletitle{{Efficient Processing of Deep Neural Networks: A
  Tutorial and Survey}}.
\newblock \bibinfo{journal}{{\em {Proceedings of the IEEE}\/}}
  \bibinfo{volume}{{105}, 12} (\bibinfo{year}{2017}).
\newblock


\bibitem[\protect\citeauthoryear{Szegedy, Ioffe, Vanhoucke, and Alemi}{Szegedy
  et~al\mbox{.}}{2017}]%
        {BG_CLASSIFICATION_2}
\bibfield{author}{\bibinfo{person}{Christian Szegedy}, \bibinfo{person}{Sergey
  Ioffe}, \bibinfo{person}{Vincent Vanhoucke}, {and}
  \bibinfo{person}{Alexander~A. Alemi}.} \bibinfo{year}{2017}\natexlab{}.
\newblock \showarticletitle{{Inception-v4, Inception-ResNet and the Impact of
  Residual Connections on Learning}}. In \bibinfo{booktitle}{{\em Association
  for the Advancement of Artificial Intelligence}}.
\newblock


\bibitem[\protect\citeauthoryear{Tai, Xiao, Zhang, and Wang}{Tai
  et~al\mbox{.}}{2016}]%
        {LRF_SVD_WITH_OPTIMISER}
\bibfield{author}{\bibinfo{person}{Cheng Tai}, \bibinfo{person}{Tong Xiao},
  \bibinfo{person}{Yi Zhang}, {and} \bibinfo{person}{Xiaogang Wang}.}
  \bibinfo{year}{2016}\natexlab{}.
\newblock \showarticletitle{{Convolutional Neural Networks with Low-rank
  Regularization}}. In \bibinfo{booktitle}{{\em International Conference on
  Learning Representations}}.
\newblock


\bibitem[\protect\citeauthoryear{Tang, Hua, and Wang}{Tang
  et~al\mbox{.}}{2017}]%
        {BNN_CNN_BINARY_CONSTRIANED_TRAINING}
\bibfield{author}{\bibinfo{person}{Wei Tang}, \bibinfo{person}{Gang Hua}, {and}
  \bibinfo{person}{Liang Wang}.} \bibinfo{year}{2017}\natexlab{}.
\newblock \showarticletitle{{How to Train a Compact Binary Neural Network with
  High Accuracy?}}. In \bibinfo{booktitle}{{\em Association for the Advancement
  of Artificial Intelligence}}.
\newblock


\bibitem[\protect\citeauthoryear{Ullrich, Meeds, and Welling}{Ullrich
  et~al\mbox{.}}{2017}]%
        {PRU_CNN_SOFT_WEIGHT_SHARING}
\bibfield{author}{\bibinfo{person}{Karen Ullrich}, \bibinfo{person}{Edward
  Meeds}, {and} \bibinfo{person}{Max Welling}.}
  \bibinfo{year}{2017}\natexlab{}.
\newblock \showarticletitle{{Soft Weight-sharing for Neural Network
  Compression}}. In \bibinfo{booktitle}{{\em International Conference on
  Learning Representations}}.
\newblock


\bibitem[\protect\citeauthoryear{Umuroglu, Fraser, Gambardella, Blott, Leong,
  Jahre, and Vissers}{Umuroglu et~al\mbox{.}}{2017}]%
        {BNN_CNN_FINN}
\bibfield{author}{\bibinfo{person}{Yaman Umuroglu},
  \bibinfo{person}{Nicholas~J. Fraser}, \bibinfo{person}{Giulio Gambardella},
  \bibinfo{person}{Michaela Blott}, \bibinfo{person}{Philip H.~W. Leong},
  \bibinfo{person}{Magnus Jahre}, {and} \bibinfo{person}{Kees Vissers}.}
  \bibinfo{year}{2017}\natexlab{}.
\newblock \showarticletitle{{FINN: A Framework for Fast, Scalable Binarized
  Neural Network Inference}}. In \bibinfo{booktitle}{{\em ACM/SIGDA
  International Symposium on Field-programmable Gate Arrays}}.
\newblock


\bibitem[\protect\citeauthoryear{Venieris and Bouganis}{Venieris and
  Bouganis}{2016}]%
        {NR_CNN_FXP_FPGACONVNET}
\bibfield{author}{\bibinfo{person}{Stylianos~I. Venieris} {and}
  \bibinfo{person}{Christos-Savvas Bouganis}.} \bibinfo{year}{2016}\natexlab{}.
\newblock \showarticletitle{{fpgaConvNet: A Framework for Mapping Convolutional
  Neural Networks on FPGAs}}. In \bibinfo{booktitle}{{\em IEEE International
  Symposium on Field-programmable Custom Computing Machines}}.
\newblock


\bibitem[\protect\citeauthoryear{Venieris and Bouganis}{Venieris and
  Bouganis}{2017}]%
        {FPGA_LATENCY_BOUGANIS_FPL}
\bibfield{author}{\bibinfo{person}{Stylianos~I. Venieris} {and}
  \bibinfo{person}{Christos-Savvas Bouganis}.} \bibinfo{year}{2017}\natexlab{}.
\newblock \showarticletitle{{Latency-driven Design for FPGA-based Convolutional
  Neural Networks}}. In \bibinfo{booktitle}{{\em International Conference on
  Field-programmable Logic and Applications}}.
\newblock


\bibitem[\protect\citeauthoryear{Wang, Davis, and Cheung}{Wang
  et~al\mbox{.}}{2018a}]%
        {FXP_CNN_PYNQ_ERWEI}
\bibfield{author}{\bibinfo{person}{Erwei Wang}, \bibinfo{person}{James~J.
  Davis}, {and} \bibinfo{person}{Peter Y.~K. Cheung}.}
  \bibinfo{year}{2018}\natexlab{a}.
\newblock \showarticletitle{{A PYNQ-based Framework for Rapid CNN
  Prototyping}}. In \bibinfo{booktitle}{{\em IEEE International Symposium on
  Field-programmable Custom Computing Machines}}.
\newblock


\bibitem[\protect\citeauthoryear{Wang, Li, Ding, Yuan, Qiu, Wang, and
  Liang}{Wang et~al\mbox{.}}{2018b}]%
        {SM_CNN_FPGA_C-LSTM}
\bibfield{author}{\bibinfo{person}{Shuo Wang}, \bibinfo{person}{Zhe Li},
  \bibinfo{person}{Caiwen Ding}, \bibinfo{person}{Bo Yuan},
  \bibinfo{person}{Qinru Qiu}, \bibinfo{person}{Yanzhi Wang}, {and}
  \bibinfo{person}{Yun Liang}.} \bibinfo{year}{2018}\natexlab{b}.
\newblock \showarticletitle{{C-LSTM: Enabling Efficient LSTM using Structured
  Compression Techniques on FPGAs}}. In \bibinfo{booktitle}{{\em ACM/SIGDA
  International Symposium on Field-programmable Gate Arrays}}.
\newblock


\bibitem[\protect\citeauthoryear{Wang, Lin, and Wang}{Wang
  et~al\mbox{.}}{2017}]%
        {SM_LSTM_HYBRID_CIRCULAR}
\bibfield{author}{\bibinfo{person}{Zhisheng Wang}, \bibinfo{person}{Jun Lin},
  {and} \bibinfo{person}{Zhongfeng Wang}.} \bibinfo{year}{2017}\natexlab{}.
\newblock \showarticletitle{{Accelerating Recurrent Neural Networks: A
  Memory-efficient Approach}}.
\newblock \bibinfo{journal}{{\em IEEE Transactions on VLSI Systems\/}}
  \bibinfo{volume}{{25}, 10} (\bibinfo{year}{2017}).
\newblock


\bibitem[\protect\citeauthoryear{Wen, Wu, Wang, Chen, and Li}{Wen
  et~al\mbox{.}}{2016}]%
        {PRU_CNN_STRUCTURED_SPARSITY}
\bibfield{author}{\bibinfo{person}{Wei Wen}, \bibinfo{person}{Chunpeng Wu},
  \bibinfo{person}{Yandan Wang}, \bibinfo{person}{Yiran Chen}, {and}
  \bibinfo{person}{Hai Li}.} \bibinfo{year}{2016}\natexlab{}.
\newblock \showarticletitle{{Learning Structured Sparsity in Deep Neural
  Networks}}. In \bibinfo{booktitle}{{\em Conference on Neural Information
  Processing Systems}}.
\newblock


\bibitem[\protect\citeauthoryear{Williamson}{Williamson}{1991}]%
        {DFXP_WILLIAMSON}
\bibfield{author}{\bibinfo{person}{Darrell Williamson}.}
  \bibinfo{year}{1991}\natexlab{}.
\newblock \showarticletitle{{Dynamically Scaled Fixed Point Arithmetic}}. In
  \bibinfo{booktitle}{{\em IEEE Pacific Rim Conference on Communications,
  Computers and Signal Processing Conference Proceedings}}.
\newblock


\bibitem[\protect\citeauthoryear{Wu, Leng, Wang, Hu, and Cheng}{Wu
  et~al\mbox{.}}{2016}]%
        {PRU_CNN_QUANTIZED_CNN}
\bibfield{author}{\bibinfo{person}{Jiaxiang Wu}, \bibinfo{person}{Cong Leng},
  \bibinfo{person}{Yuhang Wang}, \bibinfo{person}{Qinghao Hu}, {and}
  \bibinfo{person}{Jian Cheng}.} \bibinfo{year}{2016}\natexlab{}.
\newblock \showarticletitle{{Quantized Convolutional Neural Networks for Mobile
  Devices}}. In \bibinfo{booktitle}{{\em IEEE Conference on Computer Vision and
  Pattern Recognition}}.
\newblock


\bibitem[\protect\citeauthoryear{Wu, Li, Chen, and Shi}{Wu
  et~al\mbox{.}}{2018}]%
        {R_CNN_FXP_WAGE}
\bibfield{author}{\bibinfo{person}{Shuang Wu}, \bibinfo{person}{Guoqi Li},
  \bibinfo{person}{Feng Chen}, {and} \bibinfo{person}{Luping Shi}.}
  \bibinfo{year}{2018}\natexlab{}.
\newblock \showarticletitle{{Training and Inference with Integers in Deep
  Neural Networks}}. In \bibinfo{booktitle}{{\em International Conference on
  Learning Representations}}.
\newblock


\bibitem[\protect\citeauthoryear{Xilinx}{Xilinx}{2018}]%
        {XILINX_EVEREST}
\bibfield{author}{\bibinfo{person}{Xilinx}.} \bibinfo{year}{2018}\natexlab{}.
\newblock \bibinfo{title}{{Versal, the First Adaptive Compute Acceleration
  Platform}}.
\newblock   (\bibinfo{year}{2018}).
\newblock
\showURL{%
\url{https://www.xilinx.com/support/documentation/white_papers/wp505-versal-acap.pdf}}


\bibitem[\protect\citeauthoryear{Yang, Chen, and Sze}{Yang
  et~al\mbox{.}}{2017}]%
        {SALIENCY}
\bibfield{author}{\bibinfo{person}{Tien-Ju Yang}, \bibinfo{person}{Yu-Hsin
  Chen}, {and} \bibinfo{person}{Vivienne Sze}.}
  \bibinfo{year}{2017}\natexlab{}.
\newblock \showarticletitle{{Designing Energy-efficient Convolutional Neural
  Networks Using Energy-aware Pruning}}. In \bibinfo{booktitle}{{\em IEEE
  Conference on Computer Vision and Pattern Recognition}}.
\newblock


\bibitem[\protect\citeauthoryear{Yang, Howard, Chen, Zhang, Go, Sandler, Sze,
  and Adam}{Yang et~al\mbox{.}}{2018}]%
        {PRU_CNN_NETADAPT}
\bibfield{author}{\bibinfo{person}{Tien-Ju Yang}, \bibinfo{person}{Andrew
  Howard}, \bibinfo{person}{Bo Chen}, \bibinfo{person}{Xiao Zhang},
  \bibinfo{person}{Alec Go}, \bibinfo{person}{Mark Sandler},
  \bibinfo{person}{Vivienne Sze}, {and} \bibinfo{person}{Hartwig Adam}.}
  \bibinfo{year}{2018}\natexlab{}.
\newblock \showarticletitle{{NetAdapt: Platform-aware Neural Network Adaptation
  for Mobile Applications}}. In \bibinfo{booktitle}{{\em European Conference on
  Computer Vision}}.
\newblock


\bibitem[\protect\citeauthoryear{Yang, Moczulski, Denil, de~Freitas, Smola,
  Song, and Wang}{Yang et~al\mbox{.}}{2015}]%
        {SM_CNN_ADAPTIVE_FASTFOOD_TRANSFORM}
\bibfield{author}{\bibinfo{person}{Zichao Yang}, \bibinfo{person}{Marcin
  Moczulski}, \bibinfo{person}{Misha Denil}, \bibinfo{person}{Nando de
  Freitas}, \bibinfo{person}{Alex Smola}, \bibinfo{person}{Le Song}, {and}
  \bibinfo{person}{Ziyu Wang}.} \bibinfo{year}{2015}\natexlab{}.
\newblock \showarticletitle{{Deep Fried Convnets}}. In \bibinfo{booktitle}{{\em
  International Conference on Computer Vision}}.
\newblock


\bibitem[\protect\citeauthoryear{Zhang, Fang, Zhou, Pan, and Cong}{Zhang
  et~al\mbox{.}}{2016}]%
        {NR_CNN_FXP_CAFFEINE}
\bibfield{author}{\bibinfo{person}{Chen Zhang}, \bibinfo{person}{Zhenman Fang},
  \bibinfo{person}{Peipei Zhou}, \bibinfo{person}{Peichen Pan}, {and}
  \bibinfo{person}{Jason Cong}.} \bibinfo{year}{2016}\natexlab{}.
\newblock \showarticletitle{{Caffeine: Towards Uniformed Representation and
  Acceleration for Deep Convolutional Neural Networks}}. In
  \bibinfo{booktitle}{{\em International Conference On Computer Aided Design}}.
\newblock


\bibitem[\protect\citeauthoryear{Zhang and Li}{Zhang and Li}{2018}]%
        {PRU_FPGA_PQ_CNN_FCCM}
\bibfield{author}{\bibinfo{person}{Jialiang Zhang} {and} \bibinfo{person}{Jing
  Li}.} \bibinfo{year}{2018}\natexlab{}.
\newblock \showarticletitle{{PQ-CNN: Accelerating Product Quantized
  Convolutional Neural Network on FPGA}}. In \bibinfo{booktitle}{{\em
  International Symposium on Field-programmable Custom Computing Machines}}.
\newblock


\bibitem[\protect\citeauthoryear{Zhang, Liu, Ramachandran, Zhuge, Tang, Ouyang,
  Cheng, Rupnow, and Chen}{Zhang et~al\mbox{.}}{2017}]%
        {FXP_RNN_VS_GPU}
\bibfield{author}{\bibinfo{person}{Xiaofan Zhang}, \bibinfo{person}{Xinheng
  Liu}, \bibinfo{person}{Anand Ramachandran}, \bibinfo{person}{Chuanhao Zhuge},
  \bibinfo{person}{Shibin Tang}, \bibinfo{person}{Peng Ouyang},
  \bibinfo{person}{Zuofu Cheng}, \bibinfo{person}{Kyle Rupnow}, {and}
  \bibinfo{person}{Deming Chen}.} \bibinfo{year}{2017}\natexlab{}.
\newblock \showarticletitle{{High-performance Video Content Recognition with
  Long-term Recurrent Convolutional Network for FPGA}}. In
  \bibinfo{booktitle}{{\em International Conference on Field-programmable Logic
  and Applications}}.
\newblock


\bibitem[\protect\citeauthoryear{Zhao, Song, Zhang, Xing, Lin, Srivastava,
  Gupta, and Zhang}{Zhao et~al\mbox{.}}{2017}]%
        {BNN_CNN_FPGA17}
\bibfield{author}{\bibinfo{person}{Ritchie Zhao}, \bibinfo{person}{Weinan
  Song}, \bibinfo{person}{Wentao Zhang}, \bibinfo{person}{Tianwei Xing},
  \bibinfo{person}{Jeng-Hau Lin}, \bibinfo{person}{Mani Srivastava},
  \bibinfo{person}{Rajesh Gupta}, {and} \bibinfo{person}{Zhiru Zhang}.}
  \bibinfo{year}{2017}\natexlab{}.
\newblock \showarticletitle{{Accelerating Binarized Convolutional Neural
  Networks with Software-programmable FPGAs}}. In \bibinfo{booktitle}{{\em
  ACM/SIGDA International Symposium on Field-programmable Gate Arrays}}.
\newblock


\bibitem[\protect\citeauthoryear{Zhou, Yao, Guo, Xu, and Chen}{Zhou
  et~al\mbox{.}}{2016c}]%
        {EXP_CNN_INCR}
\bibfield{author}{\bibinfo{person}{Aojun Zhou}, \bibinfo{person}{Anbang Yao},
  \bibinfo{person}{Yiwen Guo}, \bibinfo{person}{Lin Xu}, {and}
  \bibinfo{person}{Yurong Chen}.} \bibinfo{year}{2016}\natexlab{c}.
\newblock \showarticletitle{{Incremental Network Quantization: Towards Lossless
  CNNs with Low-precision Weights}}. In \bibinfo{booktitle}{{\em International
  Conference on Learning Representations}}.
\newblock


\bibitem[\protect\citeauthoryear{Zhou, Alvarez, and Porikli}{Zhou
  et~al\mbox{.}}{2016a}]%
        {PRU_CNN_LESS_IS_MORE}
\bibfield{author}{\bibinfo{person}{Hao Zhou}, \bibinfo{person}{Jose~M.
  Alvarez}, {and} \bibinfo{person}{Fatih Porikli}.}
  \bibinfo{year}{2016}\natexlab{a}.
\newblock \showarticletitle{{Less is More: Towards Compact CNNs}}. In
  \bibinfo{booktitle}{{\em European Conference on Computer Vision}}.
\newblock


\bibitem[\protect\citeauthoryear{Zhou, Ni, Zhou, Wen, Wu, and Zou}{Zhou
  et~al\mbox{.}}{2016b}]%
        {BNN_CNN_DoReFa-Net}
\bibfield{author}{\bibinfo{person}{Shuchang Zhou}, \bibinfo{person}{Zekun Ni},
  \bibinfo{person}{Xinyu Zhou}, \bibinfo{person}{He Wen},
  \bibinfo{person}{Yuxin Wu}, {and} \bibinfo{person}{Yuheng Zou}.}
  \bibinfo{year}{2016}\natexlab{b}.
\newblock \showarticletitle{{DoReFa-Net: Training Low Bitwidth Convolutional
  Neural Networks with Low Bitwidth Gradients}}.
\newblock \bibinfo{journal}{{\em arXiv preprint arXiv:1606.06160\/}}
  (\bibinfo{year}{2016}).
\newblock


\bibitem[\protect\citeauthoryear{Zhu, Han, Mao, and Dally}{Zhu
  et~al\mbox{.}}{2017}]%
        {TNN_CNN_TTQ}
\bibfield{author}{\bibinfo{person}{Chenzhuo Zhu}, \bibinfo{person}{Song Han},
  \bibinfo{person}{Huizi Mao}, {and} \bibinfo{person}{William~J. Dally}.}
  \bibinfo{year}{2017}\natexlab{}.
\newblock \showarticletitle{{Trained Ternary Quantization}}. In
  \bibinfo{booktitle}{{\em International Conference on Learning
  Representations}}.
\newblock


\bibitem[\protect\citeauthoryear{Zhu, Dong, and Su}{Zhu et~al\mbox{.}}{2018}]%
        {BNN_CNN_BENN}
\bibfield{author}{\bibinfo{person}{Shilin Zhu}, \bibinfo{person}{Xin Dong},
  {and} \bibinfo{person}{Hao Su}.} \bibinfo{year}{2018}\natexlab{}.
\newblock \showarticletitle{{Binary Ensemble Neural Network: More Bits per
  Network or More Networks per Bit?}}
\newblock \bibinfo{journal}{{\em arXiv preprint arXiv:1806.07550\/}}
  (\bibinfo{year}{2018}).
\newblock


\end{thebibliography}

\end{document}